\documentclass[10pt]{article} % For LaTeX2e
\usepackage[preprint]{tmlr}

\usepackage{amsmath,amsfonts,bm}

\def\eqref#1{equation~\ref{#1}}
\def\1{\bm{1}}

\DeclareMathAlphabet{\mathsfit}{\encodingdefault}{\sfdefault}{m}{sl}
\SetMathAlphabet{\mathsfit}{bold}{\encodingdefault}{\sfdefault}{bx}{n}

\usepackage{hyperref}
\usepackage{url}
\usepackage{cleveref}
\usepackage{graphics}
\usepackage{graphicx}
\usepackage{booktabs}

\crefname{equation}{Eq.}{Eqs.}
\Crefname{equation}{Eq.}{Eqs.}
\crefname{table}{Tab.}{Tabs.}
\Crefname{table}{Tab.}{Tabs.}
\crefname{figure}{Fig.}{Figs.}
\Crefname{figure}{Fig.}{Figs.}
\crefname{section}{Sec.}{Secs.}
\Crefname{section}{Sec.}{Secs.}
\crefname{algorithm}{Algorithm}{Algorithms}
\Crefname{algorithm}{Algorithm}{Algorithms}

\definecolor{tab10blue}{rgb}{0.122,0.467,0.706}
\definecolor{tab10orange}{rgb}{1.0,0.498,0.055}
\definecolor{tab10green}{rgb}{0.173,0.627,0.173}
\definecolor{tab10red}{rgb}{0.839,0.153,0.157}
\definecolor{tab10purple}{rgb}{0.580,0.404,0.741}
\definecolor{tab10brown}{rgb}{0.549,0.337,0.294}
\definecolor{tab10pink}{rgb}{0.890,0.467,0.761}
\definecolor{tab10gray}{rgb}{0.499,0.499,0.499}
\definecolor{tab10yellow}{rgb}{0.737,0.741,0.133}
\definecolor{tab10cyan}{rgb}{0.090,0.745,0.812}

\ifdefined\HighlightModification
    \newcommand{\highlight}[1]{\textcolor{blue}{#1}}
    \newcommand{\removed}[1]{\textcolor{red}{#1}}
    \newcommand{\camhighlight}[1]{\textcolor{green}{#1}}
\else
    \newcommand{\highlight}[1]{#1}
    \newcommand{\camhighlight}[1]{#1}
    \newcommand{\removed}[1]{}
\fi

\title{Uniformity First: Uniformity-aware Test-time Adaptation of Vision-language Models against Image Corruption}

\author{\name Kazuki Adachi \email kazuki.adachi@ntt.com \\
\addr NTT, Inc. \\
Yokohama National University
\AUTHORAND
\name Shin'ya Yamaguchi \\
\addr NTT, Inc.
\AUTHORAND
\name Tomoki Hamagami \\
\addr Yokohama National University}

\def\month{08}  % Insert correct month for camera-ready version
\def\year{2026} % Insert correct year for camera-ready version
\def\openreview{\url{https://openreview.net/forum?id=YELPe35KIg}} % Insert correct link to OpenReview for camera-ready version

\usepackage{algorithm}
\usepackage{algorithmic}

\begin{document}

\maketitle

\begin{abstract}
    Pre-trained vision-language models, such as contrastive language-image pre-training (CLIP), have demonstrated a remarkable generalizability, enabling a wide range of applications, including zero-shot classification.
However, vision-language models still struggle to handle \emph{distribution shifts}, where input samples have large gaps from training ones.
We found that CLIP is especially vulnerable to image corruption, a type of realistic distribution shift caused by sensor conditions such as weather, light, or noise.
Collecting a new dataset from a test distribution for fine-tuning is highly costly since image corruption occurs unexpectedly and has a wide variety of types.
Thus, we investigate \emph{test-time adaptation (TTA)} of zero-shot classification, which enables on-the-fly adaptation to the test distribution with unlabeled test data.
Existing TTA methods for CLIP mainly focus on modifying image and text embeddings or predictions to address distribution shifts.
Although these methods can adapt to domain shifts, such as out-of-distribution or different renditions in input images, they fail to adapt to distribution shifts beyond domain shifts, e.g., image corruption.
We found that \emph{uniformity} of image embeddings, which is related to the amount of information, is a key factor that differentiates domain shifts and other distribution shifts.
To enable adaptation to image corruption, we propose a novel method called \emph{un}iformity-aware \emph{info}rmation-balanced TTA (UnInfo).
To address distribution shifts, we introduce uniformity-aware confidence maximization, information-aware loss balancing, and knowledge distillation from the exponential moving average (EMA) teacher.
Through experiments, we demonstrate that our UnInfo improves accuracy under image corruption by retaining information in terms of uniformity.
\camhighlight{The code is available at \url{https://github.com/kzkadc/uninfo}.}

\end{abstract}

\section{Introduction}\label{sec:introduction}
Vision-language models (VLMs) pre-trained on large-scale datasets, such as contrastive language-image pre-training~(CLIP)~\citep{clip_paper}, have demonstrated remarkable generalizability and rich feature representations.
Specifically, pre-trained VLMs have enabled various applications such as zero-shot transfer~\citep{clip_paper,ge2023improving,wang2023improving}, image/video retrieval~\citep{baldrati2022effective,fang2021clip2video}, and image generation~\citep{patashnik2021styleclip,ramesh2022hierarchical}.
VLMs owe their success to their rich feature representations that unify vision and language modalities and public availability of the pre-trained weights, such as OpenCLIP~\citep{ilharco_gabriel_2021_5143773,cherti2023reproducible,schuhmann2022laionb}.
However, despite their generalizability, VLMs still face a challenge in adapting to distribution shifts, i.e., making predictions on test datasets with large gaps from the training dataset~\citep{Zhang2022tip-adapter,huang2024lp++,chen2023plot,shu2022tpt,zhou2024test,karmanov2024efficient,zhang2024dual,zanella2024test,wang2024a,qian2024online}.

A naive way of adapting VLMs is to collect a dataset from the test distribution and fine-tune the entire model parameters or the head classifier.
However, labeled data from the test distribution may not be available because the distribution shift is unknown before deployment~\citep{Wang2021,cafe_adachi}.

To adapt to distribution shifts instantly after being deployed in the test distribution, \emph{test-time adaptation~(TTA)}~\citep{liang2023comprehensive,dong2025adapting}, a paradigm aiming to adapt models during testing using only unlabeled test data, has attracted attention.
In the context of VLMs, recent studies~\citep{shu2022tpt,zhou2024test,karmanov2024efficient,zhang2024dual,wang2024a,qian2024online,zanella2024test} have intensively studied the TTA for zero-shot classification, which is one of the most common applications of VLMs.
When the domain changes, typical text prompts can be suboptimal.
For example, in an art or illustration domain, text prompts such as ``\texttt{a photo of a [class name]}'' is not appropriate~\citep{shu2022tpt,zhou2022coop,zhou2022cocoop}.
In other words, there is a modality gap between text prompts and images in the embedding space, which is crucial for VLMs' generalization~\citep{liang2022mind,khattak2023maple,qian2024intra,Yamaguchi_2025_CVPR,eslami2025mitigate}, under domain shifts.
Thus, existing TTA methods for VLMs are mainly designed for domain shifts (also called natural distribution shifts), such as changes in rendition or out-of-distribution~(OOD)~\citep{imagenet-r,recht2019imagenet,Hendrycks_2021_CVPR,robust_global_representation_neurips2019} by modifying image and/or text embeddings during testing, which can be viewed as addressing the modality gap.

While existing TTA methods successfully adapt to the domain shifts, we found that they are prone to overfitting to domain shifts and degrade the performance on other types of distribution shifts, e.g., \emph{image corruption}~\citep{imagenet-c,sojka2023ar}.
When an image recognition system is deployed in the real world, the model faces various perturbations even in the same domain.
This is because of changes in weather, light conditions, noise, cameras, etc., which are crucial in a wide range of applications, such as autonomous driving or surveillance cameras~\citep{dai2018dark,volk2019robust,eastwood2022sourcefree,cafe_adachi,temp_adachi,enomoto2024test}.
Such perturbations occur unexpectedly even within a single domain.
In the existing literature on ordinary classification models, image corruption deteriorates the model's accuracy~\citep{imagenet-c,qin2022understanding}.

We examined the robustness of VLMs against various types of image corruption~\citep{imagenet-c,mintun2021on} in terms of zero-shot classification performance using CLIP-family models.
Through the experiment, we found that they significantly degrade the performance on corrupted images and that existing TTA methods can fail to improve performance.
Moreover, we analyzed image embeddings and CLIP's knowledge about image corruption to reveal the difference between domain shift and image corruption.
As a result, we experimentally found that image corruption also causes the modality gap between image and text embeddings, but the mechanism differs from domain shifts.
Under image corruption, the modality gap occurs by image embeddings being ``corrupted'' in terms of \emph{uniformity}, a measure related to the amount of input information retained in the embedding space~\citep{wang2020understanding,wang2023feature}.
In other words, the amount of input information retained in the image embeddings becomes small by image corruption.
Nevertheless, most of existing CLIP TTA methods for domain shifts address the modality gap by updating embedding vectors in a post-hoc manner without updating encoders, which implicitly assumes that embedding vectors are discriminative under domain shifts.
However, under image corruption, image embeddings retain less information, i.e., they are less discriminative.
This is the primary reason why existing CLIP TTA methods fail to maintain their performance on image corruption.
Furthermore, we found that CLIP models cannot sufficiently encode words related to image quality; thus, simple prompting techniques, such as ensembles or incorporating corruption names into prompts, cannot recover the performance degradation (\cref{sec:preliminary_experiment}).
From these observations, existing TTA methods or simple prompting techniques targeting domain shifts suffer from image corruption, highlighting the necessity of a novel TTA method suitable for a wider range of distribution shifts.

To enable the TTA of CLIP under various types of image corruption, we propose a novel method called \emph{\textbf{un}iformity-aware \textbf{info}rmation-balanced test-time adaptation~(UnInfo)}.
UnInfo addresses the fundamental challenge of image embeddings' corruption by updating the image encoder with a low-rank adapter (LoRA)~\citep{hu2022lora}.
To realize effective adaptation, UnInfo consists of three components: (i) uniformity-aware confidence maximization, (ii) information-aware loss balancing, and (iii) knowledge distillation from the exponential moving average~(EMA) teacher.
Uniformity-aware confidence maximization seeks to maximize prediction confidence in terms of entropy, while incorporating uniformity to prevent embeddings from losing input information.
The information-aware loss balancing adaptively controls the balance between confidence maximization and uniformity enhancement on the basis of mutual information so that uniformity is first leveraged and then confidence is addressed.
This balancing plays a critical role, specifically when confidence is unreliable because of severe image embedding corruption.
The knowledge distillation from the EMA teacher stabilizes the encoder update by tracking the EMA of LoRA parameters and regularizing the student's prediction to be close to the teacher's.

Through extensive experiments, our UnInfo improved the test zero-shot accuracy on various types of image corruption by incorporating uniformity and balancing priority between uniformity and entropy.

\section{Zero-shot Classification with CLIP}\label{sec:preliminary_clip}

Given a pre-trained CLIP composed of a text encoder $f^\text{txt}_{\theta_\text{txt}}: \mathcal{T} \to \mathbb{R}^d$ and image encoder $f^\text{img}_{\theta_\text{img}}:\mathcal{X} \to \mathbb{R}^d$, we first encode the text prompts to obtain text embeddings, which are used for the prototype of each class in the embedding space $\mathbb{R}^d$, where $\mathcal{T}$ and $\mathcal{X}$ are text and image input spaces, and $\theta_\text{txt}$ and $\theta_\text{img}$ are pre-trained weights of the encoders.
The text prompts typically consist of a template and class names, like ``\texttt{a photo of a [class name]},'' denoted by $\mathbf{p}_c$ for class $c$.
We denote the corresponding text embeddings as $\{ \mathbf{t}_c=f^\text{txt}_{\theta_\text{txt}}(\mathbf{p}_c) \}_{c=1}^C$, where $C$ is the total number of classes.
We assume the text embeddings are normalized, i.e., $\| \mathbf{t}_c \|_2=1$.

For a test image $\mathbf{x}\in \mathcal{X}$, we compute the image embedding $\mathbf{z}=f^\text{img}_{\theta_\text{img}}(\mathbf{x})$, where $\| \mathbf{z} \|_2=1$.
The similarity between the image and text embeddings $\mathbf{z}^\top \mathbf{t}_c$ is regarded as the logit for class $c$.
The zero-shot prediction probability is obtained by
\begin{equation}
    \hat{p}_c = \operatorname{softmax}(\mathbf{z}^\top [\mathbf{t}_1,\ldots, \mathbf{t}_C] / \tau)_c,\label{eq:clip_zeroshot_probability}
\end{equation}
where $\tau>0$ is the temperature parameter.
The final prediction is made by taking the argmax of $\hat{p}_c$.

\section{Preliminary Experiment}\label{sec:preliminary_experiment}
First, we empirically demonstrate the vulnerability of CLIP under image corruption in terms of zero-shot classification accuracy.
Moreover, CLIPs cannot properly encode images under such distribution shifts, i.e., the performance degradation cannot be recovered by simple prompting techniques because CLIPs cannot sufficiently represent concepts related to image quality.

\paragraph{Setup.} We evaluated a ViT-B/16 CLIP pre-trained on the LAION dataset~\citep{schuhmann2022laionb} downloaded via OpenCLIP~\citep{ilharco_gabriel_2021_5143773}.
We used the ImageNet-C~\citep{imagenet-c} dataset, which includes 15 types of image corruption simulating the image corruption (see \cref{ssec:exp_dataset} for dataset details).
We performed zero-shot classification with the text prompt ``\texttt{a photo of a [class name]}'' (Normal prompt).
We also used the ensemble of 80 text prompts, e.g., ``\texttt{a bad photo of a [class name]},'' ``\texttt{a photo of many [class name]},'' and so on, which is widely adopted as one baseline~\citep{clip_paper}~(Ensemble), to see the effectiveness of prompt engineering on image corruption.
To check the expressiveness of CLIP representation, we also examined text prompts that included descriptions of corruption.
Specifically, we used the prompts ``\texttt{a photo of a [class name] corrupted by [corruption name]}.''
For each corruption, we generated ten synonyms of the \texttt{corruption name} with GPT-4o~\citep{hurst2024gpt} and ensembled the text prompts (Corruption prompt).
Details of the prompt ensemble are provided in \cref{asec:text_prompt_ensemble} in the appendix.
We tested on ImageNet~\citep{imagenet}, ImageNet-A/R~\citep{Hendrycks_2021_CVPR,imagenet-r} (domain shifts), and ImageNet-C~\citep{imagenet-c} to observe the effect of image corruption.

\paragraph{Evaluation metrics.} We evaluated accuracy, entropy, uniformity loss, and the modality gap between text and image embeddings by the earth mover's distance (EMD).
The entropy measures the uncertainty of predictions (a lower value indicates high confidence), and the uniformity loss measures how image embeddings are uniformly distributed on the unit hypersphere, which is related to the amount of input information preserved in the image embedding~\citep{wang2020understanding} (a lower value indicates more information is preserved).
The modality gap measures distributional distance between image and text embeddings, which is related to the generalizability of CLIP~\citep{liang2022mind,khattak2023maple,qian2024intra,Yamaguchi_2025_CVPR,eslami2025mitigate}.
The entropy and uniformity loss are defined as follows:
\begin{align}
    \text{Entropy}         & :=\sum_{c=1}^C -\hat{p}_c \log \hat{p}_c,                                                          \\
    \text{Uniformity loss} & :=\mathbb{E}_{\mathbf{z}_1,\mathbf{z}_2}\left[ \exp(-\| \mathbf{z}_1-\mathbf{z}_2 \|_2^2) \right], \\
    \text{Modality gap}    & :=\operatorname{EMD}(\{\mathbf{z}_1,\ldots,\mathbf{z}_N\}, \{\mathbf{t}_1,\ldots, \mathbf{t}_C\}).
\end{align}
Details of EMD are provided in \cref{asec:emd_details} of the appendix.

\begin{table}
    \centering
    \caption{Zero-shot classification metrics of OpenCLIP ViT-B/16 with simple prompting techniques on ImageNet family and ImageNet-C.
    }
    \label{tab:preliminary_zs-accuracy_corruption}
    \resizebox{1\linewidth}{!}{
        \setlength{\tabcolsep}{3pt}
        \begin{tabular}{lllll|llllllllllllllll}\toprule
            Metric                                   & \rotatebox{60}{\shortstack{ImageNet}} & \rotatebox{60}{\shortstack{ImageNet-A}} & \rotatebox{60}{\shortstack{ImageNet-R}} & \rotatebox{60}{\shortstack{Domain shift\\Mean\highlight{$^*$}}} & \rotatebox{60}{\shortstack{Defocus\\blur}} & \rotatebox{60}{\shortstack{Glass\\blur}} & \rotatebox{60}{\shortstack{Motion\\blur}} & \rotatebox{60}{\shortstack{Zoom\\blur}} & \rotatebox{60}{\shortstack{Contrast}} & \rotatebox{60}{\shortstack{Elastic\\transform}} & \rotatebox{60}{\shortstack{Jpeg\\compression}} & \rotatebox{60}{\shortstack{Pixelate}} & \rotatebox{60}{\shortstack{Gaussian\\noise}} & \rotatebox{60}{\shortstack{Impulse\\noise}} & \rotatebox{60}{\shortstack{Shot\\noise}} & \rotatebox{60}{\shortstack{Brightness}} & \rotatebox{60}{\shortstack{Fog}} & \rotatebox{60}{\shortstack{Frost}} & \rotatebox{60}{\shortstack{Snow}} & \rotatebox{60}{\shortstack{Corruption\\Mean}} \\ \midrule
            Accuracy (Normal prompt, $\uparrow$)     & $66.97$                               & ${32.91}$                               & ${74.36}$                               & $58.08$                                                         & ${28.18}$                                  & ${11.81}$                                & ${19.13}$                                 & ${17.65}$                               & ${17.90}$                             & ${13.37}$                                       & ${36.77}$                                      & ${36.92}$                             & ${6.02}$                                     & ${6.12}$                                    & ${7.88}$                                 & ${54.75}$                               & ${34.39}$                        & ${27.32}$                          & ${27.77}$                         & ${23.06}$                                     \\
            Accuracy (Ensemble, $\uparrow$)          & $67.59$                               & $33.15$                                 & $76.51$                                 & $59.08$                                                         & ${29.09}$                                  & ${12.56}$                                & ${20.56}$                                 & ${19.12}$                               & ${18.55}$                             & ${14.34}$                                       & ${37.71}$                                      & ${37.89}$                             & ${6.36}$                                     & ${6.54}$                                    & ${8.27}$                                 & ${55.78}$                               & ${35.99}$                        & ${28.37}$                          & ${28.34}$                         & ${23.97}$                                     \\
            Accuracy (Corruption prompt, $\uparrow$) & -                                     & -                                       & -                                       & -                                                               & ${26.53}$                                  & ${12.15}$                                & ${18.39}$                                 & ${17.99}$                               & ${17.65}$                             & ${13.96}$                                       & ${36.29}$                                      & ${36.35}$                             & ${6.02}$                                     & ${6.10}$                                    & ${7.80}$                                 & ${53.93}$                               & ${33.43}$                        & ${27.34}$                          & ${26.55}$                         & ${22.70}$                                     \\
            Entropy ($\downarrow$)                   & $0.748$                               & $1.220$                                 & $0.595$                                 & $0.854$                                                         & ${2.011}$                                  & ${2.703}$                                & ${2.317}$                                 & ${2.245}$                               & ${3.247}$                             & ${2.297}$                                       & ${1.632}$                                      & ${1.615}$                             & ${3.773}$                                    & ${3.714}$                                   & ${3.671}$                                & ${1.061}$                               & ${1.681}$                        & ${1.905}$                          & ${2.048}$                         & ${2.395}$                                     \\
            Uniformity loss ($\downarrow$)           & $0.513$                               & $0.538$                                 & $0.500$                                 & $0.517$                                                         & ${0.682}$                                  & ${0.735}$                                & ${0.722}$                                 & ${0.715}$                               & ${0.744}$                             & ${0.706}$                                       & ${0.630}$                                      & ${0.641}$                             & ${0.855}$                                    & ${0.853}$                                   & ${0.839}$                                & ${0.601}$                               & ${0.665}$                        & ${0.655}$                          & ${0.686}$                         & ${0.715}$                                     \\
            Modality gap (EMD, $\downarrow$)         & ${1.291}$                             & ${1.333}$                               & ${1.348}$                               & ${1.324}$                                                       & ${1.298}$                                  & ${1.326}$                                & ${1.325}$                                 & ${1.327}$                               & ${1.337}$                             & ${1.341}$                                       & ${1.293}$                                      & ${1.296}$                             & ${1.299}$                                    & ${1.297}$                                   & ${1.296}$                                & ${1.293}$                               & ${1.307}$                        & ${1.315}$                          & ${1.308}$                         & ${1.311}$                                     \\ \bottomrule
            \multicolumn{21}{l}{\footnotesize \highlight{$^*$The mean over the vanilla ImageNet, A, and R.}}                                                                                                                                                                                                                                                                                                                                                                                                                                                                                                                                                                                                                                                                                                                                                                                                                                                      \\
        \end{tabular}
    }
\end{table}

\begin{table}
    \centering
    \caption{Zero-shot
        corruption type classification accuracy (\%).}
    \label{tab:preliminary_zs-corruption-accuracy}
    \resizebox{0.8\linewidth}{!}{
        \setlength{\tabcolsep}{3pt}
        \begin{tabular}{llllllllllllllll}\toprule
            %==============================
            \rotatebox{60}{\shortstack{Defocus\\blur}} & \rotatebox{60}{\shortstack{Glass\\blur}} & \rotatebox{60}{\shortstack{Motion\\blur}} & \rotatebox{60}{\shortstack{Zoom\\blur}} & \rotatebox{60}{\shortstack{Contrast}} & \rotatebox{60}{\shortstack{Elastic\\transform}} & \rotatebox{60}{\shortstack{Jpeg\\compression}} & \rotatebox{60}{\shortstack{Pixelate}} & \rotatebox{60}{\shortstack{Gaussian\\noise}} & \rotatebox{60}{\shortstack{Impulse\\noise}} & \rotatebox{60}{\shortstack{Shot\\noise}} & \rotatebox{60}{\shortstack{Brightness}} & \rotatebox{60}{\shortstack{Fog}} & \rotatebox{60}{\shortstack{Frost}} & \rotatebox{60}{\shortstack{Snow}} & Total     \\ \midrule
            ${68.76}$                                  & ${11.12}$                                & ${75.61}$                                 & ${2.61}$                                & ${0.36}$                              & ${18.91}$                                       & ${1.28}$                                       & ${8.64}$                              & ${30.28}$                                    & ${0.23}$                                    & ${0.70}$                                 & ${3.43}$                                & ${78.95}$                        & ${49.16}$                          & ${7.32}$                          & ${23.53}$ \\ \bottomrule
            %==============================
        \end{tabular}
    }
\end{table}

\paragraph{Results.} \cref{tab:preliminary_zs-accuracy_corruption} shows the results.
On the domain shifts, the prompt ensemble had 1\%pt accuracy improvement on average compared to the normal prompt.
On the other hand, ImageNet-C significantly degraded accuracy, and the ensemble did not result in significant improvement.
Moreover, including corruption information in the prompt (Corruption prompt) resulted in accuracy degradation.
Notably, the entropy and uniformity loss significantly increased compared to that of the domain shifts on all corruption types, while there is no significant difference in the modality gap.
\camhighlight{In other words, under image corruption, less semantic information is available in the image embeddings, which makes zero-shot prediction uncertain.
    Although uniformity does not directly measure semantic information, it is closely related to the mutual information between the input image and the embedding~\citep{wang2020understanding}, i.e., it measures the amount of information preserved.
    The preserved information includes semantic information, i.e., alignment with text embeddings. Thus, maintaining uniformity is necessary to preserve semantic information, as it cannot be recovered if uniformity is collapsed (i.e., all image embeddings are concentrated in a single region).}
\highlight{This is because ViTs or CNNs, which are typically used for image encoders, are vulnerable to such image corruption~\citep{imagenet-c,qin2022understanding}.
    Unlike domain shifts, image-corruption data have clean counterparts.
    In other words, image corruption can be viewed as a Markov chain $X \rightarrow X' \rightarrow Z$, where $X$ and $X'$ are the original image and its corrupted one, and $Z$ is the embedding.
    By the data processing inequality, $\mathcal{I}(X;X') \geq \mathcal{I}(X;Z)$, where $\mathcal{I}(\cdot; \cdot)$ is mutual information.
    By rewriting mutual information with entropy $\mathcal{H}(\cdot)$, we have $\mathcal{H}(X|X') + \mathcal{H}(Z) \leq \mathcal{H}(Z|X)+H(X)$.
    When corruption becomes severe, it becomes more difficult to recover the original image $X$ from the corrupted one $X'$, i.e., $\mathcal{H}(X|X')$ increases.
    Also, $\mathcal{H}(X)$ can be regarded as a constant because it depends on the original data.
    Thus, assuming that the fluctuation of $\mathcal{H}(Z|X)$ is small, $\mathcal{H}(Z)$ is likely to decrease, i.e., $Z$ becomes less diverse and distributes less uniformly.
    Intuitively, as image corruption becomes severe, the semantic content in images becomes more indistinguishable in the embedding space; image embeddings become more similar to each other, resulting in greater uniformity loss (i.e., less information).
    In such a situation, recovering semantic information from the image embeddings is impossible.
    Thus, enhancing uniformity first is necessary to maintain semantic information.}

Next, to check whether the CLIP recognizes image corruption types, we performed zero-shot classification of the 15 corruption types of input images instead of object categories using the text prompts ``\texttt{a photo corrupted by [corruption name]}.''
\cref{tab:preliminary_zs-corruption-accuracy} shows the results.
Most corruption types had poor accuracy, which suggests that CLIP cannot recognize image corruption types.
Even in the cases of corruption types with high accuracies (such as defocus blur, motion blur, and fog), the contribution of including the corruption information for object classification is limited or, even worse, as shown in \cref{tab:preliminary_zs-accuracy_corruption}.

From these observations, image corruption causes the modality gap as well as domain shifts, but has fundamentally different properties from domain shifts in terms of the uniformity and entropy, which suggests that the CLIP cannot properly encode images corrupted by the image corruption.

\begin{figure}[tb]
    \centering
    \includegraphics[width=0.8\linewidth]{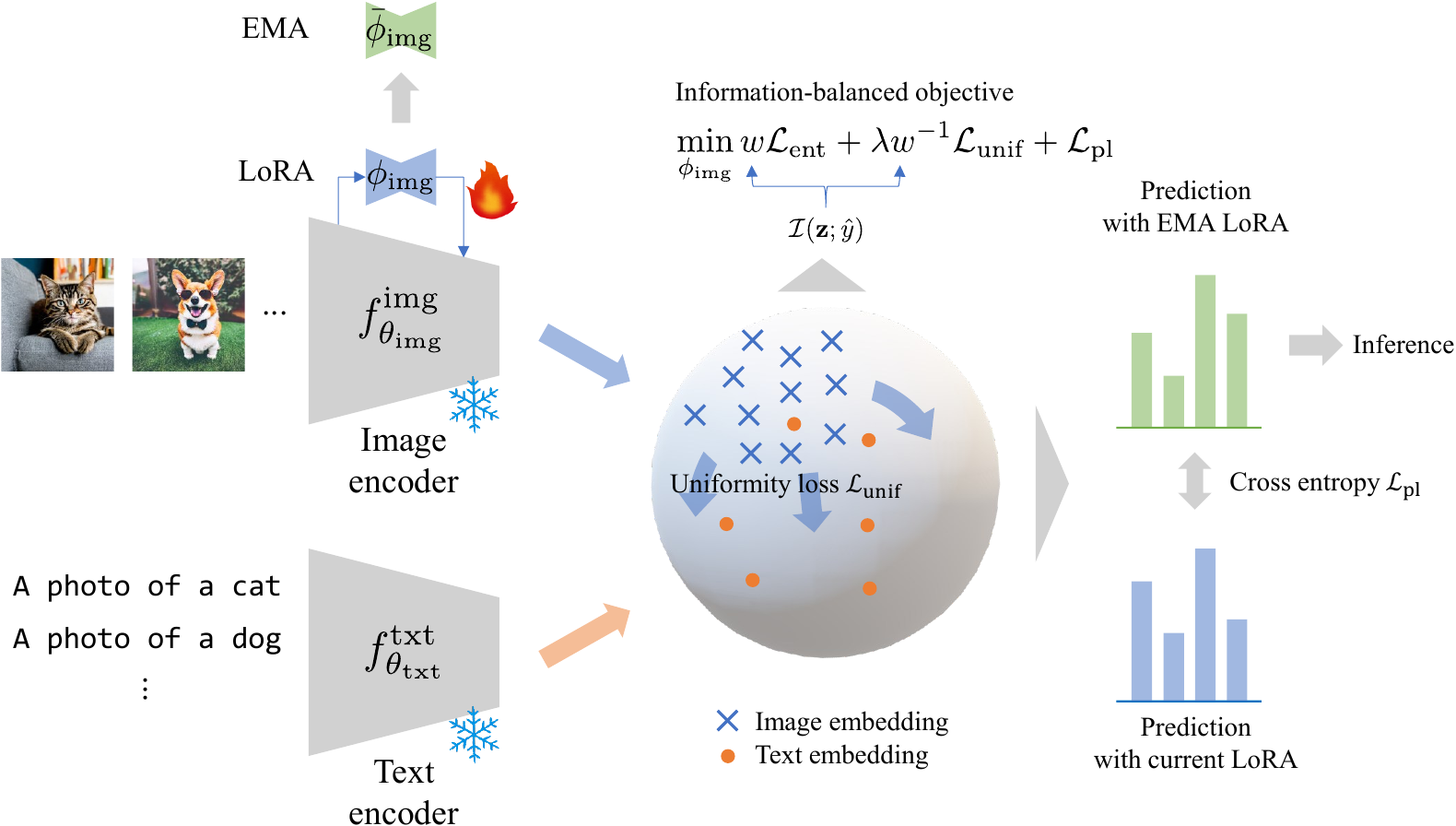}
    \caption{Overview of Uniformity-aware Information-balanced Test-time Adaptation (UnInfo).}
    \label{fig:overview}
\end{figure}

\section{Uniformity-aware Information-balanced Test-time Adaptation}\label{sec:proposed_method}

We introduce our proposed method, \emph{Uniformity-aware Information-balanced Test-time Adaptation~(UnInfo)}.
\cref{fig:overview} illustrates the overview of UnInfo.
The goal of TTA is to perform zero-shot classification on each incoming batch of test images $\{ \mathbf{x}_i \}_{i=1}^B$ with updating the model parameters to make accurate predictions, where $B$ is the batch size.
Since distribution shifts occur on input images and text embeddings are fixed in zero-shot classification, we specifically update $\theta_\text{img}$ in our method.

\subsection{Uniformity-aware Confidence Maximization}\label{ssec:proposed_method_main_loss}
Our method's basic approach is to enhance the confidence of predictions via minimizing entropy, which is widely adopted in existing TTA methods for general classification models~\citep{Wang2021,zhou2021bayesian,niu2022efficient,cafe_adachi,enomoto2024test,zhang2022memo,temp_adachi}.
In zero-shot classification with CLIP, the entropy loss is computed by using the prediction probability $\hat{p}_c$ on the basis of similarity, defined in \cref{eq:clip_zeroshot_probability}:
\begin{equation}
    \mathcal{L}_\text{ent} = \frac{1}{B}\sum_{i=1}^B \sum_{c=1}^C -\hat{p}_{i,c}\log \hat{p}_{i,c}.\label{eq:entropy_loss}
\end{equation}

Minimizing entropy is expected to improve the model performance because entropy is a promising proxy for classification accuracy when the images are appropriately embedded and input information is preserved, e.g., domain shifts.
However, the amount of input information preserved in the image embeddings decreases under image corruption as the entropy and uniformity loss increase in \cref{tab:preliminary_zs-accuracy_corruption}, as discussed in \cref{sec:preliminary_experiment}.
In such cases, the improvement by solely minimizing the entropy is limited since it attempts to leverage less information.
In other words, images are less distinguishable in the embedding space.
For enhancing information retained in image embeddings, we aim to improve the distinguishability of image embeddings from each other by minimizing the uniformity loss~\highlight{\citep{wang2020understanding}}:
\begin{equation}
    \mathcal{L}_\text{unif} = \log \frac{1}{B^2}\sum_{i=1}^B \sum_{j=1}^B \exp \left(  -\| \mathbf{z}_i - \mathbf{z}_j \|_2^2 \right). \label{eq:uniformity_loss}
\end{equation}
We minimize the entropy and uniformity losses simultaneously for refining prediction confidence and making image embeddings distinguishable (i.e., enhancing retained information):
\begin{equation}
    \mathcal{L} = \mathcal{L}_\text{ent} + \lambda \mathcal{L}_\text{unif}, \label{eq:proposed_loss_pre}
\end{equation}
where $\lambda>0$ is a hyperparameter, and $\mathbf{z}_i,\mathbf{z}_j$ are image embeddings of a batch of input images.

\subsection{Information-aware Loss Balancing}\label{ssec:proposed_method_balancing}
As described in \cref{ssec:proposed_method_main_loss,sec:preliminary_experiment}, minimizing the uniformity loss helps retain input information of image embeddings.
However, we found that the importance of uniformity and entropy can dynamically change during TTA.
For example, entropy should be prioritized when zero-shot classification works well, e.g., when image corruption is not severe.
In fact, uniformity differs depends on corruption types, which is demonstrated in the preliminary experiment (\cref{tab:preliminary_zs-accuracy_corruption}).
On the other hand, uniformity loss should be leveraged first, and then entropy should be addressed when zero-shot classification goes to a degenerated solution, e.g., classifying all images into a single class under severe image corruption.

To recognize the current regime and adaptively assign weights to the entropy and uniformity loss, we propose \emph{information-aware loss balancing}.
We employ the mutual information between the image embedding $\mathbf{z}$ and prediction $\hat{y}$, denoted by $\mathcal{I}(\mathbf{z}; \hat{y})$, to detect whether classification works without supervision.
We assign the weights to the two losses as follows:
\begin{equation}
    \mathcal{L} = w\mathcal{L}_\text{ent} + \lambda w^{-1} \mathcal{L}_\text{unif}, \quad w = \exp (\mathcal{I}(\mathbf{z};\hat{y})-\mathcal{I}_0), \label{eq:proposed_balaning_loss}
\end{equation}
where $\mathcal{I}_0$ is a hyperparameter to determine the threshold between the two regimes.
The mutual information $\mathcal{I}(\mathbf{z};\hat{y})$ is widely used in representation learning for measuring the quality of features and is computed as follows~\citep{NIPS1991_a8abb4bb,NIPS2010_42998cf3,Shi2012InformationTheoreticalLO,hu2017learning,liang2020we}:
\begin{equation}
    \mathcal{I}(\mathbf{z};\hat{y}) = \mathcal{H}(\hat{y}) - \mathcal{H}(\hat{y}|\mathbf{z})
    = \sum_{c=1}^C -\bar{p}_c \log \bar{p}_c - \frac{1}{B}\sum_{i=1}^B \sum_{c=1}^C -\hat{p}_{i,c} \log \hat{p}_{i,c},\label{eq:prediction_mutual_information}
\end{equation}
where $\mathcal{H}(\cdot)$ is entropy and $\bar{p}_c = (1/B)\sum_{i=1}^B \hat{p}_{i,c}$.
\highlight{Although $\mathcal{I}(\mathbf{z};\hat{y})$ is not always a precise estimation of the true mutual information since the batch size $B$ (e.g., $64$) can be far smaller than the number of classes $C$ (e.g., $1000$), it works on this purpose to verify whether the current classification goes well.}
Intuitively, $\mathcal{I}(\mathbf{z};\hat{y})$ takes small values when the image embedding $\mathbf{z}$ is less diverse and predictions are not confident.
In such a case, a larger weight is assigned to $\mathcal{L}_\text{unif}$ to make the image embeddings diverse and retain more information.
In the opposite case, $\mathcal{L}_\text{ent}$ is leveraged to make predictions more confident.
Note that gradients are not propagated to the weight $w$ since the balance is not an objective to be optimized.

\subsection{Update with Low-rank Adapters}\label{ssec:proposed_method_lora}
Although we aim to update the image encoder $f^\text{img}_{\theta_\text{img}}$ to minimize the proposed loss in \cref{eq:proposed_balaning_loss}, updating the whole $\theta_\text{img}$ naively leads to catastrophic model forgetting~\citep{Lai_2023_ICCV,vesdapunt2024hvclip}.
To avoid this, we fix the original pre-trained parameter $\theta_\text{img}$ and add the LoRA~\citep{hu2022lora} to the linear weights in the attention layers, inspired by \cite{Zanella_2024_CVPR}.
Specifically, given an attention layer at the $l$-th layer
\begin{equation}
    \mathbf{h}^{l+1} = \operatorname{softmax}\left( (W^l_\text{Q}\mathbf{h}^{l})(W^l_\text{K} \mathbf{h}^l)^\top / \sqrt{d^l} \right) W^l_\text{V}\mathbf{h}^l,\label{eq:attention_layer}
\end{equation}
we attach low-rank matrices to $W_\cdot^l$:
\begin{equation}
    W^l_\cdot \to W^l_\cdot + A^l_\cdot B^l_\cdot, \label{eq:lora_params}
\end{equation}
where $A^l_\cdot \in \mathbb{R}^{d^l\times d_\text{LoRA}}$ and $B^l_\cdot \in \mathbb{R}^{d_\text{LoRA}\times d^l}~(d_\text{LoRA}<d^l)$.
We denote the LoRA parameters as $\phi_\text{img}$.

\subsection{Knowledge Distillation from EMA Teacher}\label{ssec:proposed_method_ema}

For stabilizing the adaptation, we track the EMA of $\phi_\text{img}$ following previous studies~\citep{Wang_2022_CVPR,gao2022visual,Dobler_2023_CVPR,Wang_2024_WACV}.
That is, we update $\bar{\phi}_\text{img}$ in every iteration:
\begin{equation}
    \bar{\phi}_\text{img} \leftarrow m\phi_\text{img} + (1-m)\bar{\phi}_\text{img},\label{eq:ema_update}
\end{equation}
where $m\in (0,1)$ is the momentum parameter.
We adopt predictions made with $\bar{\phi}_\text{img}$ for inference.
Using $\bar{\phi}_\text{img}$ as the teacher, we penalize the current LoRA parameters $\phi_\text{img}$~(student) to make predictions close to those of the teacher.
Specifically, we take the cross entropy between the teacher and student outputs~\citep{Wang_2022_CVPR,gao2022visual}:
\begin{equation}
    \mathcal{L}_\text{pl} = \frac{1}{B} \sum_{i=1}^B \sum_{c=1}^C -\hat{q}_{i,c}\log \hat{p}_{i,c},\label{eq:cross_entropy}
\end{equation}
where $\hat{p}_{i,c}$ is the student's output defined in \cref{eq:clip_zeroshot_probability}, and $\hat{q}_{i,c}$ is the teacher's output computed in the same way with $\hat{p}_{i,c}$ but using the EMA parameter $\bar{\phi}_{\text{img}}$.

To sum up \cref{ssec:proposed_method_main_loss,ssec:proposed_method_balancing,ssec:proposed_method_lora,ssec:proposed_method_ema}, our objective is as follows:
\begin{equation}
    \min_{\phi_\text{img}} w\mathcal{L}_\text{ent} + \lambda w^{-1}\mathcal{L}_\text{unif} + \mathcal{L}_\text{pl}.\label{eq:proposed_objective}
\end{equation}

\section{Experiment}\label{sec:experiment}
We conducted experiments on TTA under image corruption.

\subsection{Datasets}\label{ssec:exp_dataset}
\paragraph{ImageNet-C~\citep{imagenet-c}:}
This dataset is constructed to evaluate the robustness of vision models.
It consists of the corrupted version of the validation set of ImageNet~\citep{imagenet}.
ImageNet-C includes 15 types of corruption, such as blur or digital noise.
Each corruption type has five severity levels.
We used the images corrupted at the highest severity level.

\paragraph{ImageNet-C-bar~\citep{mintun2021on}:}
This dataset is constructed to evaluate the robustness of the vision models on a broader range of corruption types.
Like ImageNet-C, it also consists of the corrupted version of the ImageNet validation set.
ImageNet-C-bar includes 10 corruption types that are algorithmically selected to be dissimilar from ImageNet-C.
We used the images corrupted at the highest severity level, as in the ImageNet-C case.

\subsection{Implementation}\label{ssec:exp_implementation}
We used the ViT-B/16 CLIP pre-trained on the \highlight{LAION-2B dataset~\citep{schuhmann2022laionb}}.
We downloaded the pre-trained weights via \highlight{HuggingFace\footnote{\url{https://huggingface.co/laion/CLIP-ViT-B-16-laion2B-s34B-b88K}}.}
We set the temperature of the zero-shot prediction $\tau=0.01$.
For our UnInfo, we used the AdamW optimizer~\citep{loshchilov2018decoupled} with learning rate $=0.001$, weight decay $=0.01$, and batch size $B=64$.
We set the weight of the uniformity loss $\lambda=1$ and the threshold of the information-aware loss balancing weight $\mathcal{I}_0=3$.
We chose these hyperparameters by using a few corruption types in ImageNet-C \highlight{(brightness, defocus blur, elastic transform, and Gaussian noise)} and used the selected hyperparameters for the others as default.
$\mathcal{I}_0$ was chosen on the basis of the mutual information computed on clean data (ImageNet).
For the LoRA in UnInfo, we used the implementation provided by \cite{Zanella_2024_CVPR}\footnote{\url{https://github.com/MaxZanella/CLIP-LoRA}}.
We set the LoRA hyperparameters, such as $\alpha$ and the rank $d_\text{LoRA}$, to the default values and the momentum of EMA $m=0.001$.
The LoRA matrices $A_\cdot^l$ and $B_\cdot^l$ are initialized by the Kaiming uniform initialization~\citep{He_2015_ICCV} and zero, respectively, so that image embeddings are unaffected by random $\phi_\text{img}$ at the initial state.

\subsection{Baseline}\label{ssec:exp_baseline}
We compared our UnInfo with existing TTA methods for CLIP zero-shot classification and few-shot adaptation methods.
For few-shot methods, we split $n\in \{1,5,10\}$ test samples per class, used them for adaptation, and then tested the model on the rest test samples.
\begin{description}
    \item[No-adapt:] Just performs zero-shot classification without adaptation.
    \item[LP (Linear probing):] Trains a linear classifier head with few-shot labeled samples from the test set.
    \item[Tip-adapter~\citep{Zhang2022tip-adapter}:] Modifies predictions by using cached features and logits of few-shot labeled samples from the test set.
    \item[TPT (Test-time prompt tuning)~\citep{shu2022tpt}:] Updates the text token embeddings corresponding to the words of a text template, e.g., ``\texttt{a photo of a},'' to minimize the marginal entropy over augmented views of an input image.
    \item[TDA (Training-free dynamic adapter)~\citep{karmanov2024efficient}:] Constructs positive and negative caches on the basis of prediction confidence on incoming test images and modifies predictions of subsequent inputs.
    \item[ZERO~\citep{farina2024frustratingly}:] Performs voting within predictions of augmented views of an input image.
    \item[MTA (MeanShift for test-time augmentation)~\citep{zanella2024test}:] Selects reliable image embeddings among augmented views and modifies the image embedding.
    \item[\highlight{DMN (Dual memory network)~\citep{zhang2024dual}:}] \highlight{Caches test image embeddings in a memory and utilizes similar cached embeddings for the current prediction.}
    \item[\highlight{Mint~\citep{bao2026mint}:}] \highlight{Updates the image encoder to optimize inter- and intra-class variances to maintain separability of image embeddings.}
    \item[\highlight{BATCLIP~\citep{Maharana_2025_ICCV}:}] \highlight{Updates both image and text encoders to refine confidence and improve separability while maintaining image-text matching.}
\end{description}

\subsection{Results}\label{ssec:exp_result}

\subsubsection{Adaptation Performance}\label{sssec:exp_adaptataion_performance}

\begin{table}[tb]
    \centering
    \caption{\highlight{Test accuracy (\%) on ImageNet-C. The numbers (1), (5), and (10) presented with the method names are the shot numbers per class $n$ used for the few-shot adaptation methods.}}
    \label{tab:imagenet-c_accuracy}
    \resizebox{1\linewidth}{!}{
        \setlength{\tabcolsep}{3pt}
        \begin{tabular}{lllllllllllllllll} \toprule
            %================
            Method                                         & \rotatebox{60}{\shortstack{Defocus\\blur}} & \rotatebox{60}{\shortstack{Glass\\blur}} & \rotatebox{60}{\shortstack{Motion\\blur}} & \rotatebox{60}{\shortstack{Zoom\\blur}} & \rotatebox{60}{\shortstack{Contrast}} & \rotatebox{60}{\shortstack{Elastic\\transform}} & \rotatebox{60}{\shortstack{Jpeg\\compression}} & \rotatebox{60}{\shortstack{Pixelate}} & \rotatebox{60}{\shortstack{Gaussian\\noise}} & \rotatebox{60}{\shortstack{Impulse\\noise}} & \rotatebox{60}{\shortstack{Shot\\noise}} & \rotatebox{60}{\shortstack{Brightness}} & \rotatebox{60}{\shortstack{Fog}}   & \rotatebox{60}{\shortstack{Frost}} & \rotatebox{60}{\shortstack{Snow}}  & Mean                               \\ \midrule
            No-adapt                                       & ${28.31}$                                  & ${11.89}$                                & ${19.16}$                                 & ${17.61}$                               & ${17.87}$                             & ${13.22}$                                       & ${36.79}$                                      & ${37.01}$                             & ${6.08}$                                     & ${6.17}$                                    & ${7.85}$                                 & ${54.89}$                               & ${34.55}$                          & ${27.35}$                          & ${27.67}$                          & ${23.09}$                          \\ \midrule
            Linear probing (1)                             & ${{11.53}_{\pm 0.59}}$                     & ${{6.15}_{\pm 0.20}}$                    & ${{8.79}_{\pm 0.19}}$                     & ${{9.02}_{\pm 0.41}}$                   & ${{5.87}_{\pm 0.13}}$                 & ${{8.17}_{\pm 0.39}}$                           & ${{15.57}_{\pm 0.35}}$                         & ${{16.36}_{\pm 0.46}}$                & ${{2.79}_{\pm 0.15}}$                        & ${{2.76}_{\pm 0.14}}$                       & ${{3.41}_{\pm 0.16}}$                    & ${{27.05}_{\pm 0.58}}$                  & ${{16.43}_{\pm 0.53}}$             & ${{9.88}_{\pm 0.18}}$              & ${{11.87}_{\pm 0.41}}$             & ${{10.38}_{\pm 0.02}}$             \\
            Linear probing (5)                             & ${{19.55}_{\pm 0.18}}$                     & ${{10.53}_{\pm 0.10}}$                   & ${{15.16}_{\pm 0.24}}$                    & ${{15.25}_{\pm 0.11}}$                  & ${{10.04}_{\pm 0.41}}$                & ${{14.47}_{\pm 0.05}}$                          & ${{25.68}_{\pm 0.08}}$                         & ${{27.02}_{\pm 0.09}}$                & ${{4.76}_{\pm 0.15}}$                        & ${{4.95}_{\pm 0.22}}$                       & ${{5.62}_{\pm 0.16}}$                    & ${{43.28}_{\pm 0.34}}$                  & ${{26.10}_{\pm 0.15}}$             & ${{17.31}_{\pm 0.12}}$             & ${{19.48}_{\pm 0.11}}$             & ${{17.28}_{\pm 0.04}}$             \\
            Linear probing (10)                            & ${{23.84}_{\pm 0.14}}$                     & ${{12.72}_{\pm 0.13}}$                   & ${{17.97}_{\pm 0.05}}$                    & ${{18.57}_{\pm 0.32}}$                  & ${{12.45}_{\pm 0.11}}$                & ${{17.69}_{\pm 0.10}}$                          & ${{30.50}_{\pm 0.24}}$                         & ${{32.32}_{\pm 0.18}}$                & ${{5.78}_{\pm 0.19}}$                        & ${{6.04}_{\pm 0.13}}$                       & ${{6.83}_{\pm 0.21}}$                    & ${{49.90}_{\pm 0.22}}$                  & ${{31.15}_{\pm 0.32}}$             & ${{21.49}_{\pm 0.29}}$             & ${{23.63}_{\pm 0.33}}$             & ${{20.73}_{\pm 0.06}}$             \\
            Tip-adapter~\citep{Zhang2022tip-adapter} (1)   & ${{19.00}_{\pm 0.31}}$                     & ${{9.09}_{\pm 0.22}}$                    & ${{13.92}_{\pm 0.32}}$                    & ${{14.04}_{\pm 0.36}}$                  & ${{9.53}_{\pm 0.15}}$                 & ${{12.60}_{\pm 0.31}}$                          & ${{26.98}_{\pm 0.12}}$                         & ${{27.19}_{\pm 0.29}}$                & ${{4.06}_{\pm 0.12}}$                        & ${{4.02}_{\pm 0.13}}$                       & ${{5.15}_{\pm 0.21}}$                    & ${{44.92}_{\pm 0.25}}$                  & ${{25.99}_{\pm 0.35}}$             & ${{18.10}_{\pm 0.21}}$             & ${{19.28}_{\pm 0.41}}$             & ${{16.92}_{\pm 0.07}}$             \\
            Tip-adapter (5)                                & ${{23.43}_{\pm 0.27}}$                     & ${{12.27}_{\pm 0.03}}$                   & ${{17.61}_{\pm 0.14}}$                    & ${{17.76}_{\pm 0.20}}$                  & ${{11.56}_{\pm 0.30}}$                & ${{16.82}_{\pm 0.17}}$                          & ${{30.87}_{\pm 0.21}}$                         & ${{32.09}_{\pm 0.20}}$                & ${{4.88}_{\pm 0.24}}$                        & ${{4.86}_{\pm 0.12}}$                       & ${{6.03}_{\pm 0.19}}$                    & ${{49.88}_{\pm 0.39}}$                  & ${{30.59}_{\pm 0.18}}$             & ${{21.43}_{\pm 0.21}}$             & ${{22.82}_{\pm 0.15}}$             & ${{20.19}_{\pm 0.06}}$             \\
            Tip-adapter (10)                               & ${{26.11}_{\pm 0.11}}$                     & ${{13.99}_{\pm 0.22}}$                   & ${{19.85}_{\pm 0.36}}$                    & ${{20.23}_{\pm 0.23}}$                  & ${{12.99}_{\pm 0.15}}$                & ${{19.26}_{\pm 0.22}}$                          & ${{33.03}_{\pm 0.11}}$                         & ${{34.60}_{\pm 0.21}}$                & ${{5.52}_{\pm 0.06}}$                        & ${{5.82}_{\pm 0.08}}$                       & ${{6.82}_{\pm 0.15}}$                    & ${{52.40}_{\pm 0.24}}$                  & ${{33.15}_{\pm 0.19}}$             & ${{24.01}_{\pm 0.32}}$             & ${{24.83}_{\pm 0.19}}$             & ${{22.17}_{\pm 0.05}}$             \\ \midrule
            TPT~\citep{shu2022tpt}                         & ${{29.66}_{\pm 0.05}}$                     & ${{12.87}_{\pm 0.03}}$                   & ${{21.11}_{\pm 0.02}}$                    & ${{20.54}_{\pm 0.02}}$                  & ${{20.11}_{\pm 0.08}}$                & ${{15.21}_{\pm 0.03}}$                          & ${{39.27}_{\pm 0.01}}$                         & ${{41.14}_{\pm 0.03}}$                & ${{6.48}_{\pm 0.03}}$                        & ${{6.74}_{\pm 0.02}}$                       & ${{8.50}_{\pm 0.01}}$                    & ${{57.35}_{\pm 0.04}}$                  & ${{37.05}_{\pm 0.07}}$             & ${{29.81}_{\pm 0.06}}$             & ${{30.23}_{\pm 0.04}}$             & ${{25.07}_{\pm 0.01}}$             \\
            ZERO~\citep{farina2024frustratingly}           & ${{26.85}_{\pm 0.05}}$                     & ${{8.86}_{\pm 0.06}}$                    & ${{18.11}_{\pm 0.04}}$                    & ${{19.89}_{\pm 0.04}}$                  & ${{16.46}_{\pm 0.02}}$                & ${{12.38}_{\pm 0.02}}$                          & ${{35.09}_{\pm 0.05}}$                         & ${{37.44}_{\pm 0.04}}$                & ${{3.69}_{\pm 0.02}}$                        & ${{5.33}_{\pm 0.02}}$                       & ${{4.43}_{\pm 0.03}}$                    & ${{53.35}_{\pm 0.08}}$                  & ${{33.50}_{\pm 0.01}}$             & ${{26.56}_{\pm 0.10}}$             & ${{27.42}_{\pm 0.01}}$             & ${{21.96}_{\pm 0.02}}$             \\
            MTA~\citep{zanella2024test}                    & ${{27.79}_{\pm 0.10}}$                     & ${{11.29}_{\pm 0.04}}$                   & ${{19.25}_{\pm 0.04}}$                    & ${{18.88}_{\pm 0.02}}$                  & ${{21.18}_{\pm 0.05}}$                & ${{13.92}_{\pm 0.03}}$                          & ${{37.23}_{\pm 0.04}}$                         & ${{38.95}_{\pm 0.04}}$                & ${{2.41}_{\pm 0.01}}$                        & ${{2.87}_{\pm 0.00}}$                       & ${{2.96}_{\pm 0.01}}$                    & ${{53.56}_{\pm 0.02}}$                  & ${{34.32}_{\pm 0.05}}$             & ${{28.02}_{\pm 0.07}}$             & ${{28.66}_{\pm 0.04}}$             & ${{22.75}_{\pm 0.02}}$             \\
            TDA~\citep{karmanov2024efficient}              & $\underline{{{30.13}_{\pm 0.04}}}$         & ${{14.59}_{\pm 0.08}}$                   & ${{22.10}_{\pm 0.03}}$                    & ${{21.09}_{\pm 0.06}}$                  & ${{19.59}_{\pm 0.06}}$                & ${{17.15}_{\pm 0.08}}$                          & ${{38.58}_{\pm 0.03}}$                         & ${{39.53}_{\pm 0.03}}$                & ${{7.23}_{\pm 0.01}}$                        & ${{7.45}_{\pm 0.03}}$                       & ${{9.34}_{\pm 0.07}}$                    & ${{56.99}_{\pm 0.04}}$                  & ${{38.09}_{\pm 0.06}}$             & ${{30.24}_{\pm 0.05}}$             & ${{31.02}_{\pm 0.02}}$             & ${{25.54}_{\pm 0.03}}$             \\
            \highlight{DMN~\citep{zhang2024dual}}          & ${{29.81}_{\pm 0.01}}$                     & ${{12.97}_{\pm 0.00}}$                   & ${{20.89}_{\pm 0.00}}$                    & ${{19.52}_{\pm 0.00}}$                  & ${{18.69}_{\pm 0.00}}$                & ${{14.36}_{\pm 0.00}}$                          & ${{38.93}_{\pm 0.00}}$                         & ${{39.69}_{\pm 0.00}}$                & ${{6.65}_{\pm 0.00}}$                        & ${{6.89}_{\pm 0.00}}$                       & ${{8.58}_{\pm 0.00}}$                    & ${{57.65}_{\pm 0.00}}$                  & ${{36.57}_{\pm 0.00}}$             & ${{28.98}_{\pm 0.00}}$             & ${{29.66}_{\pm 0.00}}$             & ${{24.65}_{\pm 0.00}}$             \\
            \highlight{Mint~\citep{bao2026mint}}           & ${{24.85}_{\pm 0.08}}$                     & $\mathbf{{{28.49}_{\pm 0.06}}}$          & $\mathbf{{{25.24}_{\pm 0.21}}}$           & $\mathbf{{{28.45}_{\pm 0.02}}}$         & $\mathbf{{{24.16}_{\pm 0.12}}}$       & $\underline{{{28.39}_{\pm 0.09}}}$              & $\mathbf{{{43.88}_{\pm 0.03}}}$                & $\mathbf{{{43.60}_{\pm 0.16}}}$       & $\underline{{{8.05}_{\pm 0.06}}}$            & ${{7.94}_{\pm 0.14}}$                       & ${{7.71}_{\pm 0.12}}$                    & $\mathbf{{{58.08}_{\pm 0.06}}}$         & $\mathbf{{{43.61}_{\pm 0.12}}}$    & $\underline{{{31.07}_{\pm 0.10}}}$ & $\mathbf{{{33.50}_{\pm 0.02}}}$    & $\mathbf{{{29.14}_{\pm 0.01}}}$    \\
            \highlight{BATCLIP~\citep{Maharana_2025_ICCV}} & ${{22.48}_{\pm 0.34}}$                     & $\underline{{{24.47}_{\pm 0.16}}}$       & $\underline{{{24.72}_{\pm 0.12}}}$        & $\underline{{{26.64}_{\pm 0.11}}}$      & $\underline{{{23.69}_{\pm 0.53}}}$    & $\mathbf{{{31.79}_{\pm 0.30}}}$                 & ${{33.49}_{\pm 0.17}}$                         & ${{35.93}_{\pm 0.16}}$                & $\mathbf{{{12.56}_{\pm 0.14}}}$              & $\mathbf{{{13.01}_{\pm 0.20}}}$             & $\mathbf{{{14.08}_{\pm 0.02}}}$          & ${{50.03}_{\pm 0.04}}$                  & ${{37.91}_{\pm 0.17}}$             & ${{28.33}_{\pm 0.26}}$             & ${{30.21}_{\pm 0.04}}$             & $\underline{{{27.29}_{\pm 0.03}}}$ \\
            UnInfo (ours)                                  & $\mathbf{{{31.51}_{\pm 0.19}}}$            & ${{16.76}_{\pm 1.62}}$                   & ${{23.47}_{\pm 0.74}}$                    & ${{20.40}_{\pm 0.80}}$                  & ${{22.81}_{\pm 0.27}}$                & ${{16.59}_{\pm 0.40}}$                          & $\underline{{{42.03}_{\pm 0.23}}}$             & $\underline{{{42.38}_{\pm 0.26}}}$    & ${{7.56}_{\pm 3.50}}$                        & $\underline{{{10.60}_{\pm 0.41}}}$          & $\underline{{{11.36}_{\pm 0.86}}}$       & $\underline{{{57.75}_{\pm 0.11}}}$      & $\underline{{{39.16}_{\pm 0.34}}}$ & $\mathbf{{{31.65}_{\pm 0.48}}}$    & $\underline{{{32.40}_{\pm 0.13}}}$ & ${{27.10}_{\pm 0.12}}$             \\ \bottomrule
        \end{tabular}
    }
\end{table}

\begin{table}[tb]
    \centering
    \caption{\highlight{Test accuracy~(\%) on ImageNet-C-bar. The numbers (1), (5), and (10) presented with the method names are the shot numbers per class $n$ used for the few-shot adaptation methods.}}
    \label{tab:imagenet-c-bar_accuracy}
    \resizebox{1\linewidth}{!}{
        \begin{tabular}{llllllllllll}\toprule
            Method                                         & \rotatebox{60}{\shortstack{Blue\\noise\\sample}} & \rotatebox{60}{\shortstack{Brownish\\noise}} & \rotatebox{60}{\shortstack{Caustic\\refraction}} & \rotatebox{60}{\shortstack{Checkerboard\\cutout}} & \rotatebox{60}{\shortstack{Cocentric\\sine\\waves}} & \rotatebox{60}{\shortstack{Inverse\\sparkles}} & \rotatebox{60}{\shortstack{Perlin\\noise}} & \rotatebox{60}{\shortstack{Plasma\\noise}} & \rotatebox{60}{\shortstack{Single\\frequency\\greyscale}} & \rotatebox{60}{\shortstack{Sparkles}} & Mean                               \\ \midrule
            No-adapt                                       & ${21.87}$                                        & ${46.66}$                                    & ${39.55}$                                        & ${45.51}$                                         & ${10.05}$                                           & ${19.57}$                                      & ${51.21}$                                  & ${20.69}$                                  & ${16.10}$                                                 & ${50.00}$                             & ${32.12}$                          \\ \midrule
            Linear probing (1)                             & ${{9.05}_{\pm 0.08}}$                            & ${{24.22}_{\pm 0.39}}$                       & ${{17.77}_{\pm 0.17}}$                           & ${{21.99}_{\pm 0.39}}$                            & ${{3.74}_{\pm 0.05}}$                               & ${{8.59}_{\pm 0.22}}$                          & ${{24.87}_{\pm 0.28}}$                     & ${{10.04}_{\pm 0.30}}$                     & ${{4.23}_{\pm 0.11}}$                                     & ${{26.19}_{\pm 0.45}}$                & ${{15.07}_{\pm 0.08}}$             \\
            Linear probing (5)                             & ${{15.45}_{\pm 0.17}}$                           & ${{38.05}_{\pm 0.34}}$                       & ${{29.12}_{\pm 0.08}}$                           & ${{35.58}_{\pm 0.26}}$                            & ${{6.89}_{\pm 0.19}}$                               & ${{14.28}_{\pm 0.48}}$                         & ${{39.24}_{\pm 0.33}}$                     & ${{15.43}_{\pm 0.22}}$                     & ${{6.82}_{\pm 0.17}}$                                     & ${{40.55}_{\pm 0.17}}$                & ${{24.14}_{\pm 0.08}}$             \\
            Linear probing (10)                            & ${{18.58}_{\pm 0.08}}$                           & ${{44.03}_{\pm 0.18}}$                       & ${{34.96}_{\pm 0.36}}$                           & ${{41.54}_{\pm 0.03}}$                            & ${{8.83}_{\pm 0.12}}$                               & ${{17.36}_{\pm 0.25}}$                         & ${{45.82}_{\pm 0.19}}$                     & ${{18.41}_{\pm 0.14}}$                     & ${{9.05}_{\pm 0.18}}$                                     & ${{47.20}_{\pm 0.18}}$                & ${{28.58}_{\pm 0.02}}$             \\
            Tip-adapter~\citep{Zhang2022tip-adapter} (1)   & ${{14.95}_{\pm 0.19}}$                           & ${{37.35}_{\pm 0.17}}$                       & ${{29.99}_{\pm 0.15}}$                           & ${{35.58}_{\pm 0.31}}$                            & ${{7.12}_{\pm 0.12}}$                               & ${{14.82}_{\pm 0.10}}$                         & ${{40.06}_{\pm 0.14}}$                     & ${{15.46}_{\pm 0.11}}$                     & ${{10.06}_{\pm 0.05}}$                                    & ${{41.29}_{\pm 0.18}}$                & ${{24.67}_{\pm 0.05}}$             \\
            Tip-adapter (5)                                & ${{18.41}_{\pm 0.08}}$                           & ${{43.08}_{\pm 0.26}}$                       & ${{34.86}_{\pm 0.27}}$                           & ${{40.79}_{\pm 0.09}}$                            & ${{8.37}_{\pm 0.14}}$                               & ${{17.58}_{\pm 0.15}}$                         & ${{45.58}_{\pm 0.12}}$                     & ${{18.15}_{\pm 0.09}}$                     & ${{10.24}_{\pm 0.16}}$                                    & ${{46.44}_{\pm 0.13}}$                & ${{28.35}_{\pm 0.08}}$             \\
            Tip-adapter (10)                               & ${{20.22}_{\pm 0.38}}$                           & ${{45.93}_{\pm 0.20}}$                       & ${{37.59}_{\pm 0.19}}$                           & ${{43.65}_{\pm 0.14}}$                            & ${{9.77}_{\pm 0.08}}$                               & ${{19.62}_{\pm 0.25}}$                         & ${{48.52}_{\pm 0.26}}$                     & ${{20.07}_{\pm 0.05}}$                     & ${{11.51}_{\pm 0.07}}$                                    & ${{49.26}_{\pm 0.23}}$                & ${{30.61}_{\pm 0.08}}$             \\ \midrule
            TPT~\citep{shu2022tpt}                         & ${{24.63}_{\pm 0.07}}$                           & ${{49.86}_{\pm 0.04}}$                       & ${{42.53}_{\pm 0.12}}$                           & ${{46.36}_{\pm 0.01}}$                            & ${{10.96}_{\pm 0.07}}$                              & ${{21.93}_{\pm 0.03}}$                         & ${{54.31}_{\pm 0.03}}$                     & ${{23.16}_{\pm 0.03}}$                     & ${{17.43}_{\pm 0.08}}$                                    & ${{52.20}_{\pm 0.02}}$                & ${{34.34}_{\pm 0.01}}$             \\
            ZERO~\citep{farina2024frustratingly}           & ${{25.65}_{\pm 0.06}}$                           & ${{45.60}_{\pm 0.01}}$                       & ${{41.14}_{\pm 0.04}}$                           & ${{45.30}_{\pm 0.04}}$                            & ${{10.59}_{\pm 0.02}}$                              & ${{22.79}_{\pm 0.02}}$                         & ${{49.76}_{\pm 0.02}}$                     & ${{20.42}_{\pm 0.06}}$                     & $\underline{{{19.81}_{\pm 0.07}}}$                        & ${{45.81}_{\pm 0.10}}$                & ${{32.69}_{\pm 0.02}}$             \\
            MTA~\citep{zanella2024test}                    & ${{23.75}_{\pm 0.02}}$                           & ${{45.13}_{\pm 0.03}}$                       & ${{40.28}_{\pm 0.12}}$                           & ${{45.05}_{\pm 0.08}}$                            & ${{9.93}_{\pm 0.01}}$                               & ${{20.53}_{\pm 0.04}}$                         & ${{50.50}_{\pm 0.01}}$                     & ${{20.04}_{\pm 0.02}}$                     & ${{19.32}_{\pm 0.06}}$                                    & ${{45.89}_{\pm 0.02}}$                & ${{32.04}_{\pm 0.02}}$             \\
            TDA~\citep{karmanov2024efficient}              & ${{24.53}_{\pm 0.04}}$                           & ${{50.30}_{\pm 0.02}}$                       & $\underline{{{42.90}_{\pm 0.02}}}$               & ${{49.85}_{\pm 0.02}}$                            & ${{12.69}_{\pm 0.03}}$                              & ${{22.56}_{\pm 0.04}}$                         & ${{53.89}_{\pm 0.09}}$                     & ${{24.75}_{\pm 0.09}}$                     & ${{17.77}_{\pm 0.01}}$                                    & $\underline{{{55.00}_{\pm 0.10}}}$    & ${{35.42}_{\pm 0.01}}$             \\
            \highlight{DMN~\citep{zhang2024dual}}          & ${{23.21}_{\pm 0.00}}$                           & ${{50.42}_{\pm 0.00}}$                       & ${{41.37}_{\pm 0.00}}$                           & ${{47.08}_{\pm 0.00}}$                            & ${{10.95}_{\pm 0.00}}$                              & ${{20.98}_{\pm 0.00}}$                         & ${{54.15}_{\pm 0.00}}$                     & ${{22.95}_{\pm 0.00}}$                     & ${{16.91}_{\pm 0.00}}$                                    & ${{53.09}_{\pm 0.00}}$                & ${{34.11}_{\pm 0.00}}$             \\
            \highlight{Mint~\citep{bao2026mint}}           & $\mathbf{{{29.22}_{\pm 0.07}}}$                  & $\mathbf{{{53.98}_{\pm 0.07}}}$              & ${{39.49}_{\pm 0.17}}$                           & $\mathbf{{{53.12}_{\pm 0.05}}}$                   & $\underline{{{15.48}_{\pm 0.05}}}$                  & $\mathbf{{{24.31}_{\pm 0.25}}}$                & $\underline{{{54.87}_{\pm 0.06}}}$         & $\mathbf{{{32.48}_{\pm 0.05}}}$            & $\mathbf{{{20.13}_{\pm 0.07}}}$                           & $\mathbf{{{57.56}_{\pm 0.06}}}$       & $\mathbf{{{38.06}_{\pm 0.04}}}$    \\
            \highlight{BATCLIP~\citep{Maharana_2025_ICCV}} & $\underline{{{26.92}_{\pm 0.17}}}$               & ${{47.44}_{\pm 0.25}}$                       & ${{36.60}_{\pm 0.25}}$                           & ${{46.63}_{\pm 0.10}}$                            & $\mathbf{{{16.28}_{\pm 0.15}}}$                     & ${{23.52}_{\pm 0.09}}$                         & ${{50.21}_{\pm 0.12}}$                     & $\underline{{{28.91}_{\pm 0.11}}}$         & ${{18.91}_{\pm 0.10}}$                                    & ${{50.69}_{\pm 0.20}}$                & ${{34.61}_{\pm 0.11}}$             \\
            UnInfo (ours)                                  & ${{26.78}_{\pm 0.64}}$                           & $\underline{{{51.21}_{\pm 0.32}}}$           & $\mathbf{{{43.73}_{\pm 0.40}}}$                  & $\underline{{{50.03}_{\pm 1.89}}}$                & ${{12.49}_{\pm 0.28}}$                              & $\underline{{{23.58}_{\pm 0.69}}}$             & $\mathbf{{{55.22}_{\pm 0.23}}}$            & ${{24.88}_{\pm 0.22}}$                     & ${{19.67}_{\pm 0.35}}$                                    & ${{53.74}_{\pm 0.71}}$                & $\underline{{{36.13}_{\pm 0.16}}}$ \\ \bottomrule
        \end{tabular}
    }
\end{table}

We evaluated each corruption type's test classification accuracy on ImageNet-C and C-bar.
We ran TTA three times with different random seeds for each method and corruption type, and report the mean score.
\highlight{We adopt a batched online evaluation protocol, in which model updates and batch evaluations are performed alternately (i.e., each sample is accessed once).}
\cref{tab:imagenet-c_accuracy,tab:imagenet-c-bar_accuracy} show the results.
\highlight{Our UnInfo consistently improved accuracy compared to No-adapt regardless of corruption types.}
Intriguingly, the baselines, even the few-shot methods that use labeled test samples for adaptation, sometimes underperformed No-adapt.
This is because the baseline methods aim to refine text and/or image embeddings in a post-hoc manner during testing.
While corrupted images are not appropriately encoded, as discussed in \cref{sec:preliminary_experiment}, the encoders themselves remain fixed.
In contrast, UnInfo successfully adapts to image corruption by updating the image encoder with LoRA.
\highlight{Although recent CLIP-TTA methods, Mint~\citep{bao2026mint} and BATCLIP~\citep{Maharana_2025_ICCV}, outperformed UnInfo on average in \cref{tab:imagenet-c_accuracy,tab:imagenet-c-bar_accuracy}, they performed worse than No-adapt on some corruption types and model architectures (\cref{tab:openai-vit-b-32_imagenet-c_accuracy,tab:openai-vit-b-32_imagenet-c-bar_accuracy,tab:siglip_imagenet-c_accuracy,tab:siglip_imagenet-c-bar_accuracy}).
    These results indicate that UnInfo consistently works across corruption types and model architectures under the evaluated setting.}

\highlight{We also evaluated the performance on clean data and low-severity corruption.
    In real-world applications, whether image corruption occurs cannot be known in advance.
    To examine whether UnInfo negatively affects the performance on clean or low-severity corruption, we changed the severity from 0 (clean) to 5.
    \cref{tab:imagenet-c_severity} shows the result.
    Although the accuracy gain is greater at higher severities, UnInfo showed consistent improvement across all severities, including clean images, suggesting that UnInfo does not negatively affect low-severity images.}

\begin{table}[tb]
    \centering
    \caption{\highlight{Average test accuracy (\%) of ViT-B/16 CLIP on ImageNet-C across severity.}}
    \label{tab:imagenet-c_severity}
    \resizebox{0.8\linewidth}{!}{
        \begin{tabular}{llllllll}\toprule
            ~             & \multicolumn{6}{c}{Severity}                                                                                                                                                       \\ \cmidrule(lr){2-7}
            Method        & 0 (clean)                    & 1                      & 2                      & 3                      & 4                      & 5                      & Mean                   \\ \midrule
            No-adapt      & ${66.97}$                    & ${56.09}$              & ${48.54}$              & ${42.14}$              & ${32.85}$              & ${23.06}$              & ${44.94}$              \\
            UnInfo (ours) & ${{67.82}_{\pm 0.05}}$       & ${{58.83}_{\pm 0.08}}$ & ${{52.13}_{\pm 0.01}}$ & ${{46.64}_{\pm 0.08}}$ & ${{37.81}_{\pm 0.03}}$ & ${{27.10}_{\pm 0.30}}$ & ${{48.39}_{\pm 0.04}}$ \\ \bottomrule
        \end{tabular}
    }
\end{table}

\subsubsection{Ablation Study}\label{sssec:exp_ablation}
\begin{table}[tb]
    \centering
    \caption{\highlight{Ablation of UnInfo on ImageNet-C.}}\label{tab:ablation_imagenet-c}
    \resizebox{1\linewidth}{!}{
        \setlength{\tabcolsep}{3pt}
        \begin{tabular}{lllllllllllllllll} \toprule
            Method                                                                                        & \rotatebox{60}{\shortstack{Defocus\\blur}} & \rotatebox{60}{\shortstack{Glass\\blur}} & \rotatebox{60}{\shortstack{Motion\\blur}} & \rotatebox{60}{\shortstack{Zoom\\blur}} & \rotatebox{60}{\shortstack{Contrast}} & \rotatebox{60}{\shortstack{Elastic\\transform}} & \rotatebox{60}{\shortstack{Jpeg\\compression}} & \rotatebox{60}{\shortstack{Pixelate}} & \rotatebox{60}{\shortstack{Gaussian\\noise}} & \rotatebox{60}{\shortstack{Impulse\\noise}} & \rotatebox{60}{\shortstack{Shot\\noise}} & \rotatebox{60}{\shortstack{Brightness}} & \rotatebox{60}{\shortstack{Fog}}   & \rotatebox{60}{\shortstack{Frost}} & \rotatebox{60}{\shortstack{Snow}}  & Mean                               \\ \midrule
            $\mathcal{L}_\text{ent}$                                                                      & ${{0.21}_{\pm 0.01}}$                      & ${{0.15}_{\pm 0.01}}$                    & ${{0.15}_{\pm 0.01}}$                     & ${{0.17}_{\pm 0.01}}$                   & ${{0.19}_{\pm 0.02}}$                 & ${{0.14}_{\pm 0.00}}$                           & ${{0.32}_{\pm 0.01}}$                          & ${{0.30}_{\pm 0.01}}$                 & ${{0.11}_{\pm 0.00}}$                        & ${{0.11}_{\pm 0.01}}$                       & ${{0.11}_{\pm 0.00}}$                    & ${{0.70}_{\pm 0.08}}$                   & ${{0.24}_{\pm 0.01}}$              & ${{0.19}_{\pm 0.03}}$              & ${{0.22}_{\pm 0.02}}$              & ${{0.22}_{\pm 0.01}}$              \\
            $\mathcal{L}_\text{ent} + \mathcal{L}_\text{pl}$                                              & ${{30.63}_{\pm 0.18}}$                     & ${{15.56}_{\pm 1.78}}$                   & ${{22.61}_{\pm 0.77}}$                    & ${{19.54}_{\pm 0.26}}$                  & ${{21.83}_{\pm 0.18}}$                & ${{15.99}_{\pm 0.15}}$                          & ${{41.03}_{\pm 0.18}}$                         & ${{40.82}_{\pm 0.47}}$                & ${{3.31}_{\pm 3.08}}$                        & ${{6.68}_{\pm 0.58}}$                       & ${{8.06}_{\pm 0.56}}$                    & ${{57.17}_{\pm 0.09}}$                  & ${{37.78}_{\pm 0.28}}$             & ${{30.84}_{\pm 0.18}}$             & ${{31.29}_{\pm 0.29}}$             & ${{25.54}_{\pm 0.35}}$             \\
            \highlight{$\mathcal{L}_\text{mi}$}                                                           & ${{0.66}_{\pm 0.08}}$                      & ${{0.70}_{\pm 0.07}}$                    & ${{0.66}_{\pm 0.05}}$                     & ${{1.01}_{\pm 0.09}}$                   & ${{0.58}_{\pm 0.01}}$                 & ${{1.02}_{\pm 0.16}}$                           & ${{1.26}_{\pm 0.10}}$                          & ${{1.35}_{\pm 0.15}}$                 & ${{0.22}_{\pm 0.04}}$                        & ${{0.19}_{\pm 0.04}}$                       & ${{0.21}_{\pm 0.05}}$                    & ${{2.46}_{\pm 0.30}}$                   & ${{1.73}_{\pm 0.13}}$              & ${{1.01}_{\pm 0.14}}$              & ${{0.98}_{\pm 0.13}}$              & ${{0.94}_{\pm 0.02}}$              \\
            \highlight{$\mathcal{L}_\text{mi} + \mathcal{L}_\text{pl}$}                                   & ${{30.29}_{\pm 0.25}}$                     & $\mathbf{{{17.77}_{\pm 1.07}}}$          & $\mathbf{{{23.75}_{\pm 0.49}}}$           & $\mathbf{{{21.35}_{\pm 0.36}}}$         & ${{22.21}_{\pm 0.04}}$                & $\mathbf{{{17.61}_{\pm 0.91}}}$                 & ${{40.60}_{\pm 0.19}}$                         & ${{41.08}_{\pm 0.34}}$                & $\underline{{{7.26}_{\pm 4.37}}}$            & $\mathbf{{{11.19}_{\pm 0.21}}}$             & $\mathbf{{{12.29}_{\pm 0.33}}}$          & ${{56.52}_{\pm 0.23}}$                  & ${{37.80}_{\pm 0.54}}$             & ${{30.64}_{\pm 0.61}}$             & $\underline{{{32.16}_{\pm 0.03}}}$ & $\underline{{{26.84}_{\pm 0.27}}}$ \\
            $\mathcal{L}_\text{ent} + \mathcal{L}_\text{unif} + \mathcal{L}_\text{pl}$                    & $\underline{{{31.08}_{\pm 0.07}}}$         & ${{16.48}_{\pm 1.62}}$                   & ${{23.40}_{\pm 0.48}}$                    & ${{20.32}_{\pm 0.42}}$                  & $\underline{{{22.49}_{\pm 0.12}}}$    & $\underline{{{16.67}_{\pm 0.23}}}$              & $\underline{{{41.26}_{\pm 0.35}}}$             & $\underline{{{41.26}_{\pm 0.45}}}$    & ${{3.90}_{\pm 3.70}}$                        & ${{7.87}_{\pm 0.59}}$                       & ${{7.17}_{\pm 3.64}}$                    & $\underline{{{57.24}_{\pm 0.04}}}$      & $\underline{{{38.74}_{\pm 0.20}}}$ & $\underline{{{31.24}_{\pm 0.21}}}$ & ${{32.01}_{\pm 0.42}}$             & ${{26.07}_{\pm 0.26}}$             \\
            $\mathcal{L}_\text{ent} + \mathcal{L}_\text{unif} + \mathcal{L}_\text{pl}+$Balancing (UnInfo) & $\mathbf{{{31.51}_{\pm 0.19}}}$            & $\underline{{{16.76}_{\pm 1.62}}}$       & $\underline{{{23.47}_{\pm 0.74}}}$        & $\underline{{{20.40}_{\pm 0.80}}}$      & $\mathbf{{{22.81}_{\pm 0.27}}}$       & ${{16.59}_{\pm 0.40}}$                          & $\mathbf{{{42.03}_{\pm 0.23}}}$                & $\mathbf{{{42.38}_{\pm 0.26}}}$       & $\mathbf{{{7.56}_{\pm 3.50}}}$               & $\underline{{{10.60}_{\pm 0.41}}}$          & $\underline{{{11.36}_{\pm 0.86}}}$       & $\mathbf{{{57.75}_{\pm 0.11}}}$         & $\mathbf{{{39.16}_{\pm 0.34}}}$    & $\mathbf{{{31.65}_{\pm 0.48}}}$    & $\mathbf{{{32.40}_{\pm 0.13}}}$    & $\mathbf{{{27.10}_{\pm 0.12}}}$    \\ \bottomrule
        \end{tabular}
    }
\end{table}

\begin{table}[tb]
    \centering
    \caption{\highlight{Ablation of UnInfo on ImageNet-C-bar.}}\label{tab:ablation_imagenet-c-bar}
    \resizebox{1\linewidth}{!}{
        \begin{tabular}{llllllllllll}\toprule
            Method                                                                                        & \rotatebox{60}{\shortstack{Blue\\noise\\sample}} & \rotatebox{60}{\shortstack{Brownish\\noise}} & \rotatebox{60}{\shortstack{Caustic\\refraction}} & \rotatebox{60}{\shortstack{Checkerboard\\cutout}} & \rotatebox{60}{\shortstack{Cocentric\\sine\\waves}} & \rotatebox{60}{\shortstack{Inverse\\sparkles}} & \rotatebox{60}{\shortstack{Perlin\\noise}} & \rotatebox{60}{\shortstack{Plasma\\noise}} & \rotatebox{60}{\shortstack{Single\\frequency\\greyscale}} & \rotatebox{60}{\shortstack{Sparkles}} & Mean                               \\ \midrule
            $\mathcal{L}_\text{ent}$                                                                      & ${{0.18}_{\pm 0.02}}$                            & ${{0.44}_{\pm 0.07}}$                        & ${{0.21}_{\pm 0.01}}$                            & ${{0.58}_{\pm 0.16}}$                             & ${{0.12}_{\pm 0.01}}$                               & ${{0.13}_{\pm 0.00}}$                          & ${{0.52}_{\pm 0.07}}$                      & ${{0.15}_{\pm 0.01}}$                      & ${{0.14}_{\pm 0.01}}$                                     & ${{0.41}_{\pm 0.20}}$                 & ${{0.29}_{\pm 0.01}}$              \\
            $\mathcal{L}_\text{ent} + \mathcal{L}_\text{pl}$                                              & ${{26.20}_{\pm 0.37}}$                           & ${{49.67}_{\pm 0.17}}$                       & ${{42.81}_{\pm 0.45}}$                           & ${{48.87}_{\pm 1.72}}$                            & ${{11.67}_{\pm 0.34}}$                              & ${{22.78}_{\pm 0.63}}$                         & ${{53.94}_{\pm 0.28}}$                     & ${{23.85}_{\pm 0.55}}$                     & ${{19.48}_{\pm 0.24}}$                                    & ${{49.63}_{\pm 3.89}}$                & ${{34.89}_{\pm 0.23}}$             \\
            \highlight{$\mathcal{L}_\text{mi}$}                                                           & ${{0.57}_{\pm 0.10}}$                            & ${{2.32}_{\pm 0.28}}$                        & ${{1.56}_{\pm 0.26}}$                            & ${{2.49}_{\pm 0.17}}$                             & ${{0.48}_{\pm 0.05}}$                               & ${{0.90}_{\pm 0.03}}$                          & ${{2.65}_{\pm 0.19}}$                      & ${{0.87}_{\pm 0.15}}$                      & ${{0.47}_{\pm 0.03}}$                                     & ${{3.14}_{\pm 0.10}}$                 & ${{1.54}_{\pm 0.03}}$              \\
            \highlight{$\mathcal{L}_\text{mi}+\mathcal{L}_\text{pl}$}                                     & ${{26.45}_{\pm 1.00}}$                           & ${{48.96}_{\pm 0.53}}$                       & ${{42.69}_{\pm 0.16}}$                           & $\mathbf{{{51.20}_{\pm 0.21}}}$                   & $\mathbf{{{13.52}_{\pm 0.47}}}$                     & ${{22.78}_{\pm 0.32}}$                         & ${{53.78}_{\pm 0.21}}$                     & $\underline{{{24.41}_{\pm 0.39}}}$         & ${{19.43}_{\pm 0.11}}$                                    & $\mathbf{{{55.34}_{\pm 0.18}}}$       & $\underline{{{35.86}_{\pm 0.21}}}$ \\
            $\mathcal{L}_\text{ent} + \mathcal{L}_\text{unif} + \mathcal{L}_\text{pl}$                    & $\underline{{{26.75}_{\pm 0.34}}}$               & $\underline{{{50.28}_{\pm 0.39}}}$           & $\underline{{{43.14}_{\pm 0.04}}}$               & ${{49.59}_{\pm 0.91}}$                            & ${{12.08}_{\pm 0.23}}$                              & $\underline{{{23.36}_{\pm 0.56}}}$             & $\underline{{{54.47}_{\pm 0.36}}}$         & ${{24.27}_{\pm 0.41}}$                     & $\mathbf{{{19.69}_{\pm 0.20}}}$                           & $\underline{{{54.27}_{\pm 1.50}}}$    & ${{35.79}_{\pm 0.21}}$             \\
            $\mathcal{L}_\text{ent} + \mathcal{L}_\text{unif} + \mathcal{L}_\text{pl}+$Balancing (UnInfo) & $\mathbf{{{26.78}_{\pm 0.64}}}$                  & $\mathbf{{{51.21}_{\pm 0.32}}}$              & $\mathbf{{{43.73}_{\pm 0.40}}}$                  & $\underline{{{50.03}_{\pm 1.89}}}$                & $\underline{{{12.49}_{\pm 0.28}}}$                  & $\mathbf{{{23.58}_{\pm 0.69}}}$                & $\mathbf{{{55.22}_{\pm 0.23}}}$            & $\mathbf{{{24.88}_{\pm 0.22}}}$            & $\underline{{{19.67}_{\pm 0.35}}}$                        & ${{53.74}_{\pm 0.71}}$                & $\mathbf{{{36.13}_{\pm 0.16}}}$    \\ \bottomrule
        \end{tabular}
    }
\end{table}

Here, we examined the effect of each component in UnInfo: uniformity-aware confidence maximization, information-aware loss balancing, and knowledge distillation from the EMA teacher.
\cref{tab:ablation_imagenet-c,tab:ablation_imagenet-c-bar} show the results.
Solely minimizing the entropy loss $\mathcal{L}_\text{ent}$ resulted in catastrophically poor accuracy in all cases.
This is because image corruption affects uniformity, as observed in \cref{sec:preliminary_experiment}; entropy can assign unreasonably high confidence to wrong classes, resulting in overfitting quickly.
In contrast, incorporating the knowledge distillation loss $\mathcal{L}_\text{pl}$ drastically improved the stability.
The uniformity loss $\mathcal{L}_\text{unif}$ further improved accuracy compared to $\mathcal{L}_\text{ent}+\mathcal{L}_\text{pl}$.
However, its improvement is sometimes marginal because it overlooks the balance between entropy and uniformity, as \cref{ssec:proposed_method_balancing} describes.
Thus, adding the balancing further improved accuracy by properly enhancing entropy or uniformity.
Specifically, we observed significant improvements in difficult corruption types that produce high uniformity loss, such as noise corruption.
In such cases, uniformity should be recovered first before minimizing entropy loss.
UnInfo successfully controls the priority of the losses.

\highlight{On the other hand, instead of uniformity and balancing, one may directly use mutual information in \cref{eq:prediction_mutual_information} as a loss function $\mathcal{L}_\text{mi}=-\mathcal{I}(\mathbf{z};\hat{y})$ that can enhance the embeddings' diversity while refining the prediction confidence.
    However, solely minimizing $\mathcal{L}_\text{mi}$ resulted in collapse because it is optimal for $\mathcal{L}_\text{mi}$ to make each image embedding in a batch identical to a text embedding, regardless of ground truth, making prediction diverse and confident, but suboptimal for the image embeddings' diversity.
    Combining $\mathcal{L}_\text{mi}$ with the EMA teacher ($\mathcal{L}_\text{mi}+\mathcal{L}_\text{pl}$) substantially alleviated this issue and achieved performance comparable to UnInfo across most corruption types, while outperforming UnInfo on several corruption types.
    These results suggest that $\mathcal{L}_\text{mi}$ is also an effective objective when appropriately regularized, whereas the improvements of UnInfo are likely due to the combined effects of uniformity, adaptive balancing, and pseudo-label learning rather than the uniformity objective alone.}

\subsubsection{Sensitivity Analysis}\label{sssec:exp_sensitivity_analysis}

\begin{figure}
    \centering
    \resizebox{0.7\linewidth}{!}{
        \begin{tabular}{cc}
            \includegraphics[alt={Sensitivity analysis of the uniformity loss weight.},height=4cm]{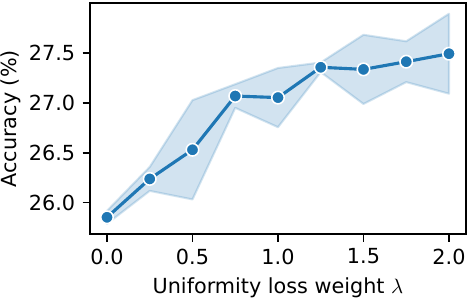}
             &
            \includegraphics[alt={Sensitivity analysis of the mutual information threshold used in the loss balancing.},height=4cm]{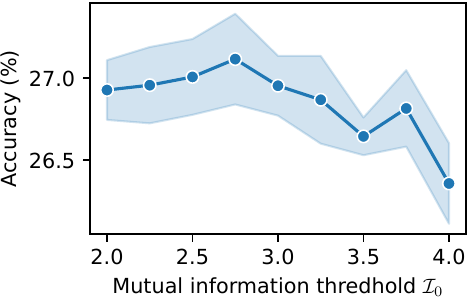}
        \end{tabular}
    }
    \caption{Sensitivity analysis of the uniformity loss weight $\lambda$ (left), and the mutual information threshold $\mathcal{I}_0$ used in the loss balancing (right).
        The mean and standard deviation of the test accuracy calculated over the ImageNet-C corruptions are plotted.}
    \label{fig:sensitivity_analysis}
\end{figure}

\begin{figure}[tb]
    \centering
    \includegraphics[alt={Sensitivity analysis of the balancing weight with being fixed.},width=0.7\linewidth]{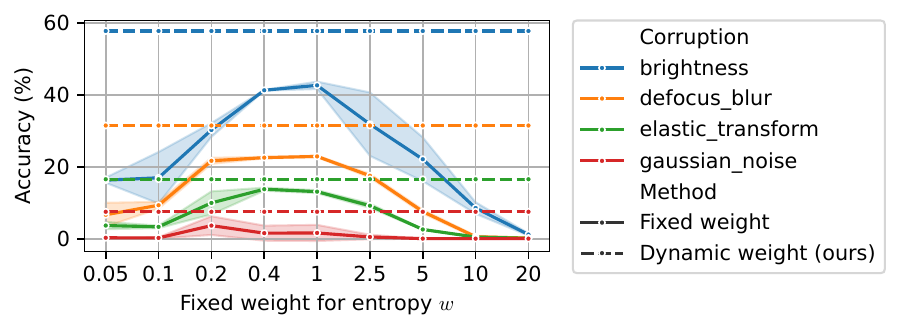}
    \caption{Sensitivity analysis of the balancing weight $w$ with being fixed. The dashed lines represent accuracies when $w$ is dynamically updated with our balancing mechanism described in \cref{ssec:proposed_method_balancing}.}
    \label{fig:sensitivity_fixed_weight}
\end{figure}

\cref{fig:sensitivity_analysis} plots the sensitivity analysis of the hyperparameters of UnInfo.
We changed the weight of the uniformity loss $\lambda$ in \cref{eq:proposed_balaning_loss} and the threshold of the mutual information $\mathcal{I}_0$ for the loss balancing.
We ran UnInfo on the ImageNet-C corruptions and reported the average test accuracy.
Increasing $\lambda$ produces better accuracies, mainly in $0.0\leq \lambda \leq 0.75$, suggesting the efficacy of the uniformity loss, and further increasing $\lambda$ beyond $1.0$ results in slightly higher accuracy.
On the other hand, varying $\mathcal{I}_0$ hits the best accuracy when $\mathcal{I}_0=2.75$ but slightly affects the accuracy within $2.0\leq \mathcal{I}_0\leq 3.25$.
However, increasing $\mathcal{I}_0$ too much deteriorates the accuracy because it controls the bias of the balance between entropy and uniformity.
When $\mathcal{I}_0$ is too high, the uniformity loss is constantly overweighted, and the entropy is no longer optimized.

For checking the effectiveness of the dynamic weight by information-aware loss balancing, we fixed the weight $w$ in \cref{eq:proposed_balaning_loss}.
\cref{fig:sensitivity_fixed_weight} shows the accuracy with fixed value of $w$ being changed.
Although leveraging both entropy and uniformity ($0.2 \leq w\leq 1$) produces higher accuracy regardless of corruption types, dynamic $w$ had significant improvements compared to the best accuracies of fixed $w$.

\subsubsection{Qualitative Analysis}\label{sssec:exp_qualitative_analysis}

\begin{figure}[tb]
    \centering
    \resizebox{1.0\linewidth}{!}{
        \begin{tabular}{cccc}
            \includegraphics[alt={Weight evolution of the information-aware loss balancing on brightness corruption.},height=2cm]{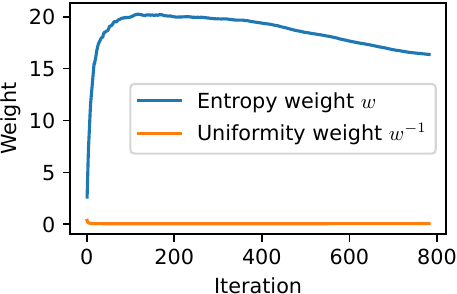}
                                         &
            \includegraphics[alt={Weight evolution of the information-aware loss balancing on defocus blur corruption.},height=2cm]{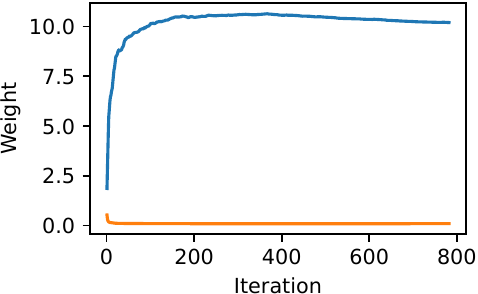}
                                         &
            \includegraphics[alt={Weight evolution of the information-aware loss balancing on elastic transform corruption.},height=2cm]{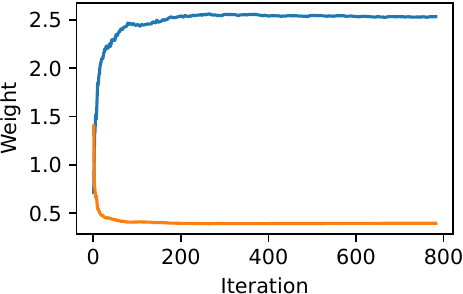}
                                         &
            \includegraphics[alt={Weight evolution of the information-aware loss balancing on Gaussian noise corruption.},height=2cm]{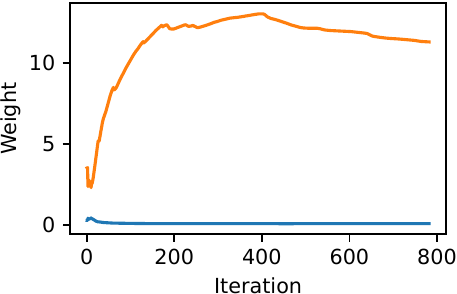}
            \\
            {\scriptsize (a) Brightness} & {\scriptsize (b) Defocus blur} & { \scriptsize(c) Elastic transform} & {\scriptsize (d) Gaussian noise}
        \end{tabular}
    }
    \caption{Evolution of the information-aware loss balancing weights.
        The weights are adaptively assigned to the entropy and uniformity losses by the difficulty of distribution shifts.
        A larger weight is assigned to the entropy for easy distribution shifts such as brightness in (a).
        In contrast, the uniformity loss is prioritized for difficult distribution shifts such as Gaussian noise in (d).}
    \label{fig:weight_evolution}
\end{figure}

\begin{figure}
    \centering
    \resizebox{0.8\linewidth}{!}{
        \begin{tabular}{cccc}
            \includegraphics[width=0.25\linewidth,alt={Visualization of image embeddings on the brightness corruption before adaptation}]{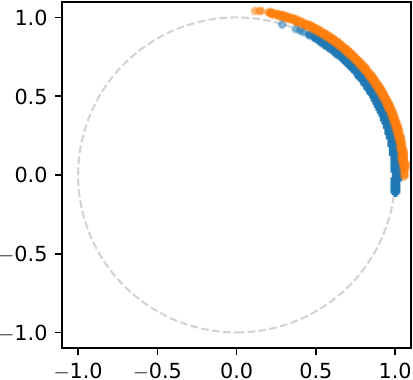}
                                                      &
            \includegraphics[width=0.25\linewidth,alt={Visualization of image embeddings on the brightness corruption after adaptation}]{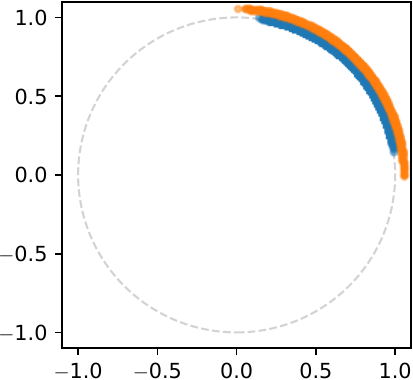}
                                                      &
            \includegraphics[width=0.25\linewidth,alt={Visualization of image embeddings on the defocus blur corruption before adaptation}]{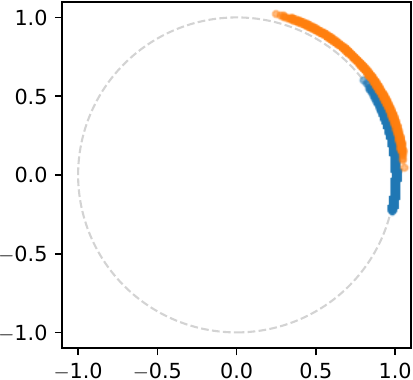}
                                                      &
            \includegraphics[width=0.25\linewidth,alt={Visualization of image embeddings on the defocus blur corruption after adaptation}]{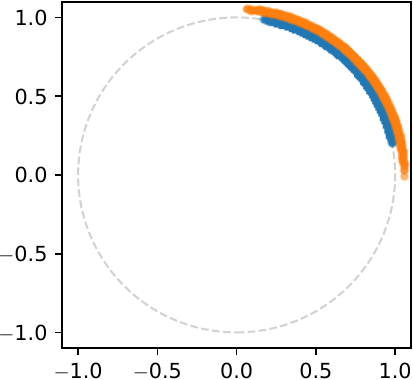}
            \\
            \multicolumn{2}{c}{(a) Brightness}        & \multicolumn{2}{c}{(b) Defocus blur}
            \\
            %============
            \includegraphics[width=0.25\linewidth,alt={Visualization of image embeddings on the elastic transform corruption before adaptation}]{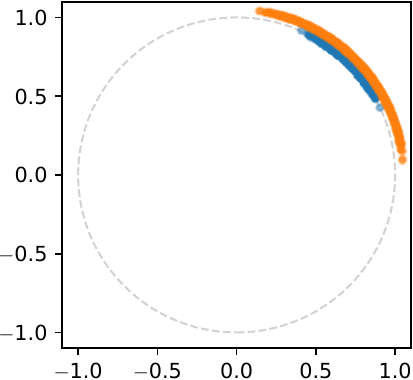}
                                                      &
            \includegraphics[width=0.25\linewidth,alt={Visualization of image embeddings on the elastic transform corruption after adaptation}]{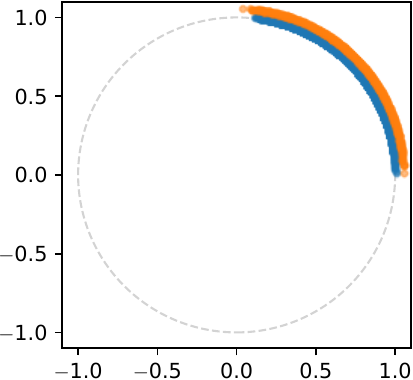}
                                                      &
            \includegraphics[width=0.25\linewidth,alt={Visualization of image embeddings on the Gaussian noise corruption before adaptation}]{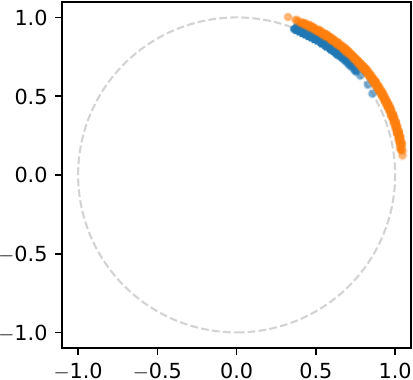}
                                                      &
            \includegraphics[width=0.25\linewidth,alt={Visualization of image embeddings on the Gaussian noise corruption after adaptation}]{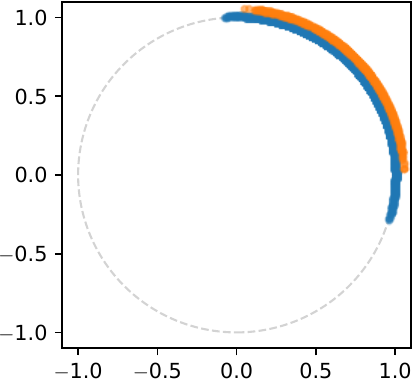}
            \\
            Before TTA                                & After TTA                              & Before TTA & After TTA
            \\
            \multicolumn{2}{c}{(c) Elastic transform} & \multicolumn{2}{c}{(d) Gaussian noise}
        \end{tabular}
    }
    \caption{Spherical PCA~\citep{liu2019spherical} visualization of image~(\textcolor{tab10blue}{blue dots}) and text~(\textcolor{tab10orange}{orange dots}) embeddings before and after TTA with UnInfo on ImageNet-C.
        The image embeddings are distributed in a broader range of the circle after TTA, which suggests that the uniformity is improved. Moreover, the distribution of the image embeddings is aligned with the text embeddings, i.e., the image embeddings are more classification-friendly.}
    \label{fig:spca_visualization}
\end{figure}

\cref{fig:weight_evolution} plots the evolution of the dynamic weights $w$ and $w^{-1}$ in \cref{eq:proposed_balaning_loss}.
For easy image corruption types on which No-adapt produced relatively high accuracy, such as brightness in (a), the weight for the entropy $w$ quickly increased, and the weight for uniformity loss $w^{-1}$ was suppressed.
This is because image embeddings under the brightness corruption retain information for classification; solely addressing entropy can improve accuracy.
The defocus blur in (b) and elastic transform in (c) also showed similar evolutions in which the entropy quickly increases.
However, the maximum value of the weight for entropy differs, and the weight for the uniformity loss is retained.
This suggests the uniformity loss needs to be minimized along with entropy to retain information for these corruptions.
On the other hand, the Gaussian noise in (d) showed a different evolution: the weight is larger for uniformity loss than for the entropy in the initial phase, indicating that retaining information of image embeddings is prioritized.

Next, we visualized embeddings to observe how the uniformity is improved.
As the CLIP's embeddings are normalized and distributed on a unit hypersphere, we used the spherical PCA~\citep{liu2019spherical}, which projects data points on a high-dimensional unit hypersphere onto a low-dimensional hypersphere (a 2D circle here).
\cref{fig:spca_visualization} visualizes the image embeddings before and after TTA with UnInfo on several corruptions of ImageNet-C, along with the text embeddings.
After TTA, the image embeddings are distributed in a broader range of the circle than those before TTA on all corruptions, which suggests that the uniformity is improved.
Moreover, the distributions of image and text embeddings are aligned after TTA, which suggests that the image embeddings become more classification-friendly.
Uniformity improved most significantly in the Gaussian noise corruption in (d) because uniformity loss was prioritized by the information-aware loss balancing, as shown in \cref{fig:weight_evolution}~(d), as we intended.
On the other hand, uniformity improved less significantly in the brightness corruption in (a) than in the other corruptions since the entropy is highly prioritized for the brightness corruption in \cref{fig:weight_evolution}~(a).
In other words, uniformity is less important for this corruption because the image embeddings are already spread in a broader range than for the other corruptions, and their distribution is already aligned with that of the text embeddings before TTA.

\subsubsection{Computational Efficiency}\label{sssec:exp_computation_time}
\begin{table}[tb]
    \centering
    \caption{Mean adaptation throughput and GPU memory usage.}
    \label{tab:computation_time}
    \resizebox{0.7\linewidth}{!}{
        \begin{tabular}{lcc}\toprule
            Method                                         & Throughput~(images/sec.)       & GPU memory~(MiB) \\ \midrule
            %=============
            No-adapt                                       & ${{299.6}_{\pm 1.0}}$          & ${1348}$         \\ \midrule
            TPT~\citep{shu2022tpt}                         & ${{1.8}_{\pm 0.0}}$            & ${14409}$        \\
            ZERO~\citep{farina2024frustratingly}           & ${{4.2}_{\pm 0.0}}$            & ${2319}$         \\
            MTA~\citep{zanella2024test}                    & ${{1.3}_{\pm 0.0}}$            & ${2331}$         \\
            TDA~\citep{karmanov2024efficient}              & ${{44.6}_{\pm 2.4}}$           & $\mathbf{1414}$  \\
            \highlight{DMN~\citep{zhang2024dual}}          & ${5.7}_{\pm 0.06}$             & $5760$           \\
            \highlight{BATCLIP~\citep{Maharana_2025_ICCV}} & ${{86.72}_{\pm 0.3}}$          & ${17456}$        \\
            \highlight{Mint~\citep{bao2026mint}}           & $\mathbf{{161.16}_{\pm 1.64}}$ & $3112$           \\
            UnInfo (ours)                                  & ${{73.7}_{\pm 0.5}}$           & ${11736}$        \\ \bottomrule
            %=============
        \end{tabular}
    }
\end{table}

\cref{tab:computation_time} shows each method's throughput~(images per second) and GPU memory usage~(MiB).
The baseline methods based on marginal confidence over augmented views of an input (TPT~\citep{shu2022tpt}, ZERO~\citep{farina2024frustratingly}, and MTA~\citep{zanella2024test}) had very low throughput because they needed to run forward passes for 64 augmented views per image.
TPT and MTA had the lowest throughput because they require a further backward pass and solve an optimization problem for each image, respectively.
Especially, TPT had a high GPU memory usage because of backward pass for prompt optimization.
\highlight{In contrast, UnInfo had higher throughput than them since it does not require data augmentation and only requires one forward and backward pass per image, though it had higher GPU memory usage.
    TDA had a higher throughput than TPT, ZERO, MTA, and DMN because it does not rely on either backpropagation or augmented views, but UnInfo had a higher throughput than TDA.
    This is because UnInfo can leverage GPUs' parallelization by processing test samples in batches and utilizing LoRA, whereas UnInfo requires backpropagation.
    In contrast, TDA requires one-by-one data processing to update the cache, which hinders GPU acceleration.
    BATCLIP and Mint had higher throughput than UnInfo because they update layer normalization parameters, which have fewer parameters than LoRA.
    Mint had less GPU memory than UnInfo and BATCLIP due to the batch size.
    Although UnInfo's throughput is not higher than BATCLIP's or Mint's, its efficiency is also reasonable.}
Moreover, UnInfo can further speed up inference and save GPU memory as much as No-adapt because the knowledge of the test distribution is accumulated in the LoRA adapters; one may stop adaptation and merge LoRA to the stem model when the test distribution is stable.
In contrast, TPT, ZERO, and MTA are episodic methods, i.e., they do not update the model or accumulate any information.
Thus, they always have to perform adaptation and inference together, unlike UnInfo.

\section{Related Work} \label{sec:related_work}
\subsection{Contrastive Language-image Pre-training}\label{ssec:related_work_clip}
CLIP~\citep{clip_paper} is a multimodal foundation model training paradigm, especially between image and text modalities.
In CLIP, two encoders (an image encoder and a text encoder), are trained to map image and text inputs into a unified embedding space so that semantically corresponding inputs are mapped to close embeddings and vice versa.
Although the training strategy is simple, CLIP has demonstrated remarkable generalization on downstream tasks by being trained on a huge dataset (e.g., hundreds of millions of image-text pairs).
However, CLIP degrades downstream performance when faced with datasets with a large gap from the training dataset~\citep{Zhang2022tip-adapter,huang2024lp++,chen2023plot,shu2022tpt,zhou2024test,karmanov2024efficient,zhang2024dual,zanella2024test,wang2024a,qian2024online}.
Re-training is often infeasible for adapting CLIP to a new dataset because it incurs a substantial computational cost, as described above.
To address this challenge, TTA of CLIP has been actively studied.

\subsection{Test-time Adaptation of Vision-language Models}\label{ssec:related_work_tta}

To adapt instantly to test distributions without incurring high computational costs, TTA for zero-shot classification with CLIP has been studied.
TTA aims to adapt a zero-shot CLIP classifier to the test distribution with only unlabeled test data.
\highlight{Existing CLIP TTA methods can be classified along two axes: episodic/cumulative and frozen/updating encoders.}

\paragraph{\highlight{Episodic / frozen encoders.}}
The representative approach of CLIP TTA is to update the image and/or text embeddings.
MTA~\citep{zanella2024test} selects reliable image embeddings and updates the feature centroid.
ZERO~\citep{farina2024frustratingly} dynamically correct the prediction probabilities for each test input.

\paragraph{\highlight{Cumulative / frozen encoders.}}
Test-time prompt tuning approaches~\citep{shu2022tpt,yoon2024c,Wang_2025_CVPR} updates the text token embeddings (e.g., four embedding vectors corresponding to words of a prompt template ``\texttt{a photo of a}'') during testing to minimize the prediction entropy marginalized with augmented views of an input image.
The text embedding corresponding to each class is updated to be more appropriate to the current domain by updating the text token embeddings.
Existing method also directly update the embeddings or logits computed from the similarity of image and text embeddings.
TDA~\citep{karmanov2024efficient} and DMN~\citep{zhang2024dual} accumulate test inputs and construct the image embedding caches to modify subsequent inputs' predictions.
\highlight{OnZeta~\citep{qian2024online}, DPE~\citep{zhang2024dpe} and TPS~\citep{sui2025just} dynamically update class weights and/or visual prototypes.
    STS~\citep{dafnis2026test} updates text prototypes within a spectral subspace.}

The above frozen encoder methods can adapt well to domain shifts, such as changes in environments, rendition, or out-of-distribution~\citep{imagenet-r,recht2019imagenet,Hendrycks_2021_CVPR,robust_global_representation_neurips2019}, using fixed image and text encoders.
This is because a pre-trained CLIP generalizes to a wide range of domains enough to encode the current domain's semantics properly.

\paragraph{\highlight{Cumulative / updating encoders.}}
\highlight{More recently, TTA methods for CLIP targeting image corruption have been developed~\citep{bao2026mint,Maharana_2025_ICCV,A_Vargas_Hakim_2025_WACV,osowiechi2024watt}.
    These methods update the image and text encoders to more effectively mitigate embedding collapse caused by image corruption.
    In particular, Mint~\citep{bao2026mint} is similar to UnInfo in that it maintains inter-class variance in image embeddings, which is close to uniformity.
    While these methods are effective to image corruption, we found that they improve accuracy on not all corruption types and model architectures (\cref{sssec:exp_adaptataion_performance}).}

\highlight{Our contributions include the finding that uniformity~\citep{wang2020understanding,wang2023feature} especially matters in the embedding space under image corruption (\cref{sec:preliminary_experiment}).}

\section{Conclusion}\label{sec:conclusion}
We proposed UnInfo, a novel test-time adaptation (TTA) method for zero-shot classification with vision-language models under image corruption.
Unlike existing methods, UnInfo updates the image encoder to address the specific challenge of image corruption, where loss input information is retained in the image embeddings unlike other natural distribution shifts.
In the experiments, UnInfo achieved higher classification performance than baselines by refining the uniformity along with the entropy, and the information-aware loss balancing further improved the performance.
One limitation of UnInfo is that it requires test data to be mini-batched \highlight{and sampled i.i.d. to compute the uniformity loss and information-aware balancing weight.
    Our future work is to extend UnInfo to a fully online setting and to broader types of realistic distribution shifts, e.g., real-world sensor shifts~\citep{Baek_2024_CVPR}, mixed severity/corruption types and non-i.i.d. test streams, which have been studied recently~\citep{yuan2023robust,zanella2025realistic}.}

\newpage

\bibliography{0_main}

\begin{thebibliography}{89}
\providecommand{\natexlab}[1]{#1}
\providecommand{\url}[1]{\texttt{#1}}
\expandafter\ifx\csname urlstyle\endcsname\relax
  \providecommand{\doi}[1]{doi: #1}\else
  \providecommand{\doi}{doi: \begingroup \urlstyle{rm}\Url}\fi

\bibitem[A~Vargas~Hakim et~al.(2025)A~Vargas~Hakim, Osowiechi, Noori, Cheraghalikhani, Bahri, Yazdanpanah, Ben~Ayed, and Desrosiers]{A_Vargas_Hakim_2025_WACV}
Gustavo A~Vargas~Hakim, David Osowiechi, Mehrdad Noori, Milad Cheraghalikhani, Ali Bahri, Moslem Yazdanpanah, Ismail Ben~Ayed, and Christian Desrosiers.
\newblock {CLIPArTT}: Adaptation of {CLIP} to new domains at test time.
\newblock In \emph{Proceedings of the Winter Conference on Applications of Computer Vision (WACV)}, pp.\  7092--7101, February 2025.

\bibitem[Adachi et~al.(2023)Adachi, Yamaguchi, and Kumagai]{cafe_adachi}
Kazuki Adachi, Shin’ya Yamaguchi, and Atsutoshi Kumagai.
\newblock Covariance-aware feature alignment with pre-computed source statistics for test-time adaptation to multiple image corruptions.
\newblock In \emph{Proceedings of the 2023 IEEE International Conference on Image Processing (ICIP)}, pp.\  800--804, 2023.

\bibitem[Adachi et~al.(2024)Adachi, Enomoto, Sasaki, and Yamaguchi]{temp_adachi}
Kazuki Adachi, Shohei Enomoto, Taku Sasaki, and Shin’ya Yamaguchi.
\newblock Test-time similarity modification for person re-identification toward temporal distribution shift.
\newblock In \emph{Proceedings of the 2024 International Joint Conference on Neural Networks (IJCNN)}, pp.\  1--8, 2024.

\bibitem[Ba et~al.(2016)Ba, Kiros, and Hinton]{ba2016layer}
Jimmy~Lei Ba, Jamie~Ryan Kiros, and Geoffrey~E Hinton.
\newblock Layer normalization.
\newblock \emph{arXiv preprint arXiv:1607.06450}, 2016.

\bibitem[Baek et~al.(2024)Baek, Park, Kim, and Kim]{Baek_2024_CVPR}
Eunsu Baek, Keondo Park, Jiyoon Kim, and Hyung-Sin Kim.
\newblock Unexplored faces of robustness and out-of-distribution: Covariate shifts in environment and sensor domains.
\newblock In \emph{Proceedings of the IEEE/CVF Conference on Computer Vision and Pattern Recognition (CVPR)}, pp.\  22294--22303, June 2024.

\bibitem[Baldrati et~al.(2022)Baldrati, Bertini, Uricchio, and Del~Bimbo]{baldrati2022effective}
Alberto Baldrati, Marco Bertini, Tiberio Uricchio, and Alberto Del~Bimbo.
\newblock Effective conditioned and composed image retrieval combining {CLIP}-based features.
\newblock In \emph{Proceedings of the IEEE/CVF Conference on Computer Vision and Pattern Recognition (CVPR)}, pp.\  21466--21474, 2022.

\bibitem[Bao et~al.(2025)Bao, Deng, and He]{bao2026mint}
Wenxuan Bao, Ruxi Deng, and Jingrui He.
\newblock Mint: A simple test-time adaptation of vision-language models against common corruptions.
\newblock \emph{Advances in Neural Information Processing Systems (NeurIPS)}, 38:\penalty0 58666--58700, 2025.

\bibitem[Bridle et~al.(1991)Bridle, Heading, and MacKay]{NIPS1991_a8abb4bb}
John Bridle, Anthony Heading, and David MacKay.
\newblock Unsupervised classifiers, mutual information and `phantom targets'.
\newblock \emph{Advances in Neural Information Processing Systems (NIPS)}, 4:\penalty0 1096--1101, 1991.

\bibitem[Chen et~al.(2023)Chen, Yao, Song, Li, Rao, and Zhang]{chen2023plot}
Guangyi Chen, Weiran Yao, Xiangchen Song, Xinyue Li, Yongming Rao, and Kun Zhang.
\newblock {PLOT}: Prompt learning with optimal transport for vision-language models.
\newblock In \emph{Proceedings of the Eleventh International Conference on Learning Representations (ICLR)}, 2023.

\bibitem[Cherti et~al.(2023)Cherti, Beaumont, Wightman, Wortsman, Ilharco, Gordon, Schuhmann, Schmidt, and Jitsev]{cherti2023reproducible}
Mehdi Cherti, Romain Beaumont, Ross Wightman, Mitchell Wortsman, Gabriel Ilharco, Cade Gordon, Christoph Schuhmann, Ludwig Schmidt, and Jenia Jitsev.
\newblock Reproducible scaling laws for contrastive language-image learning.
\newblock In \emph{Proceedings of the IEEE/CVF Conference on Computer Vision and Pattern Recognition (CVPR)}, pp.\  2818--2829, 2023.

\bibitem[Dafnis \& Metaxas(2026)Dafnis and Metaxas]{dafnis2026test}
Konstantinos Dafnis and Dimitris Metaxas.
\newblock Test-time spectrum-aware latent steering for zero-shot generalization in vision-language models.
\newblock \emph{Advances in Neural Information Processing Systems (NeurIPS)}, 38:\penalty0 151169--151194, 2026.

\bibitem[Dai \& Gool(2018)Dai and Gool]{dai2018dark}
Dengxin Dai and Luc~Van Gool.
\newblock Dark model adaptation: Semantic image segmentation from daytime to nighttime.
\newblock In \emph{Proceedings of the 21st International Conference on Intelligent Transportation Systems (ITSC)}, pp.\  3819--3824, 2018.

\bibitem[Deng et~al.(2009)Deng, Dong, Socher, Li, Li, and Fei-Fei]{imagenet}
Jia Deng, Wei Dong, Richard Socher, Li-Jia Li, Kai Li, and Li~Fei-Fei.
\newblock {ImageNet}: A large-scale hierarchical image database.
\newblock In \emph{Proceedings of the IEEE/CVF Conference on Computer Vision and Pattern Recognition (CVPR)}, pp.\  248--255, 2009.

\bibitem[D\"obler et~al.(2023)D\"obler, Marsden, and Yang]{Dobler_2023_CVPR}
Mario D\"obler, Robert~A. Marsden, and Bin Yang.
\newblock Robust mean teacher for continual and gradual test-time adaptation.
\newblock In \emph{Proceedings of the IEEE/CVF Conference on Computer Vision and Pattern Recognition (CVPR)}, pp.\  7704--7714, June 2023.

\bibitem[Dong et~al.(2026)Dong, Sheng, Liang, He, Chatzi, and Fink]{dong2025adapting}
Hao Dong, Lijun Sheng, Jian Liang, Ran He, Eleni Chatzi, and Olga Fink.
\newblock Adapting vision-language models without labels: A comprehensive survey.
\newblock \emph{International Journal of Computer Vision (IJCV)}, 134\penalty0 (8), 2026.

\bibitem[Eastwood et~al.(2022)Eastwood, Mason, Williams, and Sch{\"o}lkopf]{eastwood2022sourcefree}
Cian Eastwood, Ian Mason, Chris Williams, and Bernhard Sch{\"o}lkopf.
\newblock Source-free adaptation to measurement shift via bottom-up feature restoration.
\newblock In \emph{Proceedings of the Tenth International Conference on Learning Representations (ICLR)}, 2022.

\bibitem[Enomoto et~al.(2024)Enomoto, Hasegawa, Adachi, Sasaki, Yamaguchi, Suzuki, and Eda]{enomoto2024test}
Shohei Enomoto, Naoya Hasegawa, Kazuki Adachi, Taku Sasaki, Shin'ya Yamaguchi, Satoshi Suzuki, and Takeharu Eda.
\newblock Test-time adaptation meets image enhancement: Improving accuracy via uncertainty-aware logit switching.
\newblock In \emph{Proceedings of the 2024 International Joint Conference on Neural Networks (IJCNN)}, pp.\  1--8, 2024.

\bibitem[Eslami \& de~Melo(2025)Eslami and de~Melo]{eslami2025mitigate}
Sedigheh Eslami and Gerard de~Melo.
\newblock Mitigate the gap: Improving cross-modal alignment in {CLIP}.
\newblock In \emph{Proceedings of The Thirteenth International Conference on Learning Representations (ICLR)}, 2025.

\bibitem[Fang et~al.(2021)Fang, Xiong, Xu, and Chen]{fang2021clip2video}
Han Fang, Pengfei Xiong, Luhui Xu, and Yu~Chen.
\newblock {CLIP2Video}: Mastering video-text retrieval via image {CLIP}.
\newblock \emph{arXiv preprint arXiv:2106.11097}, 2021.

\bibitem[Farina et~al.(2024)Farina, Franchi, Iacca, Mancini, and Ricci]{farina2024frustratingly}
Matteo Farina, Gianni Franchi, Giovanni Iacca, Massimiliano Mancini, and Elisa Ricci.
\newblock Frustratingly easy test-time adaptation of vision-language models.
\newblock \emph{Advances in Neural Information Processing Systems (NeurIPS)}, 37:\penalty0 129062--129093, 2024.

\bibitem[Flamary et~al.(2021)Flamary, Courty, Gramfort, Alaya, Boisbunon, Chambon, Chapel, Corenflos, Fatras, Fournier, Gautheron, Gayraud, Janati, Rakotomamonjy, Redko, Rolet, Schutz, Seguy, Sutherland, Tavenard, Tong, and Vayer]{flamary2021pot}
R{\'e}mi Flamary, Nicolas Courty, Alexandre Gramfort, Mokhtar~Z. Alaya, Aur{\'e}lie Boisbunon, Stanislas Chambon, Laetitia Chapel, Adrien Corenflos, Kilian Fatras, Nemo Fournier, L{\'e}o Gautheron, Nathalie~T.H. Gayraud, Hicham Janati, Alain Rakotomamonjy, Ievgen Redko, Antoine Rolet, Antony Schutz, Vivien Seguy, Danica~J. Sutherland, Romain Tavenard, Alexander Tong, and Titouan Vayer.
\newblock {POT}: Python optimal transport.
\newblock \emph{Journal of Machine Learning Research (JMLR)}, 22\penalty0 (78):\penalty0 1--8, 2021.

\bibitem[Flamary et~al.(2024)Flamary, Vincent-Cuaz, Courty, Gramfort, Kachaiev, Quang~Tran, David, Bonet, Cassereau, Gnassounou, Tanguy, Delon, Collas, Mazelet, Chapel, Kerdoncuff, Yu, Feickert, Krzakala, Liu, and Fernandes~Montesuma]{flamary2024pot}
R{\'e}mi Flamary, C{\'e}dric Vincent-Cuaz, Nicolas Courty, Alexandre Gramfort, Oleksii Kachaiev, Huy Quang~Tran, Laurène David, Cl{\'e}ment Bonet, Nathan Cassereau, Th{\'e}o Gnassounou, Eloi Tanguy, Julie Delon, Antoine Collas, Sonia Mazelet, Laetitia Chapel, Tanguy Kerdoncuff, Xizheng Yu, Matthew Feickert, Paul Krzakala, Tianlin Liu, and Eduardo Fernandes~Montesuma.
\newblock {POT}: Python optimal transport (version 0.9.5), 2024.
\newblock URL \url{https://github.com/PythonOT/POT/tree/0.9.5}.

\bibitem[Gao et~al.(2022)Gao, Shi, Zhu, Wang, Tang, Zhou, Li, and Metaxas]{gao2022visual}
Yunhe Gao, Xingjian Shi, Yi~Zhu, Hao Wang, Zhiqiang Tang, Xiong Zhou, Mu~Li, and Dimitris~N Metaxas.
\newblock Visual prompt tuning for test-time domain adaptation.
\newblock \emph{arXiv preprint arXiv:2210.04831}, 2022.

\bibitem[Ge et~al.(2023)Ge, Ren, Gallagher, Wang, Yang, Adam, Itti, Lakshminarayanan, and Zhao]{ge2023improving}
Yunhao Ge, Jie Ren, Andrew Gallagher, Yuxiao Wang, Ming-Hsuan Yang, Hartwig Adam, Laurent Itti, Balaji Lakshminarayanan, and Jiaping Zhao.
\newblock Improving zero-shot generalization and robustness of multi-modal models.
\newblock In \emph{Proceedings of the IEEE/CVF Conference on Computer Vision and Pattern Recognition (CVPR)}, pp.\  11093--11101, 2023.

\bibitem[He et~al.(2015)He, Zhang, Ren, and Sun]{He_2015_ICCV}
Kaiming He, Xiangyu Zhang, Shaoqing Ren, and Jian Sun.
\newblock Delving deep into rectifiers: Surpassing human-level performance on imagenet classification.
\newblock In \emph{Proceedings of the IEEE International Conference on Computer Vision (ICCV)}, pp.\  1026--1034, December 2015.

\bibitem[Hendrycks \& Dietterich(2019)Hendrycks and Dietterich]{imagenet-c}
Dan Hendrycks and Thomas Dietterich.
\newblock Benchmarking neural network robustness to common corruptions and perturbations.
\newblock In \emph{Proceedings of the Seventh International Conference on Learning Representations (ICLR)}, 2019.

\bibitem[Hendrycks et~al.(2021{\natexlab{a}})Hendrycks, Basart, Mu, Kadavath, Wang, Dorundo, Desai, Zhu, Parajuli, Guo, Song, Steinhardt, and Gilmer]{imagenet-r}
Dan Hendrycks, Steven Basart, Norman Mu, Saurav Kadavath, Frank Wang, Evan Dorundo, Rahul Desai, Tyler Zhu, Samyak Parajuli, Mike Guo, Dawn Song, Jacob Steinhardt, and Justin Gilmer.
\newblock The many faces of robustness: A critical analysis of out-of-distribution generalization.
\newblock In \emph{Proceedings of the IEEE/CVF International Conference on Computer Vision (ICCV)}, pp.\  8340--8349, October 2021{\natexlab{a}}.

\bibitem[Hendrycks et~al.(2021{\natexlab{b}})Hendrycks, Zhao, Basart, Steinhardt, and Song]{Hendrycks_2021_CVPR}
Dan Hendrycks, Kevin Zhao, Steven Basart, Jacob Steinhardt, and Dawn Song.
\newblock Natural adversarial examples.
\newblock In \emph{Proceedings of the IEEE/CVF Conference on Computer Vision and Pattern Recognition (CVPR)}, pp.\  15262--15271, June 2021{\natexlab{b}}.

\bibitem[Hu et~al.(2022)Hu, yelong shen, Wallis, Allen-Zhu, Li, Wang, Wang, and Chen]{hu2022lora}
Edward~J Hu, yelong shen, Phillip Wallis, Zeyuan Allen-Zhu, Yuanzhi Li, Shean Wang, Lu~Wang, and Weizhu Chen.
\newblock Lo{RA}: Low-rank adaptation of large language models.
\newblock In \emph{Proceedings of the Tenth International Conference on Learning Representations (ICLR)}, 2022.

\bibitem[Hu et~al.(2017)Hu, Miyato, Tokui, Matsumoto, and Sugiyama]{hu2017learning}
Weihua Hu, Takeru Miyato, Seiya Tokui, Eiichi Matsumoto, and Masashi Sugiyama.
\newblock Learning discrete representations via information maximizing self-augmented training.
\newblock In \emph{Proceedings of the Thirty-Fourth International Conference on Machine Learning (ICML)}, pp.\  1558--1567. PMLR, 2017.

\bibitem[Hu et~al.(2025)Hu, Hu, Li, Tang, and Duan]{hu2025beyond}
Zixuan Hu, Yichun Hu, Xiaotong Li, Shixiang Tang, and Lingyu Duan.
\newblock Beyond entropy: Region confidence proxy for wild test-time adaptation.
\newblock In \emph{Proceedings of the Forty-Second International Conference on Machine Learning (ICML)}, pp.\  24371--24390. PMLR, 2025.

\bibitem[Huang et~al.(2024)Huang, Shakeri, Dolz, Boudiaf, Bahig, and Ben~Ayed]{huang2024lp++}
Yunshi Huang, Fereshteh Shakeri, Jose Dolz, Malik Boudiaf, Houda Bahig, and Ismail Ben~Ayed.
\newblock {LP++}: A surprisingly strong linear probe for few-shot {CLIP}.
\newblock In \emph{Proceedings of the IEEE/CVF Conference on Computer Vision and Pattern Recognition (CVPR)}, pp.\  23773--23782, 2024.

\bibitem[Hurst et~al.(2024)Hurst, Lerer, Goucher, Perelman, Ramesh, Clark, Ostrow, Welihinda, Hayes, Radford, Mkadry, Baker-Whitcomb, Beutel, Borzunov, Carney, Chow, Kirillov, Nichol, Paino, Renzin, Passos, Kirillov, Christakis, Conneau, Kamali, Jabri, Moyer, Tam, Crookes, Tootoochian, Tootoonchian, Kumar, Vallone, Karpathy, Braunstein, Cann, Codispoti, Galu, Kondrich, Tulloch, Mishchenko, Baek, Jiang, toine Pelisse, Woodford, Gosalia, Dhar, Pantuliano, Nayak, Oliver, Zoph, Ghorbani, Leimberger, Rossen, Sokolowsky, Wang, Zweig, Hoover, Samic, McGrew, Spero, Giertler, Cheng, Lightcap, Walkin, Quinn, Guarraci, Hsu, Kellogg, Eastman, Lugaresi, Wainwright, Bassin, Hudson, Chu, Nelson, Li, Shern, Conger, Barette, Voss, Ding, Lu, Zhang, Beaumont, Hallacy, Koch, Gibson, Kim, Choi, McLeavey, Hesse, Fischer, Winter, Czarnecki, Jarvis, Wei, Koumouzelis, Sherburn, Kappler, Levin, Levy, Carr, Farhi, M{\'e}ly, Robinson, Sasaki, Jin, Valladares, Tsipras, Li, Nguyen, Findlay, Oiwoh, Wong, Asdar, Proehl, Yang, Antonow,
  Kramer, Peterson, Sigler, Wallace, Brevdo, Mays, Khorasani, Such, Raso, Zhang, von Lohmann, Sulit, Goh, Oden, Salmon, Starace, Brockman, Salman, Bao, Hu, Wong, Wang, Schmidt, Whitney, woo Jun, Kirchner, de~Oliveira~Pinto, Ren, Chang, Chung, Kivlichan, O'Connell, Osband, Silber, Sohl, Okuyucu, Lan, Kostrikov, Sutskever, Kanitscheider, Gulrajani, Coxon, Menick, Pachocki, Aung, Betker, Crooks, Lennon, Kiros, Leike, Park, Kwon, Phang, Teplitz, Wei, Wolfe, Chen, Harris, Varavva, Lee, Shieh, Lin, Yu, Weng, Tang, Yu, Jang, Candela, Beutler, Landers, Parish, Heidecke, Schulman, Lachman, Mckay, Uesato, Ward, Kim, Huizinga, Sitkin, Kraaijeveld, Gross, Kaplan, Snyder, Achiam, Jiao, Lee, Zhuang, Harriman, Fricke, Hayashi, Singhal, Shi, Karthik, Wood, Rimbach, Hsu, Nguyen, Gu-Lemberg, Button, Liu, Howe, Muthukumar, Luther, Ahmad, Kai, Itow, Workman, Pathak, Chen, Jing, Guy, Fedus, Zhou, Mamitsuka, Weng, McCallum, Held, Long, Feuvrier, Zhang, Kondraciuk, Kaiser, Hewitt, Metz, Doshi, Aflak, Simens, laine Boyd, Thompson,
  Dukhan, Chen, Gray, Hudnall, Zhang, Aljubeh, teusz Litwin, Zeng, Johnson, Shetty, Gupta, Shah, Yatbaz, Yang, Zhong, Glaese, Chen, Janner, Lampe, Petrov, Wu, Wang, Fradin, Pokrass, Castro, Castro, Pavlov, Brundage, Wang, Khan, Murati, Bavarian, Lin, Yesildal, Soto, Gimelshein, talie Cone, Staudacher, Summers, LaFontaine, Chowdhury, Ryder, Stathas, Turley, Tezak, Felix, Kudige, Keskar, Deutsch, Bundick, Puckett, Nachum, Okelola, Boiko, Murk, Jaffe, Watkins, Godement, Campbell-Moore, Chao, McMillan, Belov, Su, Bak, Bakkum, Deng, Dolan, Hoeschele, Welinder, Tillet, Pronin, Tillet, Dhariwal, ing Yuan, Dias, Lim, Arora, Troll, Lin, Lopes, Puri, Miyara, Leike, Gaubert, Zamani, Wang, Donnelly, Honsby, Smith, Sahai, Ramchandani, Huet, Carmichael, Zellers, Chen, Chen, Nigmatullin, Cheu, Jain, Altman, Schoenholz, Toizer, Miserendino, Agarwal, Culver, Ethersmith, Gray, Grove, Metzger, Hermani, Jain, Zhao, Wu, Jomoto, Wu, Xia, Phene, Papay, Narayanan, Coffey, Lee, Hall, Balaji, Broda, Stramer, Xu, Gogineni,
  Christianson, Sanders, Patwardhan, Cunninghman, Degry, Dimson, Raoux, Shadwell, Zheng, Underwood, Markov, Sherbakov, Rubin, Stasi, Kaftan, Heywood, Peterson, Walters, Eloundou, Qi, Moeller, Monaco, Kuo, Fomenko, Chang, Zheng, Zhou, Manassra, Sheu, Zaremba, Patil, Qian, Kim, Cheng, Zhang, He, Zhang, Jin, Dai, and Malkov]{hurst2024gpt}
Aaron Hurst, Adam Lerer, Adam~P. Goucher, Adam Perelman, Aditya Ramesh, Aidan Clark, AJ~Ostrow, Akila Welihinda, Alan Hayes, Alec Radford, Aleksander Mkadry, Alex Baker-Whitcomb, Alex Beutel, Alex Borzunov, Alex Carney, Alex Chow, Alexander Kirillov, Alex Nichol, Alex Paino, Alex Renzin, Alexandre Passos, Alexander Kirillov, Alexi Christakis, Alexis Conneau, Ali Kamali, Allan Jabri, Allison Moyer, Allison Tam, Amadou Crookes, Amin Tootoochian, Amin Tootoonchian, Ananya Kumar, Andrea Vallone, Andrej Karpathy, Andrew Braunstein, Andrew Cann, Andrew Codispoti, Andrew Galu, Andrew Kondrich, Andrew Tulloch, Andrey Mishchenko, Angela Baek, Angela Jiang, An~toine Pelisse, Antonia Woodford, Anuj Gosalia, Arka Dhar, Ashley Pantuliano, Avi Nayak, Avital Oliver, Barret Zoph, B.~Ghorbani, Ben Leimberger, Ben Rossen, Benjamin Sokolowsky, Ben Wang, Benjamin Zweig, Beth Hoover, Blake Samic, Bob McGrew, Bobby Spero, Bogo Giertler, Bowen Cheng, Brad Lightcap, Brandon Walkin, Brendan Quinn, Brian Guarraci, Brian Hsu, Bright
  Kellogg, Brydon Eastman, Camillo Lugaresi, Carroll~L. Wainwright, Cary Bassin, Cary Hudson, Casey Chu, Chad Nelson, Chak Li, Chan~Jun Shern, Channing Conger, Charlotte Barette, Chelsea Voss, Chen Ding, Cheng Lu, Chong Zhang, Chris Beaumont, Chris Hallacy, Chris Koch, Christian Gibson, Christina Kim, Christine Choi, Christine McLeavey, Chris Hesse, Claudia Fischer, Clemens Winter, Coley Czarnecki, Colin Jarvis, Colin Wei, Constantin Koumouzelis, Dane Sherburn, Daniel Kappler, Daniel Levin, Daniel Levy, David Carr, David Farhi, David M{\'e}ly, David Robinson, David Sasaki, Denny Jin, Dev Valladares, Dimitris Tsipras, Doug Li, Phong~Duc Nguyen, Duncan Findlay, Edede Oiwoh, Edmund Wong, Ehsan Asdar, Elizabeth Proehl, Elle~Michelle Yang, Eric Antonow, Eric Kramer, Eric Peterson, Eric Sigler, Eric Wallace, Eugene Brevdo, Evan Mays, Farzad Khorasani, Felipe~Petroski Such, Filippo Raso, Francis Zhang, Fred von Lohmann, Freddie Sulit, Gabriel Goh, Gene Oden, Geoff Salmon, Giulio Starace, Greg Brockman, Hadi Salman,
  Hai-Biao Bao, Haitang Hu, Hannah Wong, Haoyu Wang, Heather Schmidt, Heather Whitney, Hee woo Jun, Hendrik Kirchner, Henrique~Pond{\'e} de~Oliveira~Pinto, Hongyu Ren, Hui-Wen Chang, Hyung~Won Chung, Ian Kivlichan, Ian~R. O'Connell, Ian Osband, Ian Silber, Ian Sohl, Ibrahim Okuyucu, Ikai Lan, Ilya Kostrikov, Ilya Sutskever, Ingmar Kanitscheider, Ishaan Gulrajani, Jacob Coxon, Jacob Menick, Jakub~W. Pachocki, James Aung, James Betker, James Crooks, James Lennon, Jamie~Ryan Kiros, Jan Leike, Jane Park, Jason Kwon, Jason Phang, Jason Teplitz, Jason Wei, Jason Wolfe, Jay Chen, Jeff Harris, Jenia Varavva, Jessica~Gan Lee, Jessica Shieh, Ji~Lin, Jiahui Yu, Jiayi Weng, Jie Tang, Jieqi Yu, Joanne Jang, Joaquin~Qui{\~n}onero Candela, Joe Beutler, Joe Landers, Joel Parish, Johannes Heidecke, John Schulman, Jonathan Lachman, J.~Mckay, Jonathan Uesato, Jonathan Ward, Jong~Wook Kim, Joost Huizinga, Jordan Sitkin, Jos Kraaijeveld, Joshua Gross, Josh Kaplan, Josh Snyder, Joshua Achiam, Joy Jiao, Joyce Lee, Juntang Zhuang,
  Justyn Harriman, Kai Fricke, Kai Hayashi, Karan Singhal, Katy Shi, Kavin Karthik, Kayla Wood, Kendra Rimbach, Kenny Hsu, Kenny Nguyen, Keren Gu-Lemberg, Kevin Button, Kevin Liu, Kiel Howe, Krithika Muthukumar, Kyle Luther, Lama Ahmad, Larry Kai, Lauren Itow, Lauren Workman, Leher Pathak, Leo Chen, Li~Jing, Lia Guy, Liam Fedus, Liang Zhou, Lien Mamitsuka, Lilian Weng, Lindsay McCallum, Lindsey Held, Ouyang Long, Louis Feuvrier, Lu~Zhang, Lukasz Kondraciuk, Lukasz Kaiser, Luke Hewitt, Luke Metz, Lyric Doshi, Mada Aflak, Maddie Simens, Made laine Boyd, Madeleine Thompson, Marat Dukhan, Mark Chen, Mark Gray, Mark Hudnall, Marvin Zhang, Marwan Aljubeh, Ma~teusz Litwin, Matthew Zeng, Max Johnson, Maya Shetty, Mayank~Mike Gupta, Meghan Shah, Mehmet~Ali Yatbaz, Mengxue Yang, Mengchao Zhong, Mia Glaese, Mianna Chen, Michael Janner, Michael Lampe, Michael Petrov, Michael Wu, Michele Wang, Michelle Fradin, Michelle Pokrass, Miguel Castro, Miguel Castro, Mikhail Pavlov, Miles Brundage, Miles Wang, Mina Khan, Mira
  Murati, Mo~Bavarian, Molly Lin, Murat Yesildal, Nacho Soto, Natalia Gimelshein, Na~talie Cone, Natalie~M. Staudacher, Natalie Summers, Natan LaFontaine, Neil Chowdhury, Nick Ryder, Nick Stathas, Nick Turley, Nikolas~A. Tezak, Niko Felix, Nithanth Kudige, Nitish~Shirish Keskar, Noah Deutsch, Noel Bundick, Nora Puckett, Ofir Nachum, O.~E. Okelola, Oleg Boiko, Oleg Murk, Oliver Jaffe, Olivia Watkins, Olivier Godement, Owen Campbell-Moore, Patrick Chao, Paul McMillan, Pavel Belov, Peng Su, Peter Bak, Peter Bakkum, Peter Deng, Peter Dolan, Peter Hoeschele, Peter Welinder, Phil Tillet, Philip Pronin, Phil Tillet, Prafulla Dhariwal, Qim ing Yuan, Rachel Dias, Rachel Lim, Rahul Arora, Rajan Troll, Randall Lin, Raphael~Gontijo Lopes, Raul Puri, Reah Miyara, Reimar~H. Leike, Renaud Gaubert, Reza Zamani, Ricky~Ben Wang, Rob Donnelly, Rob Honsby, Rocky Smith, Rohan Sahai, Rohit Ramchandani, Romain Huet, Rory Carmichael, Rowan Zellers, Roy Chen, Ruby Chen, Ruslan~Ramilevich Nigmatullin, Ryan Cheu, Saachi Jain, Sam
  Altman, Sam Schoenholz, Sam Toizer, Samuel Miserendino, Sandhini Agarwal, Sara Culver, Scott Ethersmith, Scott Gray, Sean Grove, Sean Metzger, Shamez Hermani, Shantanu Jain, Shengjia Zhao, Sherwin Wu, Shino Jomoto, Shirong Wu, Shuaiqi Xia, Sonia Phene, Spencer Papay, Srinivas Narayanan, Steve Coffey, Steve Yin~Lam Lee, Stewart Hall, Suchir Balaji, Tal Broda, Tal Stramer, Tao Xu, Tarun Gogineni, Taya Christianson, Ted Sanders, Tejal Patwardhan, Thomas Cunninghman, Thomas Degry, Thomas Dimson, Thomas Raoux, Thomas Shadwell, Tianhao Zheng, Todd Underwood, Todor Markov, Toki Sherbakov, Tom Rubin, Tom Stasi, Tomer Kaftan, Tristan Heywood, Troy~A. Peterson, Tyce Walters, Tyna Eloundou, Valerie Qi, Veit Moeller, Vinnie Monaco, Vishal Kuo, Vlad Fomenko, Wayne~H. Chang, Weiyi Zheng, Wenda Zhou, Wesam Manassra, Will Sheu, Wojciech Zaremba, Yash Patil, Yilei Qian, Yongjik Kim, Youlong Cheng, Yu~Zhang, Yuchen He, Yuchen Zhang, Yujia Jin, Yunxing Dai, and Yury Malkov.
\newblock {GPT}-4o system card.
\newblock \emph{arXiv preprint arXiv:2410.21276}, 2024.

\bibitem[Ilharco et~al.(2021)Ilharco, Wortsman, Wightman, Gordon, Carlini, Taori, Dave, Shankar, Namkoong, Miller, Hajishirzi, Farhadi, and Schmidt]{ilharco_gabriel_2021_5143773}
Gabriel Ilharco, Mitchell Wortsman, Ross Wightman, Cade Gordon, Nicholas Carlini, Rohan Taori, Achal Dave, Vaishaal Shankar, Hongseok Namkoong, John Miller, Hannaneh Hajishirzi, Ali Farhadi, and Ludwig Schmidt.
\newblock {OpenCLIP}, July 2021.
\newblock URL \url{https://doi.org/10.5281/zenodo.5143773}.

\bibitem[Karmanov et~al.(2024)Karmanov, Guan, Lu, El~Saddik, and Xing]{karmanov2024efficient}
Adilbek Karmanov, Dayan Guan, Shijian Lu, Abdulmotaleb El~Saddik, and Eric Xing.
\newblock Efficient test-time adaptation of vision-language models.
\newblock In \emph{Proceedings of the IEEE/CVF Conference on Computer Vision and Pattern Recognition (CVPR)}, pp.\  14162--14171, 2024.

\bibitem[Khattak et~al.(2023)Khattak, Rasheed, Maaz, Khan, and Khan]{khattak2023maple}
Muhammad~Uzair Khattak, Hanoona Rasheed, Muhammad Maaz, Salman Khan, and Fahad~Shahbaz Khan.
\newblock Maple: Multi-modal prompt learning.
\newblock In \emph{Proceedings of the IEEE/CVF Conference on Computer Vision and Pattern Recognition (CVPR)}, pp.\  19113--19122, 2023.

\bibitem[Krause et~al.(2010)Krause, Perona, and Gomes]{NIPS2010_42998cf3}
Andreas Krause, Pietro Perona, and Ryan Gomes.
\newblock Discriminative clustering by regularized information maximization.
\newblock \emph{Advances in Neural Information Processing Systems (NIPS)}, 23:\penalty0 775--783, 2010.

\bibitem[Krizhevsky(2009)]{krizhevsky2009learning}
Alex Krizhevsky.
\newblock Learning multiple layers of features from tiny images.
\newblock \emph{Master's thesis, University of Tront}, 2009.

\bibitem[Lai et~al.(2023)Lai, Vesdapunt, Zhou, Wu, Huynh, Li, Fu, and Chuah]{Lai_2023_ICCV}
Zhengfeng Lai, Noranart Vesdapunt, Ning Zhou, Jun Wu, Cong~Phuoc Huynh, Xuelu Li, Kah~Kuen Fu, and Chen-Nee Chuah.
\newblock {PADCLIP}: Pseudo-labeling with adaptive debiasing in {CLIP} for unsupervised domain adaptation.
\newblock In \emph{Proceedings of the IEEE/CVF International Conference on Computer Vision (ICCV)}, pp.\  16155--16165, October 2023.

\bibitem[Liang et~al.(2020)Liang, Hu, and Feng]{liang2020we}
Jian Liang, Dapeng Hu, and Jiashi Feng.
\newblock Do we really need to access the source data? {S}ource hypothesis transfer for unsupervised domain adaptation.
\newblock In \emph{Proceedings of the Thirty-Seventh International Conference on Machine Learning (ICML)}, pp.\  6028--6039. PMLR, 2020.

\bibitem[Liang et~al.(2025)Liang, He, and Tan]{liang2023comprehensive}
Jian Liang, Ran He, and Tieniu Tan.
\newblock A comprehensive survey on test-time adaptation under distribution shifts.
\newblock \emph{International Journal of Computer Vision (IJCV)}, 133\penalty0 (1):\penalty0 31--64, 2025.

\bibitem[Liang et~al.(2022)Liang, Zhang, Kwon, Yeung, and Zou]{liang2022mind}
Victor~Weixin Liang, Yuhui Zhang, Yongchan Kwon, Serena Yeung, and James~Y Zou.
\newblock Mind the gap: Understanding the modality gap in multi-modal contrastive representation learning.
\newblock \emph{Advances in Neural Information Processing Systems (NeurIPS)}, 35:\penalty0 17612--17625, 2022.

\bibitem[Liu et~al.(2019)Liu, Li, Wang, and Tang]{liu2019spherical}
Kai Liu, Qiuwei Li, Hua Wang, and Gongguo Tang.
\newblock Spherical principal component analysis.
\newblock In \emph{Proceedings of the 2019 SIAM International Conference on Data Mining}, pp.\  387--395. SIAM, 2019.

\bibitem[Loshchilov \& Hutter(2019)Loshchilov and Hutter]{loshchilov2018decoupled}
Ilya Loshchilov and Frank Hutter.
\newblock Decoupled weight decay regularization.
\newblock In \emph{Proceedings of the Seventh International Conference on Learning Representations (ICLR)}, 2019.

\bibitem[Maharana et~al.(2025)Maharana, Zhang, Karlinsky, Feris, and Guo]{Maharana_2025_ICCV}
Sarthak Maharana, Baoming Zhang, Leonid Karlinsky, Rogerio Feris, and Yunhui Guo.
\newblock {BATCLIP}: Bimodal online test-time adaptation for {CLIP}.
\newblock In \emph{Proceedings of the IEEE/CVF International Conference on Computer Vision (ICCV)}, pp.\  1569--1579, October 2025.

\bibitem[Mintun et~al.(2021)Mintun, Kirillov, and Xie]{mintun2021on}
Eric Mintun, Alexander Kirillov, and Saining Xie.
\newblock On interaction between augmentations and corruptions in natural corruption robustness.
\newblock \emph{Advances in Neural Information Processing Systems (NeurIPS)}, pp.\  3571--3583, 2021.

\bibitem[Niu et~al.(2022)Niu, Wu, Zhang, Chen, Zheng, Zhao, and Tan]{niu2022efficient}
Shuaicheng Niu, Jiaxiang Wu, Yifan Zhang, Yaofo Chen, Shijian Zheng, Peilin Zhao, and Mingkui Tan.
\newblock Efficient test-time model adaptation without forgetting.
\newblock In \emph{Proceedings of the Thirty-Ninth International Conference on Machine Learning (ICML)}, pp.\  16888--16905. PMLR, 2022.

\bibitem[Niu et~al.(2023)Niu, Wu, Zhang, Wen, Chen, Zhao, and Tan]{niu2023towards}
Shuaicheng Niu, Jiaxiang Wu, Yifan Zhang, Zhiquan Wen, Yaofo Chen, Peilin Zhao, and Mingkui Tan.
\newblock Towards stable test-time adaptation in dynamic wild world.
\newblock In \emph{Proceedings of the Eleventh International Conference on Learning Representations (ICLR)}, 2023.

\bibitem[Osowiechi et~al.(2024)Osowiechi, Noori, Hakim, Yazdanpanah, Bahri, Cheraghalikhani, Dastani, Beizaee, Ayed, and Desrosiers]{osowiechi2024watt}
David Osowiechi, Mehrdad Noori, Gustavo~A Hakim, Moslem Yazdanpanah, Ali Bahri, Milad Cheraghalikhani, Sahar Dastani, Farzad Beizaee, Ismail~B Ayed, and Christian Desrosiers.
\newblock {WATT}: Weight average test time adaptation of {CLIP}.
\newblock \emph{Advances in Neural Information Processing Systems (NeurIPS)}, 37:\penalty0 48015--48044, 2024.

\bibitem[Patashnik et~al.(2021)Patashnik, Wu, Shechtman, Cohen-Or, and Lischinski]{patashnik2021styleclip}
Or~Patashnik, Zongze Wu, Eli Shechtman, Daniel Cohen-Or, and Dani Lischinski.
\newblock {StyleCLIP}: Text-driven manipulation of stylegan imagery.
\newblock In \emph{Proceedings of the IEEE/CVF International Conference on Computer Vision (ICCV)}, pp.\  2085--2094, 2021.

\bibitem[Peyr{\'e} \& Cuturi(2019)Peyr{\'e} and Cuturi]{peyre2019computational}
Gabriel Peyr{\'e} and Marco Cuturi.
\newblock Computational optimal transport: With applications to data science.
\newblock \emph{Foundations and Trends in Machine Learning}, 11\penalty0 (5-6):\penalty0 355--607, 2019.

\bibitem[Qian \& Hu(2024)Qian and Hu]{qian2024online}
Qi~Qian and Juhua Hu.
\newblock Online zero-shot classification with {CLIP}.
\newblock In \emph{Proceedings of the Eighteenth European Conference on Computer Vision (ECCV)}, pp.\  462--477, 2024.

\bibitem[Qian et~al.(2024)Qian, Xu, and Hu]{qian2024intra}
Qi~Qian, Yuanhong Xu, and Juhua Hu.
\newblock Intra-modal proxy learning for zero-shot visual categorization with {CLIP}.
\newblock \emph{Advances in Neural Information Processing Systems (NeurIPS)}, 36:\penalty0 25461--25474, 2024.

\bibitem[Qin et~al.(2022)Qin, Zhang, Chen, Lakshminarayanan, Beutel, and Wang]{qin2022understanding}
Yao Qin, Chiyuan Zhang, Ting Chen, Balaji Lakshminarayanan, Alex Beutel, and Xuezhi Wang.
\newblock Understanding and improving robustness of vision transformers through patch-based negative augmentation.
\newblock \emph{Advances in Neural Information Processing Systems (NeurIPS)}, 35:\penalty0 16276--16289, 2022.

\bibitem[Radford et~al.(2021)Radford, Kim, Hallacy, Ramesh, Goh, Agarwal, Sastry, Askell, Mishkin, Clark, Krueger, and Sutskever]{clip_paper}
Alec Radford, Jong~Wook Kim, Chris Hallacy, Aditya Ramesh, Gabriel Goh, Sandhini Agarwal, Girish Sastry, Amanda Askell, Pamela Mishkin, Jack Clark, Gretchen Krueger, and Ilya Sutskever.
\newblock Learning transferable visual models from natural language supervision.
\newblock In \emph{Proceedings of the Thirty-Eighth International Conference on Machine Learning (ICML)}, pp.\  8748--8763. PMLR, 2021.

\bibitem[Ramesh et~al.(2022)Ramesh, Dhariwal, Nichol, Chu, and Chen]{ramesh2022hierarchical}
Aditya Ramesh, Prafulla Dhariwal, Alex Nichol, Casey Chu, and Mark Chen.
\newblock Hierarchical text-conditional image generation with {CLIP} latents.
\newblock \emph{arXiv preprint arXiv:2204.06125}, 1\penalty0 (2):\penalty0 3, 2022.

\bibitem[Recht et~al.(2019)Recht, Roelofs, Schmidt, and Shankar]{recht2019imagenet}
Benjamin Recht, Rebecca Roelofs, Ludwig Schmidt, and Vaishaal Shankar.
\newblock Do {ImageNet} classifiers generalize to {ImageNet}?
\newblock In \emph{Proceedings of the Thirty-Sixth International Conference on Machine Learning (ICML)}, pp.\  5389--5400. PMLR, 2019.

\bibitem[Schuhmann et~al.(2022)Schuhmann, Beaumont, Vencu, Gordon, Wightman, Cherti, Coombes, Katta, Mullis, Wortsman, Schramowski, Kundurthy, Crowson, Schmidt, Kaczmarczyk, and Jitsev]{schuhmann2022laionb}
Christoph Schuhmann, Romain Beaumont, Richard Vencu, Cade~W Gordon, Ross Wightman, Mehdi Cherti, Theo Coombes, Aarush Katta, Clayton Mullis, Mitchell Wortsman, Patrick Schramowski, Srivatsa~R Kundurthy, Katherine Crowson, Ludwig Schmidt, Robert Kaczmarczyk, and Jenia Jitsev.
\newblock {LAION-5B}: An open large-scale dataset for training next generation image-text models.
\newblock In \emph{Thirty-Sixth Conference on Neural Information Processing Systems Datasets and Benchmarks Track}, pp.\  25278--25294, 2022.

\bibitem[Shi \& Sha(2012)Shi and Sha]{Shi2012InformationTheoreticalLO}
Yuan Shi and Fei Sha.
\newblock Information-theoretical learning of discriminative clusters for unsupervised domain adaptation.
\newblock In \emph{Proceedings of the Twenty-Ninth International Conference on Machine Learning (ICML)}, pp.\  1275--1282, 2012.

\bibitem[Shu et~al.(2022)Shu, Nie, Huang, Yu, Goldstein, Anandkumar, and Xiao]{shu2022tpt}
Manli Shu, Weili Nie, De-An Huang, Zhiding Yu, Tom Goldstein, Anima Anandkumar, and Chaowei Xiao.
\newblock Test-time prompt tuning for zero-shot generalization in vision-language models.
\newblock \emph{Advances in Neural Information Processing Systems (NeurIPS)}, 35:\penalty0 14274--14289, 2022.

\bibitem[S{\'o}jka et~al.(2023)S{\'o}jka, Cygert, Twardowski, and Trzci{\'n}ski]{sojka2023ar}
Damian S{\'o}jka, Sebastian Cygert, Bart{\l}omiej Twardowski, and Tomasz Trzci{\'n}ski.
\newblock {AR-TTA}: A simple method for real-world continual test-time adaptation.
\newblock In \emph{Proceedings of the IEEE/CVF International Conference on Computer Vision (ICCV)}, pp.\  3491--3495, 2023.

\bibitem[Sui et~al.(2025)Sui, Wang, and Yeung-Levy]{sui2025just}
Elaine Sui, Xiaohan Wang, and Serena Yeung-Levy.
\newblock Just shift it: Test-time prototype shifting for zero-shot generalization with vision-language models.
\newblock In \emph{Proceedings of the IEEE/CVF Winter Conference on Applications of Computer Vision (WACV)}, pp.\  825--835, 2025.

\bibitem[Vesdapunt et~al.(2024)Vesdapunt, Fu, Wu, Zhang, and Natarajan]{vesdapunt2024hvclip}
Noranart Vesdapunt, Kah~Kuen Fu, Yue Wu, Xu~Zhang, and Pradeep Natarajan.
\newblock {HVCLIP}: High-dimensional vector in {CLIP} for unsupervised domain adaptation.
\newblock In \emph{Proceedings of the Eighteenth European Conference on Computer Vision (ECCV)}, pp.\  36--54, 2024.

\bibitem[Villani(2008)]{villani2008optimal}
C{\'e}dric Villani.
\newblock \emph{Optimal transport: old and new}, volume 338.
\newblock Springer, 2008.

\bibitem[Volk et~al.(2019)Volk, Müller, Bernuth, Hospach, and Bringmann]{volk2019robust}
Georg Volk, Stefan Müller, Alexander~von Bernuth, Dennis Hospach, and Oliver Bringmann.
\newblock Towards robust {CNN}-based object detection through augmentation with synthetic rain variations.
\newblock In \emph{Proceedings of the 2019 IEEE Intelligent Transportation Systems Conference (ITSC)}, pp.\  285--292, 2019.

\bibitem[Wang et~al.(2021)Wang, Shelhamer, Liu, Olshausen, and Darrell]{Wang2021}
Dequan Wang, Evan Shelhamer, Shaoteng Liu, Bruno Olshausen, and Trevor Darrell.
\newblock Tent: Fully test-time adaptation by entropy minimization.
\newblock In \emph{Proceedings of the Ninth International Conference on Learning Representations (ICLR)}, 2021.

\bibitem[Wang et~al.(2019)Wang, Ge, Lipton, and Xing]{robust_global_representation_neurips2019}
Haohan Wang, Songwei Ge, Zachary Lipton, and Eric~P Xing.
\newblock Learning robust global representations by penalizing local predictive power.
\newblock \emph{Advances in Neural Information Processing Systems (NeurIPS)}, 32:\penalty0 10506--10518, 2019.

\bibitem[Wang et~al.(2022)Wang, Fink, Van~Gool, and Dai]{Wang_2022_CVPR}
Qin Wang, Olga Fink, Luc Van~Gool, and Dengxin Dai.
\newblock Continual test-time domain adaptation.
\newblock In \emph{Proceedings of the IEEE/CVF Conference on Computer Vision and Pattern Recognition (CVPR)}, pp.\  7201--7211, June 2022.

\bibitem[Wang et~al.(2023{\natexlab{a}})Wang, Zhang, Yan, Zhang, and Li]{wang2023feature}
Shuai Wang, Daoan Zhang, Zipei Yan, Jianguo Zhang, and Rui Li.
\newblock Feature alignment and uniformity for test time adaptation.
\newblock In \emph{Proceedings of the IEEE/CVF Conference on Computer Vision and Pattern Recognition (CVPR)}, pp.\  20050--20060, 2023{\natexlab{a}}.

\bibitem[Wang \& Isola(2020)Wang and Isola]{wang2020understanding}
Tongzhou Wang and Phillip Isola.
\newblock Understanding contrastive representation learning through alignment and uniformity on the hypersphere.
\newblock In \emph{Proceedings of the Thirty-Seventh International Conference on Machine Learning (ICML)}, pp.\  9929--9939. PMLR, 2020.

\bibitem[Wang et~al.(2025)Wang, Chen, Zhang, Chen, and Ma]{Wang_2025_CVPR}
Xin Wang, Kai Chen, Jiaming Zhang, Jingjing Chen, and Xingjun Ma.
\newblock {TAPT}: Test-time adversarial prompt tuning for robust inference in vision-language models.
\newblock In \emph{Proceedings of the IEEE/CVF Conference on Computer Vision and Pattern Recognition (CVPR)}, pp.\  19910--19920, June 2025.

\bibitem[Wang et~al.(2024{\natexlab{a}})Wang, Hong, Cheraghian, Rahman, Ahmedt-Aristizabal, Petersson, and Harandi]{Wang_2024_WACV}
Yanshuo Wang, Jie Hong, Ali Cheraghian, Shafin Rahman, David Ahmedt-Aristizabal, Lars Petersson, and Mehrtash Harandi.
\newblock Continual test-time domain adaptation via dynamic sample selection.
\newblock In \emph{Proceedings of the IEEE/CVF Winter Conference on Applications of Computer Vision (WACV)}, pp.\  1701--1710, January 2024{\natexlab{a}}.

\bibitem[Wang et~al.(2023{\natexlab{b}})Wang, Liang, He, Xu, Wang, and Tan]{wang2023improving}
Zhengbo Wang, Jian Liang, Ran He, Nan Xu, Zilei Wang, and Tieniu Tan.
\newblock Improving zero-shot generalization for {CLIP} with synthesized prompts.
\newblock In \emph{Proceedings of the IEEE/CVF International Conference on Computer Vision (ICCV)}, pp.\  3032--3042, 2023{\natexlab{b}}.

\bibitem[Wang et~al.(2024{\natexlab{b}})Wang, Liang, Sheng, He, Wang, and Tan]{wang2024a}
Zhengbo Wang, Jian Liang, Lijun Sheng, Ran He, Zilei Wang, and Tieniu Tan.
\newblock A hard-to-beat baseline for training-free {CLIP}-based adaptation.
\newblock In \emph{Proceedings of the Twelfth International Conference on Learning Representations (ICLR)}, 2024{\natexlab{b}}.

\bibitem[Yamaguchi et~al.(2025)Yamaguchi, Feng, Kanai, Adachi, and Chijiwa]{Yamaguchi_2025_CVPR}
Shin'ya Yamaguchi, Dewei Feng, Sekitoshi Kanai, Kazuki Adachi, and Daiki Chijiwa.
\newblock Post-pre-training for modality alignment in vision-language foundation models.
\newblock In \emph{Proceedings of the IEEE/CVF Conference on Computer Vision and Pattern Recognition (CVPR)}, pp.\  4256--4266, June 2025.

\bibitem[Yoon et~al.(2024)Yoon, Yoon, Tee, Hasegawa-Johnson, Li, and Yoo]{yoon2024c}
Hee~Suk Yoon, Eunseop Yoon, Joshua Tian~Jin Tee, Mark Hasegawa-Johnson, Yingzhen Li, and Chang~D Yoo.
\newblock {C-TPT}: Calibrated test-time prompt tuning for vision-language models via text feature dispersion.
\newblock In \emph{Proceedings of the Twelfth International Conference on Learning Representations (ICLR)}, 2024.

\bibitem[Yuan et~al.(2023)Yuan, Xie, and Li]{yuan2023robust}
Longhui Yuan, Binhui Xie, and Shuang Li.
\newblock Robust test-time adaptation in dynamic scenarios.
\newblock In \emph{Proceedings of the IEEE/CVF Conference on Computer Vision and Pattern Recognition (CVPR)}, pp.\  15922--15932, 2023.

\bibitem[Zanella \& Ben~Ayed(2024{\natexlab{a}})Zanella and Ben~Ayed]{Zanella_2024_CVPR}
Maxime Zanella and Ismail Ben~Ayed.
\newblock Low-rank few-shot adaptation of vision-language models.
\newblock In \emph{Proceedings of the IEEE/CVF Conference on Computer Vision and Pattern Recognition (CVPR) Workshops}, pp.\  1593--1603, June 2024{\natexlab{a}}.

\bibitem[Zanella \& Ben~Ayed(2024{\natexlab{b}})Zanella and Ben~Ayed]{zanella2024test}
Maxime Zanella and Ismail Ben~Ayed.
\newblock On the test-time zero-shot generalization of vision-language models: Do we really need prompt learning?
\newblock In \emph{Proceedings of the IEEE/CVF Conference on Computer Vision and Pattern Recognition (CVPR)}, pp.\  23783--23793, 2024{\natexlab{b}}.

\bibitem[Zanella et~al.(2025)Zanella, Fuchs, De~Vleeschouwer, and Ben~Ayed]{zanella2025realistic}
Maxime Zanella, Cl{\'e}ment Fuchs, Christophe De~Vleeschouwer, and Ismail Ben~Ayed.
\newblock Realistic test-time adaptation of vision-language models.
\newblock In \emph{Proceedings of the IEEE/CVF Conference on Computer Vision and Pattern Recognition (CVPR)}, pp.\  25103--25112, 2025.

\bibitem[Zhai et~al.(2023)Zhai, Mustafa, Kolesnikov, and Beyer]{zhai2023sigmoid}
Xiaohua Zhai, Basil Mustafa, Alexander Kolesnikov, and Lucas Beyer.
\newblock Sigmoid loss for language image pre-training.
\newblock In \emph{Proceedings of the IEEE/CVF International Conference on Computer Vision (ICCV)}, pp.\  11975--11986, 2023.

\bibitem[Zhang et~al.(2024{\natexlab{a}})Zhang, Stepputtis, Sycara, and Xie]{zhang2024dpe}
Ce~Zhang, Simon Stepputtis, Katia Sycara, and Yaqi Xie.
\newblock Dual prototype evolving for test-time generalization of vision-language models.
\newblock \emph{Advances in Neural Information Processing Systems (NeurIPS)}, 37:\penalty0 32111--32136, 2024{\natexlab{a}}.

\bibitem[Zhang et~al.(2022{\natexlab{a}})Zhang, Levine, and Finn]{zhang2022memo}
Marvin Zhang, Sergey Levine, and Chelsea Finn.
\newblock {MEMO}: Test time robustness via adaptation and augmentation.
\newblock \emph{Advances in Neural Information Processing Systems (NeurIPS)}, 35:\penalty0 38629--38642, 2022{\natexlab{a}}.

\bibitem[Zhang et~al.(2022{\natexlab{b}})Zhang, Fang, Zhang, Gao, Li, Dai, Qiao, and Li]{Zhang2022tip-adapter}
Renrui Zhang, Rongyao Fang, Wei Zhang, Peng Gao, Kunchang Li, Jifeng Dai, Yu~Qiao, and Hongsheng Li.
\newblock Tip-adapter: Training-free adaption of {CLIP} for few-shot classification.
\newblock In \emph{Proceedings of the Seventeenth European Conference on Computer Vision (ECCV)}, pp.\  493--510, 2022{\natexlab{b}}.

\bibitem[Zhang et~al.(2024{\natexlab{b}})Zhang, Zhu, Tang, Ma, Zhou, and Zhang]{zhang2024dual}
Yabin Zhang, Wenjie Zhu, Hui Tang, Zhiyuan Ma, Kaiyang Zhou, and Lei Zhang.
\newblock Dual memory networks: A versatile adaptation approach for vision-language models.
\newblock In \emph{Proceedings of the IEEE/CVF Conference on Computer Vision and Pattern Recognition (CVPR)}, pp.\  28718--28728, 2024{\natexlab{b}}.

\bibitem[Zhou \& Levine(2021)Zhou and Levine]{zhou2021bayesian}
Aurick Zhou and Sergey Levine.
\newblock Bayesian adaptation for covariate shift.
\newblock \emph{Advances in Neural Information Processing Systems (NeurIPS)}, 34:\penalty0 914--927, 2021.

\bibitem[Zhou et~al.(2022{\natexlab{a}})Zhou, Yang, Loy, and Liu]{zhou2022cocoop}
Kaiyang Zhou, Jingkang Yang, Chen~Change Loy, and Ziwei Liu.
\newblock Conditional prompt learning for vision-language models.
\newblock In \emph{Proceedings of the IEEE/CVF Conference on Computer Vision and Pattern Recognition (CVPR)}, pp.\  16816--16825, June 2022{\natexlab{a}}.

\bibitem[Zhou et~al.(2022{\natexlab{b}})Zhou, Yang, Loy, and Liu]{zhou2022coop}
Kaiyang Zhou, Jingkang Yang, Chen~Change Loy, and Ziwei Liu.
\newblock Learning to prompt for vision-language models.
\newblock \emph{International Journal of Computer Vision (IJCV)}, 130\penalty0 (9):\penalty0 2337--2348, 2022{\natexlab{b}}.

\bibitem[Zhou et~al.(2023)Zhou, Ren, Li, Zabih, and Lim]{zhou2024test}
Yifei Zhou, Juntao Ren, Fengyu Li, Ramin Zabih, and Ser~Nam Lim.
\newblock Test-time distribution normalization for contrastively learned visual-language models.
\newblock \emph{Advances in Neural Information Processing Systems (NeurIPS)}, 36:\penalty0 47105--47123, 2023.

\end{thebibliography}
\bibliographystyle{tmlr}

\newpage
\appendix
\section{\highlight{Broader Impact Concerns}}\label{asec:broader_impact}
\highlight{While UnInfo enables test-time adaptation for zero-shot VLM classification in a fully unsupervised manner, it may also adapt to biased data, which could affect model behavior, such as fairness.
  In sensitive applications, model behavior should be carefully monitored and appropriately intervened in as needed.}

\section{Details of UnInfo}\label{asec:details_uninfo}

% Algorithm =================

\renewcommand{\algorithmicrequire}{\textbf{Input:}}
\renewcommand{\algorithmicensure}{\textbf{Output:}}

\begin{algorithm}[tb]
  \caption{Procedure of UnInfo.}
  \label{alg:proposed_method}
  \begin{algorithmic}
    \REQUIRE Pre-trained image encoder $f^\text{img}_{\theta_\text{img}}$, class text embeddings $\{ \mathbf{t}_c \}_{c=1}^C$, initialized LoRA $\phi_\text{img}$, target dataset $\mathcal{D}$
    \ENSURE Adapted LoRA parameter (EMA) $\bar{\phi}_\text{img}$
    \FOR{mini-batch $\{ \mathbf{x}_i \}_{i=1}^B$ in $\mathcal{D}$}
    \STATE Compute image embeddings with the current LoRA $\{ \mathbf{z}_i=f^\text{img}_{\theta_\text{img},\phi_\text{img}}(\mathbf{x}_i) \}_{i=1}^B$.
    \STATE Compute zero-shot prediction probabilities $\{ \hat{p}_{i,c} \}_{i=1}^B$ in accordance with \cref{eq:clip_zeroshot_probability}.
    \STATE Compute teacher zero-shot prediction probabilities $\{\hat{q}_{i,c}\}_{i=1}^B$ using the EMA LoRA $\bar{\phi}_\text{img}$.
    \STATE Compute the loss in accordance with \cref{eq:proposed_objective}.
    \STATE Update $\phi_\text{img}$.
    \STATE Update EMA teacher $\bar{\phi}_\text{img}$ in accordance with \cref{eq:ema_update}.
    \ENDFOR
  \end{algorithmic}
\end{algorithm}

\cref{alg:proposed_method} lists the procedure of UnInfo.

\section{Text Prompt Ensemble}\label{asec:text_prompt_ensemble}
Here, we describe the details of the prompt ensemble in the preliminary experiment (\cref{sec:preliminary_experiment}).
We used template texts listed in \cref{atab:ensemble_prompt_list} for the ensemble of multiple templates (denoted by ``Ensemble'' in \cref{tab:preliminary_zs-accuracy_corruption}).
We generated the prompts using the templates for each class and encoded them with the text encoder.
Then, we calculated the mean of the text embeddings.
We normalized the embedding and used it for the class prototype.

For the ensemble of corruption synonyms (denoted by ``Corruption prompt'' in \cref{tab:preliminary_zs-accuracy_corruption}), we ensembled the corruption synonyms listed in \cref{atab:ensemble_corruption_list}, which are generated by GPT-4o~\citep{hurst2024gpt} with the instruction ``\texttt{You are an expert of image processing. List the synonyms of the word "[corruption name]," which represents image quality.}''

\begin{table}[tb]
  \centering
  \caption{Text prompt templates used for ensemble in the preliminary experiment (\cref{sec:preliminary_experiment}). These templates are proposed by~\citet{clip_paper}\protect\footnotemark.}
  \label{atab:ensemble_prompt_list}
  \resizebox{1.0\linewidth}{!}{
    \setlength{\tabcolsep}{3pt}
    \begin{tabular}{llll}\toprule
      \texttt{a bad photo of a \{\}.}              & \texttt{a photo of many \{\}.}                & \texttt{a sculpture of a \{\}.}             & \texttt{a photo of the hard to see \{\}.} \\
      \texttt{a low resolution photo of the \{\}.} & \texttt{a rendering of a \{\}.}               & \texttt{graffiti of a \{\}.}                & \texttt{a bad photo of the \{\}.}         \\
      \texttt{a cropped photo of the \{\}.}        & \texttt{a tattoo of a \{\}.}                  & \texttt{the embroidered \{\}.}              & \texttt{a photo of a hard to see \{\}.}   \\
      \texttt{a bright photo of a \{\}.}           & \texttt{a photo of a clean \{\}.}             & \texttt{a photo of a dirty \{\}.}           & \texttt{a dark photo of the \{\}.}        \\
      \texttt{a drawing of a \{\}.}                & \texttt{a photo of my \{\}.}                  & \texttt{the plastic \{\}.}                  & \texttt{a photo of the cool \{\}.}        \\
      \texttt{a close-up photo of a \{\}.}         & \texttt{a black and white photo of the \{\}.} & \texttt{a painting of the \{\}.}            & \texttt{a painting of a \{\}.}            \\
      \texttt{a pixelated photo of the \{\}.}      & \texttt{a sculpture of the \{\}.}             & \texttt{a bright photo of the \{\}.}        & \texttt{a cropped photo of a \{\}.}       \\
      \texttt{a plastic \{\}.}                     & \texttt{a photo of the dirty \{\}.}           & \texttt{a jpeg corrupted photo of a \{\}.}  & \texttt{a blurry photo of the \{\}.}      \\
      \texttt{a photo of the \{\}.}                & \texttt{a good photo of the \{\}.}            & \texttt{a rendering of the \{\}.}           & \texttt{a \{\} in a video game.}          \\
      \texttt{a photo of one \{\}.}                & \texttt{a doodle of a \{\}.}                  & \texttt{a close-up photo of the \{\}.}      & \texttt{a photo of a \{\}.}               \\
      \texttt{the origami \{\}.}                   & \texttt{the \{\} in a video game.}            & \texttt{a sketch of a \{\}.}                & \texttt{a doodle of the \{\}.}            \\
      \texttt{a origami \{\}.}                     & \texttt{a low resolution photo of a \{\}.}    & \texttt{the toy \{\}.}                      & \texttt{a rendition of the \{\}.}         \\
      \texttt{a photo of the clean \{\}.}          & \texttt{a photo of a large \{\}.}             & \texttt{a rendition of a \{\}.}             & \texttt{a photo of a nice \{\}.}          \\
      \texttt{a photo of a weird \{\}.}            & \texttt{a blurry photo of a \{\}.}            & \texttt{a cartoon \{\}.}                    & \texttt{art of a \{\}.}                   \\
      \texttt{a sketch of the \{\}.}               & \texttt{a embroidered \{\}.}                  & \texttt{a pixelated photo of a \{\}.}       & \texttt{itap of the \{\}.}                \\
      \texttt{a jpeg corrupted photo of the \{\}.} & \texttt{a good photo of a \{\}.}              & \texttt{a plushie \{\}.}                    & \texttt{a photo of the nice \{\}.}        \\
      \texttt{a photo of the small \{\}.}          & \texttt{a photo of the weird \{\}.}           & \texttt{the cartoon \{\}.}                  & \texttt{art of the \{\}.}                 \\
      \texttt{a drawing of the \{\}.}              & \texttt{a photo of the large \{\}.}           & \texttt{a black and white photo of a \{\}.} & \texttt{the plushie \{\}.}                \\
      \texttt{a dark photo of a \{\}.}             & \texttt{itap of a \{\}.}                      & \texttt{graffiti of the \{\}.}              & \texttt{a toy \{\}.}                      \\
      \texttt{itap of my \{\}.}                    & \texttt{a photo of a cool \{\}.}              & \texttt{a photo of a small \{\}.}           & \texttt{a tattoo of the \{\}.}            \\ \bottomrule
    \end{tabular}
  }
\end{table}

\begin{table}[tb]
  \centering
  \caption{Synonyms of the corruption names of ImageNet-C used in the preliminary experiment in \cref{sec:preliminary_experiment}.}
  \label{atab:ensemble_corruption_list}
  \resizebox{1.0\linewidth}{!}{
    \setlength{\tabcolsep}{3.5pt}
    \begin{tabular}{lllll}\toprule
      Defocus blur                 & Glass blur                        & Motion blur                    & Zoom blur                          & Contrast                       \\ \cmidrule(lr){1-1} \cmidrule(lr){2-2} \cmidrule(lr){3-3} \cmidrule(lr){4-4} \cmidrule(lr){5-5}
      \texttt{defocus blur}        & \texttt{glass blur}               & \texttt{motion blur}           & \texttt{zoom blur}                 & \texttt{contrast}              \\
      \texttt{out-of-focus blur}   & \texttt{frosted blur}             & \texttt{directional blur}      & \texttt{radial blur}               & \texttt{tonal contrast}        \\
      \texttt{soft focus}          & \texttt{glazing blur}             & \texttt{linear blur}           & \texttt{zooming effect}            & \texttt{brightness difference} \\
      \texttt{bokeh}               & \texttt{diffuse blur}             & \texttt{dynamic blur}          & \texttt{dynamic zoom blur}         & \texttt{clarity}               \\
      \texttt{lens blur}           & \texttt{smudged blur}             & \texttt{streaking}             & \texttt{burst blur}                & \texttt{definition}            \\
      \texttt{gaussian blur}       & \texttt{hazy blur}                & \texttt{trail blur}            & \texttt{focus expansion blur}      & \texttt{distinction}           \\
      \texttt{depth blur}          & \texttt{translucent blur}         & \texttt{speed blur}            & \texttt{depth blur}                & \texttt{sharpness}             \\
      \texttt{background blur}     & \texttt{refractive blur}          & \texttt{panning blur}          & \texttt{lens zoom blur}            & \texttt{intensity diffenrece}  \\
      \texttt{field blur}          & \texttt{distortion blur}          & \texttt{motion streak}         & \texttt{outward motion blur}       & \texttt{dynamic range}         \\
      \texttt{focus softness}      & \texttt{veiled blur}              & \texttt{kinetic blur}          & \texttt{radian streak blur}        & \texttt{separation}            \\ \midrule
      %=============================
      Elastic transform            & Jpeg compression                  & Pixelate                       & Gaussian noise                     & Impulse noise                  \\\cmidrule(lr){1-1} \cmidrule(lr){2-2} \cmidrule(lr){3-3} \cmidrule(lr){4-4} \cmidrule(lr){5-5}
      \texttt{elastic transform}   & \texttt{jpeg compression}         & \texttt{pixelate}              & \texttt{Gaussian noise}            & \texttt{impulse noise}         \\
      \texttt{warping}             & \texttt{image compression}        & \texttt{blockify}              & \texttt{normal noise}              & \texttt{salt-and-pepper noise} \\
      \texttt{distortion}          & \texttt{lossy compression}        & \texttt{rasterize}             & \texttt{additive noise}            & \texttt{spiky noise}           \\
      \texttt{deformation}         & \texttt{JPEG encoding}            & \texttt{mosaic}                & \texttt{white Gaussian noise}      & \texttt{outlier noise}         \\
      \texttt{stretching}          & \texttt{file compression}         & \texttt{chunkify}              & \texttt{statistical noise}         & \texttt{random noise}          \\
      \texttt{bending}             & \texttt{quantization artifacting} & \texttt{grid effect}           & \texttt{random noise}              & \texttt{shot noise}            \\
      \texttt{geometric transform} & \texttt{data compression}         & \texttt{quantization}          & \texttt{luminance noise}           & \texttt{transitional noise}    \\
      \texttt{morphing}            & \texttt{image encoding}           & \texttt{low-resolution effect} & \texttt{stochastic interference}   & \texttt{burst noise}           \\
      \texttt{image warping}       & \texttt{compression artifacts}    & \texttt{bitmapping}            & \texttt{signal perturbation}       & \texttt{pulsed noise}          \\
      \texttt{spatial transform}   & \texttt{JPEG artifacts}           & \texttt{aliased effect}        & \texttt{normal distribution noise} & \texttt{point noise}           \\ \midrule
      %============================
      Shot noise                   & Brightness                        & Fog                            & Frost                              & Snow                           \\\cmidrule(lr){1-1} \cmidrule(lr){2-2} \cmidrule(lr){3-3} \cmidrule(lr){4-4} \cmidrule(lr){5-5}
      \texttt{shot noise}          & \texttt{brightness}               & \texttt{fog}                   & \texttt{frost}                     & \texttt{snow}                  \\
      \texttt{photon noise}        & \texttt{luminance}                & \texttt{haze}                  & \texttt{frosting}                  & \texttt{noise}                 \\
      \texttt{Poisson noise}       & \texttt{illumination}             & \texttt{mist}                  & \texttt{glare}                     & \texttt{grain}                 \\
      \texttt{quantum noise}       & \texttt{lightness}                & \texttt{obscuration}           & \texttt{haze}                      & \texttt{salt-and-pepper noise} \\
      \texttt{statistical noise}   & \texttt{intensity}                & \texttt{cloudiness}            & \texttt{mist}                      & \texttt{static}                \\
      \texttt{random noise}        & \texttt{radiance}                 & \texttt{smog}                  & \texttt{veiling}                   & \texttt{visual noise}          \\
      \texttt{electronic noise}    & \texttt{glow}                     & \texttt{blur}                  & \texttt{soft-focus}                & \texttt{pixel noise}           \\
      \texttt{counting noise}      & \texttt{shininess}                & \texttt{glare}                 & \texttt{diffusion}                 & \texttt{random noise}          \\
      \texttt{current noise}       & \texttt{exposure}                 & \texttt{veiling}               & \texttt{cloudiness}                & \texttt{white noise}           \\
      \texttt{flicker noise}       & \texttt{highlighting}             & \texttt{dimming}               & \texttt{blur}                      & \texttt{dither}                \\ \bottomrule
    \end{tabular}
  }
\end{table}

\footnotetext{\url{https://github.com/openai/CLIP/blob/main/notebooks/Prompt_Engineering_for_ImageNet.ipynb}}

\section{Earth Mover's Distance}\label{asec:emd_details}
In \cref{sec:preliminary_experiment}, we evaluated the modality gap between image and text embeddings using the earth mover's distance (EMD).
Here, we describe the details of EMD~\citep{villani2008optimal,peyre2019computational}.

EMD is computed as the minimum cost of transporting image embeddings $\{ \mathbf{z}_1,\ldots, \mathbf{z}_N \}$ to match text embeddings $\{ \mathbf{t}_1, \ldots, \mathbf{t}_C \}$.
More formally, we computed EMD between two distributions $p(\mathbf{z})=(1/N)\sum_{n=1}^N \delta(\mathbf{z}-\mathbf{z}_n)$ and $q(\mathbf{z})=(1/C)\sum_{c=1}^C \delta(\mathbf{z}-\mathbf{t}_c)$ over $\mathbb{R}^d$, where $\delta(\cdot)$ is the Dirac function.
Specifically, since the embedding vectors are normalized and on the unit hypersphere, we adopted the spherical distance $d(\mathbf{z}_n, \mathbf{t}_c):=\arccos (\mathbf{z}_n^\top \mathbf{t}_c)$ for the transportation cost.
We implemented the computation with the Python Optimal Transport (POT) library~\citep{flamary2021pot,flamary2024pot}.

\section{Additional Experimental Results}\label{asec:additional_results}

\subsection{Adaptation Performance}\label{assec:adaptation_performance}

\paragraph{Other CLIP architectures.}
\begin{table}[tb]
    \centering
    \caption{\highlight{Test accuracy (\%) of OpenAI ViT-B/32 on ImageNet-C. The numbers (1), (5), and (10) presented with the method names are the shot numbers per class $n$ used for the few-shot adaptation methods.}}
    \label{tab:openai-vit-b-32_imagenet-c_accuracy}
    \resizebox{1\linewidth}{!}{
        \setlength{\tabcolsep}{3pt}
        \begin{tabular}{lllllllllllllllll} \toprule
            Method                                         & \rotatebox{60}{\shortstack{Defocus\\blur}} & \rotatebox{60}{\shortstack{Glass\\blur}} & \rotatebox{60}{\shortstack{Motion\\blur}} & \rotatebox{60}{\shortstack{Zoom\\blur}} & \rotatebox{60}{\shortstack{Contrast}} & \rotatebox{60}{\shortstack{Elastic\\transform}} & \rotatebox{60}{\shortstack{Jpeg\\compression}} & \rotatebox{60}{\shortstack{Pixelate}} & \rotatebox{60}{\shortstack{Gaussian\\noise}} & \rotatebox{60}{\shortstack{Impulse\\noise}} & \rotatebox{60}{\shortstack{Shot\\noise}} & \rotatebox{60}{\shortstack{Brightness}} & \rotatebox{60}{\shortstack{Fog}}   & \rotatebox{60}{\shortstack{Frost}} & \rotatebox{60}{\shortstack{Snow}}  & Mean                               \\ \midrule
            No-adapt                                       & ${22.31}$                                  & ${11.39}$                                & ${20.07}$                                 & ${17.50}$                               & ${17.30}$                             & ${18.13}$                                       & ${29.36}$                                      & ${30.24}$                             & ${13.01}$                                    & ${13.54}$                                   & ${13.14}$                                & ${48.38}$                               & ${28.40}$                          & ${24.84}$                          & ${23.81}$                          & ${22.09}$                          \\ \midrule
            Linear probing (1)                             & ${{7.34}_{\pm 0.16}}$                      & ${{3.98}_{\pm 0.04}}$                    & ${{6.93}_{\pm 0.05}}$                     & ${{6.25}_{\pm 0.09}}$                   & ${{5.60}_{\pm 0.02}}$                 & ${{7.53}_{\pm 0.33}}$                           & ${{10.39}_{\pm 0.07}}$                         & ${{11.38}_{\pm 0.31}}$                & ${{5.05}_{\pm 0.03}}$                        & ${{5.03}_{\pm 0.09}}$                       & ${{5.24}_{\pm 0.06}}$                    & ${{20.45}_{\pm 0.30}}$                  & ${{10.80}_{\pm 0.35}}$             & ${{8.10}_{\pm 0.13}}$              & ${{7.59}_{\pm 0.10}}$              & ${{8.11}_{\pm 0.07}}$              \\
            Linear probing (5)                             & ${{12.27}_{\pm 0.18}}$                     & ${{6.91}_{\pm 0.23}}$                    & ${{11.11}_{\pm 0.05}}$                    & ${{9.82}_{\pm 0.20}}$                   & ${{8.93}_{\pm 0.11}}$                 & ${{12.29}_{\pm 0.21}}$                          & ${{16.34}_{\pm 0.13}}$                         & ${{18.54}_{\pm 0.24}}$                & ${{7.63}_{\pm 0.06}}$                        & ${{7.88}_{\pm 0.20}}$                       & ${{8.22}_{\pm 0.15}}$                    & ${{31.84}_{\pm 0.08}}$                  & ${{17.36}_{\pm 0.27}}$             & ${{13.49}_{\pm 0.11}}$             & ${{12.55}_{\pm 0.53}}$             & ${{13.01}_{\pm 0.02}}$             \\
            Linear probing (10)                            & ${{14.52}_{\pm 0.03}}$                     & ${{8.41}_{\pm 0.20}}$                    & ${{13.08}_{\pm 0.11}}$                    & ${{11.73}_{\pm 0.21}}$                  & ${{10.55}_{\pm 0.28}}$                & ${{15.38}_{\pm 0.54}}$                          & ${{19.38}_{\pm 0.24}}$                         & ${{22.34}_{\pm 0.20}}$                & ${{9.55}_{\pm 0.36}}$                        & ${{9.44}_{\pm 0.29}}$                       & ${{9.72}_{\pm 0.14}}$                    & ${{37.52}_{\pm 0.32}}$                  & ${{20.71}_{\pm 0.37}}$             & ${{16.38}_{\pm 0.14}}$             & ${{15.21}_{\pm 0.28}}$             & ${{15.59}_{\pm 0.08}}$             \\
            Tip-adapter~\citep{Zhang2022tip-adapter} (1)   & ${{10.40}_{\pm 0.16}}$                     & ${{5.67}_{\pm 0.10}}$                    & ${{9.49}_{\pm 0.08}}$                     & ${{8.94}_{\pm 0.07}}$                   & ${{7.45}_{\pm 0.03}}$                 & ${{11.22}_{\pm 0.14}}$                          & ${{15.07}_{\pm 0.15}}$                         & ${{16.36}_{\pm 0.24}}$                & ${{6.53}_{\pm 0.16}}$                        & ${{6.73}_{\pm 0.10}}$                       & ${{6.97}_{\pm 0.17}}$                    & ${{29.89}_{\pm 0.18}}$                  & ${{15.36}_{\pm 0.20}}$             & ${{11.98}_{\pm 0.17}}$             & ${{10.97}_{\pm 0.26}}$             & ${{11.53}_{\pm 0.07}}$             \\
            Tip-adapter (5)                                & ${{14.07}_{\pm 0.10}}$                     & ${{7.94}_{\pm 0.07}}$                    & ${{12.34}_{\pm 0.12}}$                    & ${{11.24}_{\pm 0.13}}$                  & ${{9.39}_{\pm 0.11}}$                 & ${{14.91}_{\pm 0.28}}$                          & ${{18.77}_{\pm 0.39}}$                         & ${{21.68}_{\pm 0.25}}$                & ${{8.72}_{\pm 0.08}}$                        & ${{9.08}_{\pm 0.21}}$                       & ${{9.12}_{\pm 0.26}}$                    & ${{36.48}_{\pm 0.21}}$                  & ${{19.45}_{\pm 0.21}}$             & ${{15.83}_{\pm 0.09}}$             & ${{14.00}_{\pm 0.23}}$             & ${{14.87}_{\pm 0.05}}$             \\
            Tip-adapter (10)                               & ${{16.14}_{\pm 0.25}}$                     & ${{9.40}_{\pm 0.27}}$                    & ${{14.30}_{\pm 0.18}}$                    & ${{12.69}_{\pm 0.29}}$                  & ${{11.23}_{\pm 0.02}}$                & ${{17.21}_{\pm 0.26}}$                          & ${{21.25}_{\pm 0.20}}$                         & ${{24.76}_{\pm 0.18}}$                & ${{9.92}_{\pm 0.04}}$                        & ${{10.47}_{\pm 0.12}}$                      & ${{10.48}_{\pm 0.25}}$                   & ${{40.20}_{\pm 0.28}}$                  & ${{22.15}_{\pm 0.23}}$             & ${{18.04}_{\pm 0.20}}$             & ${{15.93}_{\pm 0.05}}$             & ${{16.94}_{\pm 0.02}}$             \\ \midrule
            TPT~\citep{shu2022tpt}                         & ${{23.29}_{\pm 0.06}}$                     & ${{12.16}_{\pm 0.03}}$                   & ${{21.01}_{\pm 0.03}}$                    & ${{19.62}_{\pm 0.04}}$                  & ${{18.62}_{\pm 0.02}}$                & ${{19.50}_{\pm 0.08}}$                          & ${{31.74}_{\pm 0.09}}$                         & ${{32.57}_{\pm 0.08}}$                & ${{12.81}_{\pm 0.05}}$                       & ${{13.24}_{\pm 0.02}}$                      & ${{13.13}_{\pm 0.03}}$                   & ${{50.47}_{\pm 0.03}}$                  & ${{30.06}_{\pm 0.04}}$             & ${{26.89}_{\pm 0.09}}$             & ${{25.68}_{\pm 0.02}}$             & ${{23.39}_{\pm 0.02}}$             \\
            ZERO~\citep{farina2024frustratingly}           & ${{21.48}_{\pm 0.04}}$                     & ${{10.17}_{\pm 0.03}}$                   & ${{18.93}_{\pm 0.02}}$                    & ${{20.12}_{\pm 0.08}}$                  & ${{14.33}_{\pm 0.03}}$                & ${{17.27}_{\pm 0.03}}$                          & ${{30.04}_{\pm 0.05}}$                         & ${{31.22}_{\pm 0.04}}$                & ${{9.19}_{\pm 0.05}}$                        & ${{9.80}_{\pm 0.01}}$                       & ${{9.31}_{\pm 0.06}}$                    & ${{48.44}_{\pm 0.07}}$                  & ${{29.21}_{\pm 0.03}}$             & ${{25.33}_{\pm 0.08}}$             & ${{25.02}_{\pm 0.03}}$             & ${{21.32}_{\pm 0.00}}$             \\
            MTA~\citep{zanella2024test}                    & ${{20.76}_{\pm 0.05}}$                     & ${{11.19}_{\pm 0.06}}$                   & ${{18.60}_{\pm 0.06}}$                    & ${{18.56}_{\pm 0.04}}$                  & $\underline{{{19.42}_{\pm 0.07}}}$    & ${{19.29}_{\pm 0.03}}$                          & ${{30.25}_{\pm 0.11}}$                         & ${{32.09}_{\pm 0.05}}$                & ${{10.86}_{\pm 0.01}}$                       & ${{11.40}_{\pm 0.04}}$                      & ${{11.43}_{\pm 0.02}}$                   & ${{49.33}_{\pm 0.06}}$                  & ${{29.33}_{\pm 0.06}}$             & ${{26.22}_{\pm 0.05}}$             & ${{24.65}_{\pm 0.08}}$             & ${{22.23}_{\pm 0.02}}$             \\
            TDA~\citep{karmanov2024efficient}              & $\underline{{{23.93}_{\pm 0.08}}}$         & ${{12.97}_{\pm 0.06}}$                   & ${{22.29}_{\pm 0.03}}$                    & ${{19.49}_{\pm 0.04}}$                  & ${{18.22}_{\pm 0.05}}$                & ${{21.10}_{\pm 0.08}}$                          & ${{31.41}_{\pm 0.02}}$                         & ${{32.89}_{\pm 0.06}}$                & ${{14.65}_{\pm 0.04}}$                       & ${{15.32}_{\pm 0.02}}$                      & ${{15.24}_{\pm 0.05}}$                   & ${{{50.49}_{\pm 0.04}}}$                & ${{31.66}_{\pm 0.13}}$             & $\underline{{{27.33}_{\pm 0.03}}}$ & $\underline{{{26.02}_{\pm 0.14}}}$ & ${{24.20}_{\pm 0.02}}$             \\
            \highlight{DMN~\citep{zhang2024dual}}          & ${{23.30}_{\pm 0.00}}$                     & ${{12.02}_{\pm 0.00}}$                   & ${{21.43}_{\pm 0.00}}$                    & ${{18.85}_{\pm 0.00}}$                  & ${{16.41}_{\pm 0.00}}$                & ${{18.93}_{\pm 0.00}}$                          & ${{30.75}_{\pm 0.00}}$                         & ${{31.67}_{\pm 0.00}}$                & ${{13.83}_{\pm 0.00}}$                       & ${{14.53}_{\pm 0.00}}$                      & ${{14.13}_{\pm 0.00}}$                   & $\underline{{50.84}_{\pm 0.00}}$        & ${{29.28}_{\pm 0.00}}$             & ${{26.24}_{\pm 0.00}}$             & ${{24.89}_{\pm 0.00}}$             & ${{23.14}_{\pm 0.00}}$             \\
            \highlight{Mint~\citep{bao2026mint}}           & ${{19.89}_{\pm 0.16}}$                     & $\mathbf{{{20.67}_{\pm 0.08}}}$          & $\underline{{{23.57}_{\pm 0.17}}}$        & $\mathbf{{{21.59}_{\pm 0.13}}}$         & ${{15.99}_{\pm 0.03}}$                & $\mathbf{{{25.99}_{\pm 0.09}}}$                 & $\mathbf{{{36.56}_{\pm 0.04}}}$                & $\mathbf{{{36.88}_{\pm 0.09}}}$       & $\mathbf{{{19.39}_{\pm 0.10}}}$              & $\mathbf{{{19.00}_{\pm 0.12}}}$             & $\mathbf{{{19.88}_{\pm 0.11}}}$          & ${{45.06}_{\pm 0.16}}$                  & $\mathbf{{{36.02}_{\pm 0.11}}}$    & ${{21.18}_{\pm 0.06}}$             & ${{22.88}_{\pm 0.10}}$             & $\underline{{{25.64}_{\pm 0.03}}}$ \\
            \highlight{BATCLIP~\citep{Maharana_2025_ICCV}} & ${{6.47}_{\pm 0.19}}$                      & ${{9.42}_{\pm 0.61}}$                    & ${{18.08}_{\pm 0.72}}$                    & ${{19.42}_{\pm 0.43}}$                  & ${{3.74}_{\pm 0.36}}$                 & $\underline{{{24.24}_{\pm 0.32}}}$              & ${{28.11}_{\pm 1.36}}$                         & ${{26.54}_{\pm 0.62}}$                & ${{9.39}_{\pm 0.05}}$                        & ${{8.18}_{\pm 0.39}}$                       & ${{8.77}_{\pm 0.09}}$                    & ${{43.25}_{\pm 0.33}}$                  & ${{29.35}_{\pm 0.09}}$             & ${{14.91}_{\pm 0.49}}$             & ${{20.49}_{\pm 0.21}}$             & ${{18.03}_{\pm 0.25}}$             \\
            UnInfo (ours)                                  & $\mathbf{{{26.80}_{\pm 0.28}}}$            & $\underline{{{14.77}_{\pm 1.00}}}$       & $\mathbf{{{26.10}_{\pm 0.74}}}$           & $\underline{{{21.41}_{\pm 0.57}}}$      & $\mathbf{{{21.51}_{\pm 0.27}}}$       & ${{20.94}_{\pm 0.71}}$                          & $\underline{{{34.73}_{\pm 0.10}}}$             & $\underline{{{36.87}_{\pm 0.50}}}$    & $\underline{{{16.71}_{\pm 0.15}}}$           & $\underline{{{18.88}_{\pm 0.64}}}$          & $\underline{{{17.14}_{\pm 0.47}}}$       & $\mathbf{{{51.16}_{\pm 0.25}}}$         & $\underline{{{33.09}_{\pm 0.44}}}$ & $\mathbf{{{27.44}_{\pm 0.13}}}$    & $\mathbf{{{27.39}_{\pm 0.34}}}$    & $\mathbf{{{26.33}_{\pm 0.04}}}$    \\ \bottomrule
        \end{tabular}
    }
\end{table}

\begin{table}[tb]
    \centering
    \caption{\highlight{Test accuracy (\%) of SigLIP on ImageNet-C. The numbers (1), (5), and (10) presented with the method names are the shot numbers per class $n$ used for the few-shot adaptation methods.}}
    \label{tab:siglip_imagenet-c_accuracy}
    \resizebox{1\linewidth}{!}{
        \setlength{\tabcolsep}{3pt}
        \begin{tabular}{lllllllllllllllll} \toprule
            Method                                         & \rotatebox{60}{\shortstack{Defocus\\blur}} & \rotatebox{60}{\shortstack{Glass\\blur}} & \rotatebox{60}{\shortstack{Motion\\blur}} & \rotatebox{60}{\shortstack{Zoom\\blur}} & \rotatebox{60}{\shortstack{Contrast}} & \rotatebox{60}{\shortstack{Elastic\\transform}} & \rotatebox{60}{\shortstack{Jpeg\\compression}} & \rotatebox{60}{\shortstack{Pixelate}} & \rotatebox{60}{\shortstack{Gaussian\\noise}} & \rotatebox{60}{\shortstack{Impulse\\noise}} & \rotatebox{60}{\shortstack{Shot\\noise}} & \rotatebox{60}{\shortstack{Brightness}} & \rotatebox{60}{\shortstack{Fog}}   & \rotatebox{60}{\shortstack{Frost}} & \rotatebox{60}{\shortstack{Snow}}  & Mean                               \\ \midrule
            No-adapt                                       & ${26.49}$                                  & ${12.33}$                                & ${16.60}$                                 & ${20.38}$                               & ${16.48}$                             & ${16.09}$                                       & ${36.61}$                                      & ${45.29}$                             & ${6.18}$                                     & ${8.09}$                                    & ${8.03}$                                 & ${62.21}$                               & ${42.49}$                          & ${32.75}$                          & ${31.94}$                          & ${25.46}$                          \\ \midrule
            Linear probing (1)                             & ${{9.79}_{\pm 1.48}}$                      & ${{5.80}_{\pm 0.10}}$                    & ${{6.56}_{\pm 1.32}}$                     & ${{9.75}_{\pm 0.23}}$                   & ${{6.76}_{\pm 0.26}}$                 & ${{8.33}_{\pm 0.20}}$                           & ${{14.46}_{\pm 0.24}}$                         & ${{20.78}_{\pm 0.41}}$                & ${{3.13}_{\pm 0.29}}$                        & ${{3.23}_{\pm 0.03}}$                       & ${{3.34}_{\pm 0.14}}$                    & ${{31.11}_{\pm 2.40}}$                  & ${{19.56}_{\pm 0.34}}$             & ${{10.68}_{\pm 0.23}}$             & ${{13.48}_{\pm 0.38}}$             & ${{11.12}_{\pm 0.19}}$             \\
            Linear probing (5)                             & ${{18.15}_{\pm 0.34}}$                     & ${{9.83}_{\pm 0.17}}$                    & ${{12.28}_{\pm 0.17}}$                    & ${{15.72}_{\pm 0.28}}$                  & ${{10.50}_{\pm 0.18}}$                & ${{14.17}_{\pm 0.21}}$                          & ${{23.23}_{\pm 0.08}}$                         & ${{32.55}_{\pm 0.03}}$                & ${{1.61}_{\pm 2.13}}$                        & ${{0.10}_{\pm 0.00}}$                       & ${{2.26}_{\pm 3.06}}$                    & ${{49.34}_{\pm 0.13}}$                  & ${{30.55}_{\pm 0.26}}$             & ${{18.31}_{\pm 0.38}}$             & ${{21.20}_{\pm 0.35}}$             & ${{17.32}_{\pm 0.21}}$             \\
            Linear probing (10)                            & ${{21.64}_{\pm 0.05}}$                     & ${{12.08}_{\pm 0.12}}$                   & ${{14.54}_{\pm 0.23}}$                    & ${{18.97}_{\pm 0.28}}$                  & ${{12.62}_{\pm 0.26}}$                & ${{17.40}_{\pm 0.36}}$                          & ${{27.96}_{\pm 0.15}}$                         & ${{37.78}_{\pm 0.37}}$                & ${{0.10}_{\pm 0.00}}$                        & ${{0.10}_{\pm 0.00}}$                       & ${{2.85}_{\pm 3.88}}$                    & ${{55.23}_{\pm 0.10}}$                  & ${{35.82}_{\pm 0.43}}$             & ${{22.93}_{\pm 0.23}}$             & ${{24.72}_{\pm 0.04}}$             & ${{20.31}_{\pm 0.29}}$             \\
            Tip-adapter~\citep{Zhang2022tip-adapter} (1)   & ${{15.16}_{\pm 0.24}}$                     & ${{7.86}_{\pm 0.17}}$                    & ${{10.10}_{\pm 0.27}}$                    & ${{13.73}_{\pm 0.24}}$                  & ${{9.25}_{\pm 0.19}}$                 & ${{11.93}_{\pm 0.15}}$                          & ${{21.88}_{\pm 0.23}}$                         & ${{29.03}_{\pm 0.37}}$                & ${{4.49}_{\pm 0.07}}$                        & ${{5.51}_{\pm 0.13}}$                       & ${{5.53}_{\pm 0.09}}$                    & ${{44.67}_{\pm 0.12}}$                  & ${{27.45}_{\pm 0.41}}$             & ${{16.71}_{\pm 0.20}}$             & ${{18.62}_{\pm 0.11}}$             & ${{16.13}_{\pm 0.10}}$             \\
            Tip-adapter (5)                                & ${{20.04}_{\pm 0.13}}$                     & ${{10.99}_{\pm 0.16}}$                   & ${{13.81}_{\pm 0.18}}$                    & ${{17.90}_{\pm 0.17}}$                  & ${{11.88}_{\pm 0.06}}$                & ${{16.26}_{\pm 0.16}}$                          & ${{26.36}_{\pm 0.24}}$                         & ${{36.26}_{\pm 0.41}}$                & ${{5.72}_{\pm 0.18}}$                        & ${{7.01}_{\pm 0.04}}$                       & ${{6.95}_{\pm 0.07}}$                    & ${{52.24}_{\pm 0.12}}$                  & ${{33.39}_{\pm 0.20}}$             & ${{20.28}_{\pm 0.46}}$             & ${{23.22}_{\pm 0.10}}$             & ${{20.15}_{\pm 0.02}}$             \\
            Tip-adapter (10)                               & ${{22.86}_{\pm 0.14}}$                     & ${{12.91}_{\pm 0.14}}$                   & ${{15.80}_{\pm 0.28}}$                    & ${{20.27}_{\pm 0.22}}$                  & ${{13.44}_{\pm 0.14}}$                & ${{18.60}_{\pm 0.08}}$                          & ${{29.41}_{\pm 0.20}}$                         & ${{39.62}_{\pm 0.15}}$                & ${{6.47}_{\pm 0.08}}$                        & ${{7.88}_{\pm 0.18}}$                       & ${{8.06}_{\pm 0.16}}$                    & ${{55.65}_{\pm 0.16}}$                  & ${{36.67}_{\pm 0.29}}$             & ${{23.37}_{\pm 0.16}}$             & ${{25.94}_{\pm 0.19}}$             & ${{22.46}_{\pm 0.05}}$             \\ \midrule
            TPT~\citep{shu2022tpt}                         & ${{11.32}_{\pm 0.01}}$                     & ${{4.87}_{\pm 0.00}}$                    & ${{6.56}_{\pm 0.02}}$                     & ${{8.94}_{\pm 0.01}}$                   & $\underline{{{22.77}_{\pm 0.02}}}$    & ${{2.68}_{\pm 0.01}}$                           & ${{9.08}_{\pm 0.01}}$                          & ${{18.02}_{\pm 0.01}}$                & ${{0.90}_{\pm 0.00}}$                        & ${{0.97}_{\pm 0.00}}$                       & ${{0.88}_{\pm 0.00}}$                    & ${{2.81}_{\pm 0.02}}$                   & ${{33.48}_{\pm 0.02}}$             & ${{3.44}_{\pm 0.01}}$              & ${{0.44}_{\pm 0.00}}$              & ${{8.48}_{\pm 0.00}}$              \\
            ZERO~\citep{farina2024frustratingly}           & ${{10.20}_{\pm 0.00}}$                     & ${{3.74}_{\pm 0.05}}$                    & ${{5.63}_{\pm 0.01}}$                     & ${{9.09}_{\pm 0.03}}$                   & ${{21.11}_{\pm 0.04}}$                & ${{3.65}_{\pm 0.00}}$                           & ${{8.10}_{\pm 0.03}}$                          & ${{19.41}_{\pm 0.02}}$                & ${{0.81}_{\pm 0.01}}$                        & ${{0.89}_{\pm 0.02}}$                       & ${{0.69}_{\pm 0.01}}$                    & ${{2.65}_{\pm 0.02}}$                   & ${{36.91}_{\pm 0.06}}$             & ${{3.86}_{\pm 0.02}}$              & ${{0.36}_{\pm 0.00}}$              & ${{8.47}_{\pm 0.01}}$              \\
            MTA~\citep{zanella2024test}                    & ${{11.35}_{\pm 0.05}}$                     & ${{5.04}_{\pm 0.02}}$                    & ${{6.27}_{\pm 0.03}}$                     & ${{9.04}_{\pm 0.06}}$                   & $\mathbf{{{26.00}_{\pm 0.06}}}$       & ${{4.33}_{\pm 0.01}}$                           & ${{11.18}_{\pm 0.03}}$                         & ${{21.69}_{\pm 0.02}}$                & ${{1.24}_{\pm 0.00}}$                        & ${{1.26}_{\pm 0.02}}$                       & ${{1.13}_{\pm 0.01}}$                    & ${{5.10}_{\pm 0.02}}$                   & ${{38.29}_{\pm 0.07}}$             & ${{4.93}_{\pm 0.03}}$              & ${{0.70}_{\pm 0.01}}$              & ${{9.84}_{\pm 0.01}}$              \\
            TDA~\citep{karmanov2024efficient}              & $\underline{{{29.07}_{\pm 0.02}}}$         & ${{14.75}_{\pm 0.05}}$                   & ${{19.09}_{\pm 0.02}}$                    & ${{23.57}_{\pm 0.02}}$                  & ${{18.48}_{\pm 0.03}}$                & ${{19.95}_{\pm 0.04}}$                          & ${{38.35}_{\pm 0.06}}$                         & $\underline{{{47.45}_{\pm 0.02}}}$    & ${{7.87}_{\pm 0.04}}$                        & ${{9.76}_{\pm 0.01}}$                       & ${{9.71}_{\pm 0.03}}$                    & $\underline{{{64.13}_{\pm 0.05}}}$      & ${{45.62}_{\pm 0.10}}$             & $\underline{{{35.45}_{\pm 0.05}}}$ & ${{34.81}_{\pm 0.11}}$             & ${{27.87}_{\pm 0.01}}$             \\
            \highlight{DMN~\citep{zhang2024dual}}          & ${{13.31}_{\pm 0.00}}$                     & ${{5.92}_{\pm 0.00}}$                    & ${{7.36}_{\pm 0.00}}$                     & ${{9.48}_{\pm 0.00}}$                   & ${{27.58}_{\pm 0.00}}$                & ${{3.78}_{\pm 0.00}}$                           & ${{10.75}_{\pm 0.00}}$                         & ${{21.66}_{\pm 0.00}}$                & ${{0.90}_{\pm 0.00}}$                        & ${{0.92}_{\pm 0.00}}$                       & ${{0.77}_{\pm 0.00}}$                    & ${{2.90}_{\pm 0.00}}$                   & ${{41.07}_{\pm 0.00}}$             & ${{4.39}_{\pm 0.00}}$              & ${{0.36}_{\pm 0.00}}$              & ${{10.08}_{\pm 0.00}}$             \\
            \highlight{Mint~\citep{bao2026mint}}           & ${{21.41}_{\pm 0.06}}$                     & $\mathbf{{{24.13}_{\pm 0.07}}}$          & ${{19.50}_{\pm 0.41}}$                    & $\mathbf{{{26.06}_{\pm 0.10}}}$         & ${{12.98}_{\pm 0.21}}$                & $\underline{{{28.46}_{\pm 0.10}}}$              & $\underline{{{39.42}_{\pm 0.03}}}$             & ${{46.80}_{\pm 0.13}}$                & $\underline{{{15.56}_{\pm 0.05}}}$           & $\mathbf{{{19.26}_{\pm 0.11}}}$             & $\underline{{{15.13}_{\pm 0.10}}}$       & ${{63.29}_{\pm 0.10}}$                  & $\underline{{{47.13}_{\pm 0.08}}}$ & ${{34.29}_{\pm 0.12}}$             & $\mathbf{{{39.43}_{\pm 0.05}}}$    & $\mathbf{{{30.19}_{\pm 0.02}}}$    \\
            \highlight{BATCLIP~\citep{Maharana_2025_ICCV}} & ${{19.44}_{\pm 1.60}}$                     & $\underline{{{18.85}_{\pm 0.32}}}$       & $\mathbf{{{22.28}_{\pm 0.16}}}$           & ${{11.78}_{\pm 3.17}}$                  & ${{4.67}_{\pm 0.17}}$                 & $\mathbf{{{33.33}_{\pm 0.17}}}$                 & ${{33.97}_{\pm 0.08}}$                         & ${{37.88}_{\pm 0.28}}$                & $\mathbf{{{16.80}_{\pm 0.06}}}$              & $\underline{{{17.33}_{\pm 0.16}}}$          & $\mathbf{{{17.98}_{\pm 0.15}}}$          & ${{59.99}_{\pm 0.09}}$                  & ${{42.60}_{\pm 0.32}}$             & ${{11.07}_{\pm 0.31}}$             & ${{35.82}_{\pm 0.07}}$             & ${{25.59}_{\pm 0.31}}$             \\
            UnInfo (ours)                                  & $\mathbf{{{30.82}_{\pm 0.05}}}$            & ${{15.28}_{\pm 0.37}}$                   & $\underline{{{20.41}_{\pm 0.38}}}$        & $\underline{{{24.48}_{\pm 0.46}}}$      & ${{20.80}_{\pm 0.43}}$                & ${{19.89}_{\pm 0.14}}$                          & $\mathbf{{{41.73}_{\pm 0.30}}}$                & $\mathbf{{{49.98}_{\pm 0.13}}}$       & ${{9.72}_{\pm 0.13}}$                        & ${{11.41}_{\pm 0.15}}$                      & ${{9.09}_{\pm 2.74}}$                    & $\mathbf{{{64.96}_{\pm 0.23}}}$         & $\mathbf{{{48.22}_{\pm 0.19}}}$    & $\mathbf{{{36.73}_{\pm 0.23}}}$    & $\underline{{{37.11}_{\pm 0.30}}}$ & $\underline{{{29.38}_{\pm 0.26}}}$ \\ \bottomrule
        \end{tabular}
    }
\end{table}

\begin{table}[tb]
    \centering
    \caption{\highlight{Test accuracy~(\%) of OpenAI ViT-B/32 on ImageNet-C-bar. The numbers (1), (5), and (10) presented with the method names are the shot numbers per class $n$ used for the few-shot adaptation methods.}}
    \label{tab:openai-vit-b-32_imagenet-c-bar_accuracy}
    \resizebox{1\linewidth}{!}{
        \begin{tabular}{llllllllllll}\toprule
            Method                                         & \rotatebox{60}{\shortstack{Blue\\noise\\sample}} & \rotatebox{60}{\shortstack{Brownish\\noise}} & \rotatebox{60}{\shortstack{Caustic\\refraction}} & \rotatebox{60}{\shortstack{Checkerboard\\cutout}} & \rotatebox{60}{\shortstack{Cocentric\\sine\\waves}} & \rotatebox{60}{\shortstack{Inverse\\sparkles}} & \rotatebox{60}{\shortstack{Perlin\\noise}} & \rotatebox{60}{\shortstack{Plasma\\noise}} & \rotatebox{60}{\shortstack{Single\\frequency\\greyscale}} & \rotatebox{60}{\shortstack{Sparkles}} & Mean                               \\ \midrule
            No-adapt                                       & ${29.44}$                                        & ${40.46}$                                    & ${32.98}$                                        & ${35.76}$                                         & ${10.59}$                                           & ${14.09}$                                      & ${44.83}$                                  & ${16.58}$                                  & ${26.53}$                                                 & ${44.70}$                             & ${29.60}$                          \\ \midrule
            Linear probing (1)                             & ${{11.85}_{\pm 0.40}}$                           & ${{17.61}_{\pm 0.17}}$                       & ${{12.22}_{\pm 0.16}}$                           & ${{14.22}_{\pm 0.02}}$                            & ${{3.17}_{\pm 0.15}}$                               & ${{4.76}_{\pm 0.04}}$                          & ${{17.89}_{\pm 0.39}}$                     & ${{6.70}_{\pm 0.28}}$                      & ${{6.94}_{\pm 0.17}}$                                     & ${{19.51}_{\pm 0.47}}$                & ${{11.49}_{\pm 0.04}}$             \\
            Linear probing (5)                             & ${{18.91}_{\pm 0.24}}$                           & ${{26.94}_{\pm 0.11}}$                       & ${{19.19}_{\pm 0.05}}$                           & ${{23.10}_{\pm 0.17}}$                            & ${{5.08}_{\pm 0.17}}$                               & ${{7.71}_{\pm 0.29}}$                          & ${{27.78}_{\pm 0.10}}$                     & ${{9.97}_{\pm 0.24}}$                      & ${{10.80}_{\pm 0.26}}$                                    & ${{30.39}_{\pm 0.17}}$                & ${{17.99}_{\pm 0.03}}$             \\
            Linear probing (10)                            & ${{22.96}_{\pm 0.33}}$                           & ${{30.90}_{\pm 0.17}}$                       & ${{23.29}_{\pm 0.29}}$                           & ${{27.44}_{\pm 0.03}}$                            & ${{6.63}_{\pm 0.05}}$                               & ${{9.32}_{\pm 0.09}}$                          & ${{33.01}_{\pm 0.41}}$                     & ${{11.52}_{\pm 0.18}}$                     & ${{13.51}_{\pm 0.31}}$                                    & ${{35.42}_{\pm 0.57}}$                & ${{21.40}_{\pm 0.04}}$             \\
            Tip-adapter~\citep{Zhang2022tip-adapter} (1)   & ${{17.16}_{\pm 0.37}}$                           & ${{24.37}_{\pm 0.08}}$                       & ${{18.03}_{\pm 0.17}}$                           & ${{20.65}_{\pm 0.30}}$                            & ${{4.94}_{\pm 0.17}}$                               & ${{6.83}_{\pm 0.10}}$                          & ${{26.05}_{\pm 0.17}}$                     & ${{9.00}_{\pm 0.16}}$                      & ${{11.84}_{\pm 0.11}}$                                    & ${{27.31}_{\pm 0.30}}$                & ${{16.62}_{\pm 0.03}}$             \\
            Tip-adapter (5)                                & ${{21.75}_{\pm 0.21}}$                           & ${{30.07}_{\pm 0.17}}$                       & ${{22.59}_{\pm 0.20}}$                           & ${{26.24}_{\pm 0.50}}$                            & ${{6.02}_{\pm 0.11}}$                               & ${{8.96}_{\pm 0.15}}$                          & ${{32.16}_{\pm 0.26}}$                     & ${{11.05}_{\pm 0.03}}$                     & ${{13.95}_{\pm 0.21}}$                                    & ${{33.69}_{\pm 0.31}}$                & ${{20.65}_{\pm 0.08}}$             \\
            Tip-adapter (10)                               & ${{24.19}_{\pm 0.35}}$                           & ${{32.85}_{\pm 0.09}}$                       & ${{25.15}_{\pm 0.36}}$                           & ${{29.52}_{\pm 0.19}}$                            & ${{7.10}_{\pm 0.02}}$                               & ${{10.42}_{\pm 0.04}}$                         & ${{35.13}_{\pm 0.11}}$                     & ${{12.59}_{\pm 0.14}}$                     & ${{16.01}_{\pm 0.31}}$                                    & ${{36.78}_{\pm 0.23}}$                & ${{22.97}_{\pm 0.06}}$             \\ \midrule
            TPT~\citep{shu2022tpt}                         & ${{30.43}_{\pm 0.09}}$                           & ${{42.77}_{\pm 0.04}}$                       & ${{35.87}_{\pm 0.03}}$                           & ${{38.12}_{\pm 0.04}}$                            & ${{11.32}_{\pm 0.04}}$                              & ${{16.35}_{\pm 0.05}}$                         & ${{46.75}_{\pm 0.04}}$                     & ${{18.03}_{\pm 0.04}}$                     & ${{28.87}_{\pm 0.09}}$                                    & ${{47.69}_{\pm 0.04}}$                & ${{31.62}_{\pm 0.02}}$             \\
            ZERO~\citep{farina2024frustratingly}           & ${{23.87}_{\pm 0.02}}$                           & ${{40.90}_{\pm 0.06}}$                       & $\mathbf{{{36.66}_{\pm 0.07}}}$                  & ${{39.51}_{\pm 0.04}}$                            & ${{9.68}_{\pm 0.08}}$                               & $\mathbf{{{19.63}_{\pm 0.01}}}$                & ${{44.81}_{\pm 0.08}}$                     & ${{17.72}_{\pm 0.04}}$                     & ${{28.03}_{\pm 0.09}}$                                    & ${{46.27}_{\pm 0.02}}$                & ${{30.71}_{\pm 0.02}}$             \\
            MTA~\citep{zanella2024test}                    & ${{28.94}_{\pm 0.05}}$                           & ${{40.85}_{\pm 0.01}}$                       & ${{35.61}_{\pm 0.01}}$                           & ${{37.84}_{\pm 0.05}}$                            & ${{11.17}_{\pm 0.03}}$                              & ${{16.68}_{\pm 0.01}}$                         & ${{45.65}_{\pm 0.03}}$                     & ${{16.61}_{\pm 0.02}}$                     & $\underline{{{29.65}_{\pm 0.06}}}$                        & ${{46.46}_{\pm 0.02}}$                & ${{30.95}_{\pm 0.01}}$             \\
            TDA~\citep{karmanov2024efficient}              & ${{33.49}_{\pm 0.12}}$                           & $\underline{{{43.47}_{\pm 0.07}}}$           & ${{35.45}_{\pm 0.04}}$                           & $\underline{{{39.97}_{\pm 0.06}}}$                & $\underline{{{12.42}_{\pm 0.05}}}$                  & ${{15.61}_{\pm 0.06}}$                         & $\underline{{{46.94}_{\pm 0.05}}}$         & $\underline{{{18.96}_{\pm 0.07}}}$         & ${{27.94}_{\pm 0.07}}$                                    & $\mathbf{{{48.50}_{\pm 0.06}}}$       & $\underline{{{32.27}_{\pm 0.01}}}$ \\
            \highlight{DMN~\citep{zhang2024dual}}          & ${{30.85}_{\pm 0.00}}$                           & ${{42.99}_{\pm 0.00}}$                       & ${{34.55}_{\pm 0.00}}$                           & ${{37.23}_{\pm 0.00}}$                            & ${{11.03}_{\pm 0.00}}$                              & ${{14.94}_{\pm 0.00}}$                         & ${{46.81}_{\pm 0.00}}$                     & ${{17.75}_{\pm 0.00}}$                     & ${{27.99}_{\pm 0.00}}$                                    & ${{47.35}_{\pm 0.00}}$                & ${{31.15}_{\pm 0.00}}$             \\
            \highlight{Mint~\citep{bao2026mint}}           & $\underline{{{34.85}_{\pm 0.25}}}$               & ${{42.73}_{\pm 0.01}}$                       & ${{29.31}_{\pm 0.07}}$                           & ${{39.76}_{\pm 0.09}}$                            & $\mathbf{{{14.05}_{\pm 0.07}}}$                     & ${{14.36}_{\pm 0.10}}$                         & ${{45.23}_{\pm 0.08}}$                     & $\mathbf{{{24.07}_{\pm 0.07}}}$            & ${{27.43}_{\pm 0.08}}$                                    & ${{46.48}_{\pm 0.11}}$                & ${{31.83}_{\pm 0.04}}$             \\
            \highlight{BATCLIP~\citep{Maharana_2025_ICCV}} & ${{31.13}_{\pm 0.05}}$                           & ${{36.48}_{\pm 0.06}}$                       & ${{25.79}_{\pm 0.18}}$                           & ${{31.04}_{\pm 0.07}}$                            & ${{9.43}_{\pm 0.51}}$                               & ${{11.51}_{\pm 0.19}}$                         & ${{39.30}_{\pm 0.30}}$                     & ${{3.70}_{\pm 0.43}}$                      & ${{17.83}_{\pm 0.25}}$                                    & ${{38.68}_{\pm 0.19}}$                & ${{24.49}_{\pm 0.19}}$             \\
            UnInfo (ours)                                  & $\mathbf{{{35.88}_{\pm 0.31}}}$                  & $\mathbf{{{43.86}_{\pm 0.17}}}$              & $\underline{{{36.34}_{\pm 0.31}}}$               & $\mathbf{{{40.03}_{\pm 0.21}}}$                   & ${{12.34}_{\pm 0.10}}$                              & $\underline{{{17.09}_{\pm 0.30}}}$             & $\mathbf{{{48.46}_{\pm 0.17}}}$            & ${{18.82}_{\pm 0.21}}$                     & $\mathbf{{{29.67}_{\pm 0.19}}}$                           & $\underline{{{48.03}_{\pm 0.30}}}$    & $\mathbf{{{33.05}_{\pm 0.09}}}$    \\ \bottomrule
        \end{tabular}
    }
\end{table}

\begin{table}[tb]
    \centering
    \caption{\highlight{Test accuracy~(\%) of SigLIP on ImageNet-C-bar. The numbers (1), (5), and (10) presented with the method names are the shot numbers per class $n$ used for the few-shot adaptation methods.}}
    \label{tab:siglip_imagenet-c-bar_accuracy}
    \resizebox{1\linewidth}{!}{
        \begin{tabular}{llllllllllll}\toprule
            Method                                         & \rotatebox{60}{\shortstack{Blue\\noise\\sample}} & \rotatebox{60}{\shortstack{Brownish\\noise}} & \rotatebox{60}{\shortstack{Caustic\\refraction}} & \rotatebox{60}{\shortstack{Checkerboard\\cutout}} & \rotatebox{60}{\shortstack{Cocentric\\sine\\waves}} & \rotatebox{60}{\shortstack{Inverse\\sparkles}} & \rotatebox{60}{\shortstack{Perlin\\noise}} & \rotatebox{60}{\shortstack{Plasma\\noise}} & \rotatebox{60}{\shortstack{Single\\frequency\\greyscale}} & \rotatebox{60}{\shortstack{Sparkles}} & Mean                               \\ \midrule
            %===================
            No-adapt                                       & ${32.20}$                                        & ${55.95}$                                    & ${46.04}$                                        & ${56.68}$                                         & ${13.33}$                                           & ${24.72}$                                      & ${59.48}$                                  & ${27.46}$                                  & ${21.30}$                                                 & ${60.34}$                             & ${39.75}$                          \\ \midrule
            Linear probing (1)                             & ${{14.39}_{\pm 0.18}}$                           & ${{28.10}_{\pm 0.35}}$                       & ${{20.19}_{\pm 0.20}}$                           & ${{26.71}_{\pm 2.33}}$                            & ${{4.46}_{\pm 0.10}}$                               & ${{9.99}_{\pm 0.37}}$                          & ${{29.67}_{\pm 0.14}}$                     & ${{10.57}_{\pm 0.26}}$                     & ${{6.95}_{\pm 0.34}}$                                     & ${{30.44}_{\pm 2.83}}$                & ${{18.15}_{\pm 0.50}}$             \\
            Linear probing (5)                             & ${{22.61}_{\pm 0.10}}$                           & ${{43.51}_{\pm 0.08}}$                       & ${{32.23}_{\pm 0.31}}$                           & ${{43.69}_{\pm 0.23}}$                            & ${{7.74}_{\pm 0.11}}$                               & ${{16.64}_{\pm 0.42}}$                         & ${{45.60}_{\pm 0.23}}$                     & ${{17.11}_{\pm 0.18}}$                     & ${{11.80}_{\pm 0.28}}$                                    & ${{47.98}_{\pm 0.16}}$                & ${{28.89}_{\pm 0.03}}$             \\
            Linear probing (10)                            & ${{26.64}_{\pm 0.40}}$                           & ${{49.88}_{\pm 0.33}}$                       & ${{38.19}_{\pm 0.18}}$                           & ${{49.51}_{\pm 0.16}}$                            & ${{9.89}_{\pm 0.10}}$                               & ${{19.82}_{\pm 0.20}}$                         & ${{51.88}_{\pm 0.30}}$                     & ${{20.63}_{\pm 0.40}}$                     & ${{15.22}_{\pm 0.25}}$                                    & ${{54.09}_{\pm 0.12}}$                & ${{33.57}_{\pm 0.01}}$             \\
            Tip-adapter~\citep{Zhang2022tip-adapter} (1)   & ${{19.91}_{\pm 0.25}}$                           & ${{39.06}_{\pm 0.15}}$                       & ${{29.38}_{\pm 0.08}}$                           & ${{39.06}_{\pm 0.14}}$                            & ${{6.93}_{\pm 0.21}}$                               & ${{14.89}_{\pm 0.30}}$                         & ${{41.90}_{\pm 0.07}}$                     & ${{15.63}_{\pm 0.27}}$                     & ${{11.58}_{\pm 0.20}}$                                    & ${{42.54}_{\pm 0.27}}$                & ${{26.09}_{\pm 0.03}}$             \\
            Tip-adapter (5)                                & ${{25.00}_{\pm 0.31}}$                           & ${{46.29}_{\pm 0.10}}$                       & ${{35.35}_{\pm 0.23}}$                           & ${{46.78}_{\pm 0.37}}$                            & ${{8.57}_{\pm 0.09}}$                               & ${{18.50}_{\pm 0.22}}$                         & ${{48.91}_{\pm 0.28}}$                     & ${{18.72}_{\pm 0.37}}$                     & ${{13.56}_{\pm 0.08}}$                                    & ${{50.05}_{\pm 0.11}}$                & ${{31.17}_{\pm 0.06}}$             \\
            Tip-adapter (10)                               & ${{27.78}_{\pm 0.14}}$                           & ${{49.58}_{\pm 0.26}}$                       & ${{38.62}_{\pm 0.47}}$                           & ${{50.29}_{\pm 0.21}}$                            & ${{10.08}_{\pm 0.04}}$                              & ${{20.43}_{\pm 0.34}}$                         & ${{52.48}_{\pm 0.18}}$                     & ${{20.86}_{\pm 0.23}}$                     & ${{15.78}_{\pm 0.13}}$                                    & ${{53.58}_{\pm 0.19}}$                & ${{33.95}_{\pm 0.04}}$             \\ \midrule
            TPT~\citep{shu2022tpt}                         & ${{4.86}_{\pm 0.00}}$                            & ${{17.57}_{\pm 0.00}}$                       & ${{16.77}_{\pm 0.01}}$                           & ${{22.19}_{\pm 0.02}}$                            & ${{2.40}_{\pm 0.01}}$                               & ${{0.66}_{\pm 0.00}}$                          & ${{37.78}_{\pm 0.01}}$                     & ${{8.47}_{\pm 0.01}}$                      & ${{2.23}_{\pm 0.00}}$                                     & ${{31.88}_{\pm 0.02}}$                & ${{14.48}_{\pm 0.00}}$             \\
            ZERO~\citep{farina2024frustratingly}           & ${{6.32}_{\pm 0.03}}$                            & ${{19.31}_{\pm 0.01}}$                       & ${{22.32}_{\pm 0.02}}$                           & ${{27.12}_{\pm 0.03}}$                            & ${{2.09}_{\pm 0.01}}$                               & ${{4.64}_{\pm 0.02}}$                          & ${{43.11}_{\pm 0.09}}$                     & ${{8.54}_{\pm 0.01}}$                      & ${{5.53}_{\pm 0.02}}$                                     & ${{35.37}_{\pm 0.06}}$                & ${{17.44}_{\pm 0.00}}$             \\
            MTA~\cite{zanella2024test}                     & ${{8.42}_{\pm 0.01}}$                            & ${{20.05}_{\pm 0.07}}$                       & ${{21.69}_{\pm 0.03}}$                           & ${{27.69}_{\pm 0.05}}$                            & ${{2.29}_{\pm 0.02}}$                               & ${{2.12}_{\pm 0.00}}$                          & ${{44.93}_{\pm 0.07}}$                     & ${{8.62}_{\pm 0.01}}$                      & ${{6.92}_{\pm 0.04}}$                                     & ${{36.89}_{\pm 0.04}}$                & ${{17.96}_{\pm 0.02}}$             \\
            TDA~\citep{karmanov2024efficient}              & ${{35.11}_{\pm 0.04}}$                           & ${{59.13}_{\pm 0.11}}$                       & $\underline{{{48.80}_{\pm 0.05}}}$               & ${{58.94}_{\pm 0.05}}$                            & $\underline{{{16.30}_{\pm 0.06}}}$                  & $\underline{{{28.05}_{\pm 0.02}}}$             & $\underline{{{61.59}_{\pm 0.06}}}$         & ${{30.90}_{\pm 0.05}}$                     & ${{23.55}_{\pm 0.08}}$                                    & ${{63.03}_{\pm 0.01}}$                & ${{42.54}_{\pm 0.02}}$             \\
            \highlight{DMN~\citep{zhang2024dual}}          & ${{5.85}_{\pm 0.00}}$                            & ${{20.02}_{\pm 0.00}}$                       & ${{19.82}_{\pm 0.00}}$                           & ${{27.43}_{\pm 0.00}}$                            & ${{2.29}_{\pm 0.00}}$                               & ${{0.64}_{\pm 0.00}}$                          & ${{46.16}_{\pm 0.00}}$                     & ${{9.36}_{\pm 0.00}}$                      & ${{2.39}_{\pm 0.00}}$                                     & ${{37.91}_{\pm 0.00}}$                & ${{17.19}_{\pm 0.00}}$             \\
            \highlight{Mint~\citep{bao2026mint}}           & $\mathbf{{{44.16}_{\pm 0.05}}}$                  & $\underline{{{59.33}_{\pm 0.02}}}$           & ${{45.66}_{\pm 0.13}}$                           & $\underline{{{59.43}_{\pm 0.12}}}$                & $\mathbf{{{20.53}_{\pm 0.19}}}$                     & ${{27.13}_{\pm 0.10}}$                         & ${{59.74}_{\pm 0.10}}$                     & $\mathbf{{{33.77}_{\pm 0.10}}}$            & $\mathbf{{{31.82}_{\pm 0.08}}}$                           & $\mathbf{{{63.92}_{\pm 0.04}}}$       & $\mathbf{{{44.55}_{\pm 0.04}}}$    \\
            \highlight{BATCLIP~\citep{Maharana_2025_ICCV}} & ${{36.87}_{\pm 0.27}}$                           & ${{56.55}_{\pm 0.06}}$                       & ${{45.37}_{\pm 0.03}}$                           & ${{55.64}_{\pm 0.13}}$                            & ${{2.73}_{\pm 0.11}}$                               & ${{3.18}_{\pm 0.08}}$                          & ${{57.18}_{\pm 0.16}}$                     & ${{8.48}_{\pm 0.07}}$                      & ${{6.21}_{\pm 0.49}}$                                     & ${{58.93}_{\pm 0.04}}$                & ${{33.12}_{\pm 0.07}}$             \\
            UnInfo (ours)                                  & $\underline{{{38.76}_{\pm 0.19}}}$               & $\mathbf{{{60.79}_{\pm 0.18}}}$              & $\mathbf{{{50.93}_{\pm 0.30}}}$                  & $\mathbf{{{60.92}_{\pm 0.04}}}$                   & ${{15.74}_{\pm 2.17}}$                              & $\mathbf{{{28.82}_{\pm 0.60}}}$                & $\mathbf{{{63.03}_{\pm 0.13}}}$            & $\underline{{{31.70}_{\pm 0.27}}}$         & $\underline{{{25.57}_{\pm 0.12}}}$                        & $\underline{{{63.57}_{\pm 0.53}}}$    & $\underline{{{43.98}_{\pm 0.15}}}$ \\ \bottomrule
        \end{tabular}
    }
\end{table}

Similar to \cref{sssec:exp_adaptataion_performance}, we conducted experiments with OpenAI ViT-B/32 CLIP\footnote{\url{https://github.com/openai/CLIP}} and SigLIP~\citep{zhai2023sigmoid}.
\cref{tab:openai-vit-b-32_imagenet-c_accuracy,tab:siglip_imagenet-c_accuracy,tab:openai-vit-b-32_imagenet-c-bar_accuracy,tab:siglip_imagenet-c-bar_accuracy} show the results.
The trend aligns with the cases of the ViT-B/16 CLIP in \cref{sssec:exp_adaptataion_performance}.
\highlight{Similar to the results shown in \cref{sssec:exp_adaptataion_performance}, UnInfo consistently improved the performance compared to No-adapt.
  In contrast, although the average accuracies of Mint and BATCLIP are sometimes higher than UnInfo, they are also sometimes lower than No-adapt, suggesting that UnInfo is more stable across corruption types and model architectures.}

\paragraph{\highlight{CIFAR-C datasets.}}
\highlight{We conducted experiments on CIFAR10/100-C~\citep{imagenet-c}, which are the corrupted versions of CIFAR10/100~\citep{krizhevsky2009learning}.
  \cref{tab:vit-b-16_cifar10c_accuracy,tab:vit-b-32_cifar10c_accuracy,tab:siglip_cifar10c_accuracy} and \cref{tab:vit-b-16_cifar100c_accuracy,tab:vit-b-32_cifar100c_accuracy,tab:siglip_cifar100c_accuracy} show the results on CIFAR10-C and CIFAR100-C, respectively.
  Although Mint and BATCLIP achieved higher accuracy in most cases, UnInfo also improved accuracy compared to the other baselines, suggesting the efficacy against image corruption.}
\begin{table}[tb]
    \centering
    \caption{\highlight{Test accuracy (\%) of OpenCLIP ViT-B/16 on CIFAR10-C. The numbers (1), (5), and (10) presented with the method names are the shot numbers per class $n$ used for the few-shot adaptation methods.}}
    \label{tab:vit-b-16_cifar10c_accuracy}
    \resizebox{1\linewidth}{!}{
        \setlength{\tabcolsep}{3pt}
\begin{tabular}{lllllllllllllllll} \toprule
            %================
            Method                                       & \rotatebox{60}{\shortstack{Defocus\\blur}} & \rotatebox{60}{\shortstack{Glass\\blur}} & \rotatebox{60}{\shortstack{Motion\\blur}} & \rotatebox{60}{\shortstack{Zoom\\blur}} & \rotatebox{60}{\shortstack{Contrast}} & \rotatebox{60}{\shortstack{Elastic\\transform}} & \rotatebox{60}{\shortstack{Jpeg\\compression}} & \rotatebox{60}{\shortstack{Pixelate}} & \rotatebox{60}{\shortstack{Gaussian\\noise}} & \rotatebox{60}{\shortstack{Impulse\\noise}} & \rotatebox{60}{\shortstack{Shot\\noise}} & \rotatebox{60}{\shortstack{Brightness}} & \rotatebox{60}{\shortstack{Fog}} & \rotatebox{60}{\shortstack{Frost}} & \rotatebox{60}{\shortstack{Snow}} & Mean                \\ \midrule
            No-adapt                                     & ${80.68}$                                  & ${43.23}$                                & ${73.91}$                                 & ${82.09}$                               & ${69.40}$                             & ${64.51}$                                       & ${62.81}$                                      & ${51.05}$                             & ${47.96}$                                    & ${60.98}$                                   & ${53.50}$                                & ${90.82}$                               & ${81.27}$                        & ${83.88}$                          & ${83.09}$                         & ${68.61}$           \\
\bottomrule
        \end{tabular}
    }
\end{table}

\begin{table}[tb]
    \centering
    \caption{\highlight{Test accuracy (\%) of OpenAI ViT-B/32 CLIP on CIFAR10-C. The numbers (1), (5), and (10) presented with the method names are the shot numbers per class $n$ used for the few-shot adaptation methods.}}
    \label{tab:vit-b-32_cifar10c_accuracy}
    \resizebox{1\linewidth}{!}{
        \setlength{\tabcolsep}{3pt}
        \begin{tabular}{lllllllllllllllll} \toprule
            %================
            Method                                       & \rotatebox{60}{\shortstack{Defocus\\blur}} & \rotatebox{60}{\shortstack{Glass\\blur}} & \rotatebox{60}{\shortstack{Motion\\blur}} & \rotatebox{60}{\shortstack{Zoom\\blur}} & \rotatebox{60}{\shortstack{Contrast}} & \rotatebox{60}{\shortstack{Elastic\\transform}} & \rotatebox{60}{\shortstack{Jpeg\\compression}} & \rotatebox{60}{\shortstack{Pixelate}} & \rotatebox{60}{\shortstack{Gaussian\\noise}} & \rotatebox{60}{\shortstack{Impulse\\noise}} & \rotatebox{60}{\shortstack{Shot\\noise}} & \rotatebox{60}{\shortstack{Brightness}} & \rotatebox{60}{\shortstack{Fog}} & \rotatebox{60}{\shortstack{Frost}} & \rotatebox{60}{\shortstack{Snow}} & Mean                \\ \midrule
            No-adapt                                     & ${69.72}$                                  & ${42.40}$                                & ${63.95}$                                 & ${69.84}$                               & ${64.47}$                             & ${60.87}$                                       & ${55.40}$                                      & ${50.53}$                             & ${35.49}$                                    & ${42.80}$                                   & ${39.77}$                                & ${81.92}$                               & ${67.03}$                        & ${72.92}$                          & ${71.78}$                         & ${59.26}$           \\
\bottomrule
        \end{tabular}
    }
\end{table}

\begin{table}[tb]
    \centering
    \caption{\highlight{Test accuracy (\%) of SigLIP on CIFAR10-C. The numbers (1), (5), and (10) presented with the method names are the shot numbers per class $n$ used for the few-shot adaptation methods.}}
    \label{tab:siglip_cifar10c_accuracy}
    \resizebox{1\linewidth}{!}{
        \setlength{\tabcolsep}{3pt}
        \begin{tabular}{lllllllllllllllll} \toprule
            %================
            Method                                       & \rotatebox{60}{\shortstack{Defocus\\blur}} & \rotatebox{60}{\shortstack{Glass\\blur}} & \rotatebox{60}{\shortstack{Motion\\blur}} & \rotatebox{60}{\shortstack{Zoom\\blur}} & \rotatebox{60}{\shortstack{Contrast}} & \rotatebox{60}{\shortstack{Elastic\\transform}} & \rotatebox{60}{\shortstack{Jpeg\\compression}} & \rotatebox{60}{\shortstack{Pixelate}} & \rotatebox{60}{\shortstack{Gaussian\\noise}} & \rotatebox{60}{\shortstack{Impulse\\noise}} & \rotatebox{60}{\shortstack{Shot\\noise}} & \rotatebox{60}{\shortstack{Brightness}} & \rotatebox{60}{\shortstack{Fog}} & \rotatebox{60}{\shortstack{Frost}} & \rotatebox{60}{\shortstack{Snow}} & Mean                \\ \midrule
            No-adapt                                     & ${65.65}$                                  & ${39.06}$                                & ${65.55}$                                 & ${66.51}$                               & ${55.36}$                             & ${50.81}$                                       & ${43.96}$                                      & ${57.18}$                             & ${28.35}$                                    & ${41.48}$                                   & ${32.61}$                                & ${86.60}$                               & ${73.28}$                        & ${74.88}$                          & ${76.80}$                         & ${57.21}$           \\
\bottomrule
        \end{tabular}
    }
\end{table}

\begin{table}[tb]
    \centering
    \caption{\highlight{Test accuracy (\%) of OpenCLIP ViT-B/16 on CIFAR100-C. The numbers (1), (5), and (10) presented with the method names are the shot numbers per class $n$ used for the few-shot adaptation methods.}}
    \label{tab:vit-b-16_cifar100c_accuracy}
    \resizebox{1\linewidth}{!}{
        \setlength{\tabcolsep}{3pt}
        \begin{tabular}{lllllllllllllllll} \toprule
            %================
            Method                                       & \rotatebox{60}{\shortstack{Defocus\\blur}} & \rotatebox{60}{\shortstack{Glass\\blur}} & \rotatebox{60}{\shortstack{Motion\\blur}} & \rotatebox{60}{\shortstack{Zoom\\blur}} & \rotatebox{60}{\shortstack{Contrast}} & \rotatebox{60}{\shortstack{Elastic\\transform}} & \rotatebox{60}{\shortstack{Jpeg\\compression}} & \rotatebox{60}{\shortstack{Pixelate}} & \rotatebox{60}{\shortstack{Gaussian\\noise}} & \rotatebox{60}{\shortstack{Impulse\\noise}} & \rotatebox{60}{\shortstack{Shot\\noise}} & \rotatebox{60}{\shortstack{Brightness}} & \rotatebox{60}{\shortstack{Fog}} & \rotatebox{60}{\shortstack{Frost}} & \rotatebox{60}{\shortstack{Snow}} & Mean                \\ \midrule
            No-adapt                                     & ${53.49}$                                  & ${16.15}$                                & ${48.43}$                                 & ${56.98}$                               & ${39.75}$                             & ${32.74}$                                       & ${34.52}$                                      & ${25.62}$                             & ${22.08}$                                    & ${26.82}$                                   & ${24.83}$                                & ${65.20}$                               & ${51.83}$                        & ${55.53}$                          & ${56.89}$                         & ${40.72}$           \\
\bottomrule
        \end{tabular}
    }
\end{table}

\begin{table}[tb]
    \centering
    \caption{\highlight{Test accuracy (\%) of OpenAI ViT-B/32 CLIP on CIFAR100-C. The numbers (1), (5), and (10) presented with the method names are the shot numbers per class $n$ used for the few-shot adaptation methods.}}
    \label{tab:vit-b-32_cifar100c_accuracy}
    \resizebox{1\linewidth}{!}{
        \setlength{\tabcolsep}{3pt}
        \begin{tabular}{lllllllllllllllll} \toprule
            %================
            Method                                       & \rotatebox{60}{\shortstack{Defocus\\blur}} & \rotatebox{60}{\shortstack{Glass\\blur}} & \rotatebox{60}{\shortstack{Motion\\blur}} & \rotatebox{60}{\shortstack{Zoom\\blur}} & \rotatebox{60}{\shortstack{Contrast}} & \rotatebox{60}{\shortstack{Elastic\\transform}} & \rotatebox{60}{\shortstack{Jpeg\\compression}} & \rotatebox{60}{\shortstack{Pixelate}} & \rotatebox{60}{\shortstack{Gaussian\\noise}} & \rotatebox{60}{\shortstack{Impulse\\noise}} & \rotatebox{60}{\shortstack{Shot\\noise}} & \rotatebox{60}{\shortstack{Brightness}} & \rotatebox{60}{\shortstack{Fog}} & \rotatebox{60}{\shortstack{Frost}} & \rotatebox{60}{\shortstack{Snow}} & Mean                \\ \midrule
            No-adapt                                     & ${36.76}$                                  & ${14.21}$                                & ${36.10}$                                 & ${40.31}$                               & ${29.54}$                             & ${26.36}$                                       & ${25.96}$                                      & ${22.02}$                             & ${14.78}$                                    & ${13.83}$                                   & ${15.99}$                                & ${48.18}$                               & ${38.00}$                        & ${40.57}$                          & ${38.92}$                         & ${29.44}$           \\
\bottomrule
        \end{tabular}
    }
\end{table}

\begin{table}[tb]
    \centering
    \caption{\highlight{Test accuracy (\%) of SigLIP on CIFAR100-C. The numbers (1), (5), and (10) presented with the method names are the shot numbers per class $n$ used for the few-shot adaptation methods.}}
    \label{tab:siglip_cifar100c_accuracy}
    \resizebox{1\linewidth}{!}{
        \setlength{\tabcolsep}{3pt}
        \begin{tabular}{lllllllllllllllll} \toprule
            %================
            Method                                       & \rotatebox{60}{\shortstack{Defocus\\blur}} & \rotatebox{60}{\shortstack{Glass\\blur}} & \rotatebox{60}{\shortstack{Motion\\blur}} & \rotatebox{60}{\shortstack{Zoom\\blur}} & \rotatebox{60}{\shortstack{Contrast}} & \rotatebox{60}{\shortstack{Elastic\\transform}} & \rotatebox{60}{\shortstack{Jpeg\\compression}} & \rotatebox{60}{\shortstack{Pixelate}} & \rotatebox{60}{\shortstack{Gaussian\\noise}} & \rotatebox{60}{\shortstack{Impulse\\noise}} & \rotatebox{60}{\shortstack{Shot\\noise}} & \rotatebox{60}{\shortstack{Brightness}} & \rotatebox{60}{\shortstack{Fog}} & \rotatebox{60}{\shortstack{Frost}} & \rotatebox{60}{\shortstack{Snow}} & Mean                \\ \midrule
            No-adapt                                     & ${40.19}$                                  & ${15.63}$                                & ${40.63}$                                 & ${44.75}$                               & ${29.20}$                             & ${24.75}$                                       & ${21.60}$                                      & ${24.48}$                             & ${14.00}$                                    & ${25.16}$                                   & ${15.59}$                                & ${60.89}$                               & ${47.84}$                        & ${47.56}$                          & ${49.88}$                         & ${33.48}$           \\
\bottomrule
        \end{tabular}
    }
\end{table}

\paragraph{Domain Shifts.}
\begin{table}[tb]
    \centering
    \caption{Test accuracy (\%) on domain shifts.
        The numbers (1), (5), and (10) presented with the method names are the shot numbers per class $n$ used for the few-shot adaptation methods.
        `-' represents that the setting is infeasible due to the dataset size.}
    \label{tab:domain-shift_accuracy}
    \resizebox{0.7\linewidth}{!}{
        \begin{tabular}{lllll}\toprule
            Method                                       & ImageNet  & ImageNet-A & ImageNet-R & Mean      \\ \midrule
            No-adapt                                     & ${66.97}$ & ${32.91}$  & ${74.36}$  & ${58.08}$ \\ \midrule
            Linear probing (1)                           & ${37.86}$ & ${14.07}$  & ${23.48}$  & ${25.13}$ \\
            Linear probing (5)                           & ${55.95}$ & -          & ${52.91}$  & -         \\
            Linear probing (10)                          & ${62.50}$ & -          & ${62.51}$  & -         \\
            Tip-adapter~\citep{Zhang2022tip-adapter} (1) & ${57.55}$ & ${30.80}$  & ${63.70}$  & ${50.68}$ \\
            Tip-adapter (5)                              & ${61.76}$ & -          & ${63.82}$  & -         \\
            Tip-adapter (10)                             & ${64.00}$ & -          & ${66.08}$  & -         \\ \midrule
            TPT~\citep{shu2022tpt}                       & ${70.70}$ & ${19.92}$  & ${80.26}$  & ${56.96}$ \\
            ZERO~\citep{farina2024frustratingly}         & ${69.02}$ & ${43.23}$  & ${75.53}$  & ${62.59}$ \\
            MTA~\citep{zanella2024test}                  & ${68.25}$ & ${35.93}$  & ${74.72}$  & ${59.63}$ \\
            TDA~\citep{karmanov2024efficient}            & ${70.33}$ & ${33.24}$  & ${76.81}$  & ${60.13}$ \\
            UnInfo (ours)                                & ${67.82}$ & ${33.13}$  & ${76.53}$  & ${59.16}$ \\ \bottomrule
        \end{tabular}
    }
\end{table}

We also tested TTA baselines and our UnInfo on domain shift datasets: ImageNet~\citep{imagenet}, ImageNet-A~\citep{Hendrycks_2021_CVPR}, and R~\citep{imagenet-r}.
\cref{tab:domain-shift_accuracy} shows the results.
Although UnInfo did not outperform the existing TTA baselines since they are specifically designed for domain shifts, UnInfo had slight improvements and no negative effects compared to No-adapt.
This implies that the applicability of our method is not limited within image corruption by adaptively controlling the balance between entropy and uniformity.
Moreover, one may incorporate UnInfo with other methods.

\subsection{\highlight{Ablation Study}}\label{assec:additional_ablation}
\paragraph{\highlight{Efficacy of updating the image encoder.}}
\highlight{To verify the efficacy of updating the image encoder rather than image embeddings, we compared a variant of UnInfo that freezes the image encoder and updates image embeddings as optimized parameters.
  The loss is the same as \cref{eq:proposed_balaning_loss}, but the knowledge distillation loss $\mathcal{L}_\text{pl}$ is omitted because the EMA teacher is not available in this setting.
  \cref{tab:accuracy_uninfo_variants} shows the result.
  UnInfo (embedding) had only a slight effect on accuracy compared to No-adapt, suggesting that updating the image encoder is efficient.
  This is because UnInfo can accumulate knowledge of the target distribution by updating the image encoder, whereas updating image embeddings is episodic and cannot leverage knowledge from past inputs.}
\camhighlight{Note that the effect of updating the image encoder cannot be fully separated from that of directly updating image embeddings.
  When directly updating image embeddings instead of the encoder, an equivalent EMA teacher $\mathcal{L}_\text{pl}$ is not well-defined because the EMA teacher is obtained by accumulating the history of the updated encoder.}

\begin{table}[tb]
    \centering
    \caption{\highlight{Test accuracy (\%) of UnInfo variants on ImageNet-C with ViT-B/16 CLIP.}}
    \label{tab:accuracy_uninfo_variants}
    \resizebox{1.0\linewidth}{!}{
        \begin{tabular}{lllllllllllllllll} \toprule
            %================
            Method             & \rotatebox{60}{\shortstack{Defocus\\blur}} & \rotatebox{60}{\shortstack{Glass\\blur}} & \rotatebox{60}{\shortstack{Motion\\blur}} & \rotatebox{60}{\shortstack{Zoom\\blur}} & \rotatebox{60}{\shortstack{Contrast}} & \rotatebox{60}{\shortstack{Elastic\\transform}} & \rotatebox{60}{\shortstack{Jpeg\\compression}} & \rotatebox{60}{\shortstack{Pixelate}} & \rotatebox{60}{\shortstack{Gaussian\\noise}} & \rotatebox{60}{\shortstack{Impulse\\noise}} & \rotatebox{60}{\shortstack{Shot\\noise}} & \rotatebox{60}{\shortstack{Brightness}} & \rotatebox{60}{\shortstack{Fog}} & \rotatebox{60}{\shortstack{Frost}} & \rotatebox{60}{\shortstack{Snow}} & Mean      \\ \midrule
            No-adapt           & ${28.31}$                                  & ${11.89}$                                & ${19.16}$                                 & ${17.61}$                               & ${17.87}$                             & ${13.22}$                                       & ${36.79}$                                      & ${37.01}$                             & ${6.08}$                                     & ${6.17}$                                    & ${7.85}$                                 & ${54.89}$                               & ${34.55}$                        & ${27.35}$                          & ${27.67}$                         & ${23.09}$ \\
            UnInfo (ours)      & ${31.51}$                                  & ${16.76}$                                & ${23.47}$                                 & ${20.40}$                               & ${22.81}$                             & ${16.59}$                                       & ${42.03}$                                      & ${42.38}$                             & ${7.56}$                                     & ${10.60}$                                   & ${11.36}$                                & ${57.75}$                               & ${39.16}$                        & ${31.65}$                          & ${32.40}$                         & ${27.10}$ \\
            UnInfo (embedding) & ${28.19}$                                  & ${11.81}$                                & ${19.13}$                                 & ${17.65}$                               & ${17.89}$                             & ${13.37}$                                       & ${36.76}$                                      & ${36.92}$                             & ${6.02}$                                     & ${6.13}$                                    & ${7.87}$                                 & ${54.74}$                               & ${34.39}$                        & ${27.31}$                          & ${27.78}$                         & ${23.06}$ \\ \bottomrule
        \end{tabular}
    }
\end{table}

\begin{figure}[tb]
  \centering
  \includegraphics[width=0.3\linewidth]{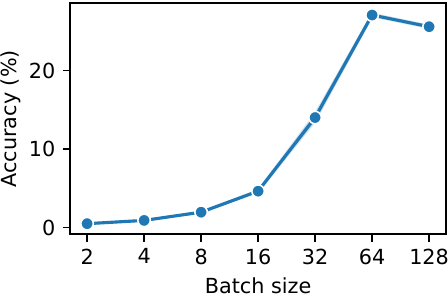}
  \caption{\highlight{Sensitivity analysis of the batch size.
      The mean and standard deviation of the test accuracy calculated over the ImageNet-C corruptions are plotted.}}
  \label{fig:sensitivity_batch-size}
\end{figure}

\paragraph{\highlight{Ablation on the batch size.}}
\highlight{\cref{fig:sensitivity_batch-size} shows the mean accuracy of UnInfo over ImageNet-C corruptions with ViT-B/16 CLIP.
  When the batch size $\leq 32$, the accuracy dropped significantly.
  This is because the uniformity estimation becomes poor with small batch sizes, which is one limitation of UnInfo, as mentioned in \cref{sec:conclusion}.
  One possible modification is to accumulate batch statistics, such as a moving average of uniformity, to compensate for the estimation when using a single small batch size.}

\subsection{\highlight{Hyperparameter Selection}}\label{assec:hyperparameter_selection}
\begin{figure}[tb]
  \centering
  \includegraphics[width=0.3\linewidth]{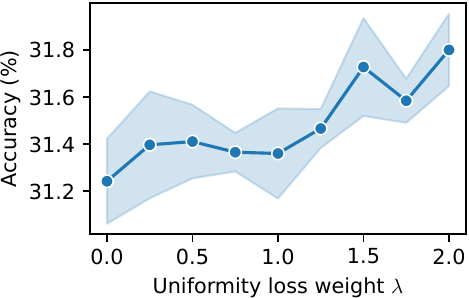}
  \caption{\highlight{Hyperparameter sensitivity on only defocus blur.}}
  \label{fig:lambda_sensitivity_defocus}
\end{figure}

\highlight{In \cref{ssec:exp_implementation}, we selected the weight of uniformity loss $\lambda$ based on brightness, defocus blur, elastic transform, and Gaussian noise.
  Here, we searched $\lambda$ based on only defocus blur from $\{ 0.0, 0.25, \ldots,  2.0 \}$ and checked the selected $\lambda$'s performance to examine whether the hyperparameter setting does not overfit.
  \cref{fig:lambda_sensitivity_defocus} shows the result.
  The best values based on defocus blur were $1.5$ and $2.0$.
  From the sensitivity analysis in \cref{fig:sensitivity_analysis}, these values do not have a large effect on the average accuracy across the other corruption types, suggesting the reasonableness of the $\lambda$ used in the experiments.}

\highlight{Moreover, the hyperparameter setting we used in the experiments also works on ImageNet-C-bar.
  The corruption types in ImageNet-C-bar are selected to be dissimilar to ImageNet-C~\citep{mintun2021on}, suggesting that our hyperparameter setting does not overfit to the selected corruption types.}

\subsection{\highlight{Continual Adaptation}}\label{assec:continual_adaptation}
\highlight{We tested UnInfo in the continual TTA setting, where adaptation continues over all corruption types sequentially without resetting the model.
  Here, we used the same hyperparameter setting as in the main experiments, except setting the learning rate to $10^{-4}$ to avoid catastrophic forgetting.
  \cref{tab:imagenet-c_continual} shows the result on ImageNet-C with ViT-B/16 CLIP.
  Although UnInfo improved accuracy compared to No-adapt on most corruption types, the accuracy gains for the latter half of corruption types are smaller than those for the former half.
  This is because the EMA teacher retains the LoRA parameters adapted to previous corruptions, which affect current adaptation.
  One possible modification would be to adaptively reset LoRA using change detection, which is one direction for future work.}

\begin{table}[tb]
    \centering
    \caption{\highlight{Test accuracy (\%) of UnInfo on ImageNet-C under the continual adaptation setting.}}
    \label{tab:imagenet-c_continual}
    \resizebox{1\linewidth}{!}{
        \setlength{\tabcolsep}{3pt}
        \begin{tabular}{lllllllllllllllll} \toprule
            %================
            Method   & \rotatebox{60}{\shortstack{Defocus\\blur}} & \rotatebox{60}{\shortstack{Glass\\blur}} & \rotatebox{60}{\shortstack{Motion\\blur}} & \rotatebox{60}{\shortstack{Zoom\\blur}} & \rotatebox{60}{\shortstack{Contrast}} & \rotatebox{60}{\shortstack{Elastic\\transform}} & \rotatebox{60}{\shortstack{Jpeg\\compression}} & \rotatebox{60}{\shortstack{Pixelate}} & \rotatebox{60}{\shortstack{Gaussian\\noise}} & \rotatebox{60}{\shortstack{Impulse\\noise}} & \rotatebox{60}{\shortstack{Shot\\noise}} & \rotatebox{60}{\shortstack{Brightness}} & \rotatebox{60}{\shortstack{Fog}} & \rotatebox{60}{\shortstack{Frost}} & \rotatebox{60}{\shortstack{Snow}} & Mean                   \\ \midrule
            No-adapt & ${28.31}$                                  & ${11.89}$                                & ${19.16}$                                 & ${17.61}$                               & ${17.87}$                             & ${13.22}$                                       & ${36.79}$                                      & ${37.01}$                             & ${6.08}$                                     & ${6.17}$                                    & ${7.85}$                                 & ${54.89}$                               & ${34.55}$                        & ${27.35}$                          & ${27.67}$                         & ${23.09}$              \\
            UnInfo   & ${{29.21}_{\pm 0.05}}$                     & ${{13.62}_{\pm 0.06}}$                   & ${{22.14}_{\pm 0.36}}$                    & ${{20.70}_{\pm 0.31}}$                  & ${{20.21}_{\pm 0.32}}$                & ${{17.24}_{\pm 0.70}}$                          & ${{36.33}_{\pm 0.33}}$                         & ${{36.64}_{\pm 0.50}}$                & ${{6.06}_{\pm 0.07}}$                        & ${{6.41}_{\pm 0.10}}$                       & ${{8.23}_{\pm 0.11}}$                    & ${{52.08}_{\pm 0.22}}$                  & ${{34.87}_{\pm 0.37}}$           & ${{28.71}_{\pm 0.03}}$             & ${{27.88}_{\pm 0.34}}$            & ${{24.02}_{\pm 0.09}}$ \\ \bottomrule
        \end{tabular}
    }
\end{table}

\subsection{\highlight{Analysis of Model Collapse with Entropy Loss}}\label{assec:entropy_collapse_analysis}
\highlight{In \cref{tab:ablation_imagenet-c,tab:ablation_imagenet-c-bar}, we observed that the entropy-only loss ($\mathcal{L}_\text{ent}$) resulted in model collapse.
  Here, we investigated the statistics of the model prediction to understand this phenomenon.
  \cref{tab:pred_class_ratio} shows the ratio of the class to which an unadapted ViT-B/16 CLIP classified the largest number of ImageNet-C samples.
  We can observe a significant imbalance due to image corruption.
  A large number of samples ($>1/1000=0.1\%$), especially around $20\%$ on noise corruptions, were classified to a single class, where the entropy-only loss exacerbates the imbalance.
  This is due to the degradation of uniformity observed in \cref{sec:preliminary_experiment}; By image corruption, a large number of image embeddings are distributed within the region of a single class in the embedding space.}

\highlight{The instability of simple entropy minimization is also addressed in classification TTA studies, such as SAR~\citep{niu2023towards} and ReCAP~\citep{hu2025beyond}.
  Combining these techniques is one direction of our future work.}

\begin{table}[tb]
  \centering
  \caption{\highlight{The ratio of the class to which an unadapted ViT-B/16 CLIP classified the largest number of ImageNet-C samples.}}
  \label{tab:pred_class_ratio}
  \resizebox{0.9\linewidth}{!}{
    \begin{tabular}{lllllllllllllll}\toprule
      \rotatebox{60}{\shortstack{Defocus\\blur}} & \rotatebox{60}{\shortstack{Glass\\blur}} & \rotatebox{60}{\shortstack{Motion\\blur}} & \rotatebox{60}{\shortstack{Zoom\\blur}} & \rotatebox{60}{\shortstack{Contrast}} & \rotatebox{60}{\shortstack{Elastic\\transform}} & \rotatebox{60}{\shortstack{Jpeg\\compression}} & \rotatebox{60}{\shortstack{Pixelate}} & \rotatebox{60}{\shortstack{Gaussian\\noise}} & \rotatebox{60}{\shortstack{Impulse\\noise}} & \rotatebox{60}{\shortstack{Shot\\noise}} & \rotatebox{60}{\shortstack{Brightness}} & \rotatebox{60}{\shortstack{Fog}} & \rotatebox{60}{\shortstack{Frost}} & \rotatebox{60}{\shortstack{Snow}} \\ \midrule
      1.16                                       & 6.42                                     & 5.77                                      & 5.09                                    & 3.88                                  & 5.85                                            & 0.84                                           & 0.93                                  & 24.45                                        & 23.96                                       & 18.89                                    & 0.73                                    & 2.86                             & 5.59                               & 7.01                              \\ \bottomrule
    \end{tabular}
  }
\end{table}

\subsection{\highlight{Optimized Parameters}}\label{asec:optimized_parameters}
\highlight{To validate the effect of using LoRA in UnInfo, we changed the optimized parameter set.
  Here, we compared two variants of UnInfo:}

\paragraph{\highlight{LayerNorm:}}
\highlight{updates the affine parameters in LayerNorm layers~\citep{ba2016layer} in the image encoder instead of LoRA, as existing studies (e.g., BATCLIP~\citep{Maharana_2025_ICCV} and Mint~\citep{bao2026mint})}

\paragraph{\highlight{LayerNorm (w/o EMA teacher):}}
\highlight{also updates the affine parameters in LayerNorm layers, but the EMA teacher regularization term $\mathcal{L}_\text{pl}$ is omitted.
  We compared this variant to verify whether model collapse can be avoided by limiting the number of optimized parameters, as we introduced $\mathcal{L}_\text{pl}$ to avoid model collapse.}

\highlight{\cref{tab:uninfo_opt_params} shows the result.
  Although LayerNorm (w/o EMA teacher) outperformed LoRA (original UnInfo) on several corruption types, LoRA achieved the best average accuracy.
  While LayerNorm sometimes had significantly higher accuracy than LayerNorm (w/o EMA teacher), e.g., defocus blur, it also sometimes had lower accuracy.
  When optimizing LayerNorm parameters, model collapse can be avoided without $\mathcal{L}_\text{pl}$ because LayerNorm has fewer degrees of freedom than LoRA.
  However, UnInfo (LoRA with $\mathcal{L}_\text{pl}$) performed the best because it hits a better balance between adaptation flexibility and regularization.}

\camhighlight{Further, to validate that the accuracy improvement comes from UnInfo's loss design rather than just using LoRA, we applied LoRA to BATCLIP~\citep{Maharana_2025_ICCV}, a LayerNorm-based TTA method.
  In \cref{tab:uninfo_opt_params}, BATCLIP (LoRA) significantly degraded accuracy compared to the original one in \cref{tab:imagenet-c_accuracy}.
  This result suggests that simply replacing LayerNorm with LoRA does not yield an improvement.
  Owing to UnInfo's loss design, UnInfo can benefit from LoRA's flexibility while keeping stability.}

\begin{table}[tb]
  \centering
  \caption{\highlight{LoRA vs. LayerNorm affine parameters in UnInfo on ImageNet-C.}}
  \label{tab:uninfo_opt_params}
  \resizebox{1.0\linewidth}{!}{
    \begin{tabular}{lllllllllllllllll} \toprule
      %================
      Method                        & \rotatebox{60}{\shortstack{Defocus\\blur}} & \rotatebox{60}{\shortstack{Glass\\blur}} & \rotatebox{60}{\shortstack{Motion\\blur}} & \rotatebox{60}{\shortstack{Zoom\\blur}} & \rotatebox{60}{\shortstack{Contrast}} & \rotatebox{60}{\shortstack{Elastic\\transform}} & \rotatebox{60}{\shortstack{Jpeg\\compression}} & \rotatebox{60}{\shortstack{Pixelate}} & \rotatebox{60}{\shortstack{Gaussian\\noise}} & \rotatebox{60}{\shortstack{Impulse\\noise}} & \rotatebox{60}{\shortstack{Shot\\noise}} & \rotatebox{60}{\shortstack{Brightness}} & \rotatebox{60}{\shortstack{Fog}} & \rotatebox{60}{\shortstack{Frost}} & \rotatebox{60}{\shortstack{Snow}} & Mean                            \\ \midrule
      No-adapt                      & ${28.31}$                                  & ${11.89}$                                & ${19.16}$                                 & ${17.61}$                               & ${17.87}$                             & ${13.22}$                                       & ${36.79}$                                      & ${37.01}$                             & ${6.08}$                                     & ${6.17}$                                    & ${7.85}$                                 & ${54.89}$                               & ${34.55}$                        & ${27.35}$                          & ${27.67}$                         & ${23.09}$                       \\
      \camhighlight{BATCLIP (LoRA)} & ${{2.00}_{\pm 0.31}}$                      & ${{1.49}_{\pm 0.11}}$                    & ${{1.77}_{\pm 0.29}}$                     & ${{2.10}_{\pm 0.38}}$                   & ${{1.22}_{\pm 0.13}}$                 & ${{2.09}_{\pm 0.18}}$                           & ${{3.19}_{\pm 0.48}}$                          & ${{3.63}_{\pm 0.38}}$                 & ${{0.20}_{\pm 0.00}}$                        & ${{0.48}_{\pm 0.28}}$                       & ${{0.61}_{\pm 0.43}}$                    & ${{5.55}_{\pm 1.08}}$                   & ${{3.50}_{\pm 0.36}}$            & ${{2.70}_{\pm 0.39}}$              & ${{2.53}_{\pm 0.47}}$             & ${{2.20}_{\pm 0.10}}$           \\ \midrule
      LayerNorm                     & ${{{29.22}_{\pm 0.04}}}$                   & ${{12.22}_{\pm 0.00}}$                   & ${{{19.78}_{\pm 0.02}}}$                  & ${{18.31}_{\pm 0.03}}$                  & ${{18.62}_{\pm 0.06}}$                & ${{13.90}_{\pm 0.01}}$                          & ${{38.12}_{\pm 0.03}}$                         & ${{38.56}_{\pm 0.01}}$                & ${{6.29}_{\pm 0.01}}$                        & ${{6.40}_{\pm 0.01}}$                       & ${{{8.17}_{\pm 0.01}}}$                  & ${{56.23}_{\pm 0.02}}$                  & ${{{35.84}_{\pm 0.01}}}$         & ${{{28.20}_{\pm 0.03}}}$           & ${{{28.69}_{\pm 0.04}}}$          & ${{23.90}_{\pm 0.01}}$          \\
      LayerNorm (w/o EMA teacher)   & ${{20.74}_{\pm 0.88}}$                     & ${{{15.84}_{\pm 2.13}}}$                 & ${{15.24}_{\pm 0.87}}$                    & ${{{18.87}_{\pm 3.05}}}$                & ${{{22.22}_{\pm 0.69}}}$              & $\mathbf{{{21.74}_{\pm 1.07}}}$                 & $\mathbf{{{42.08}_{\pm 1.85}}}$                & $\mathbf{{{45.51}_{\pm 0.30}}}$       & ${{{6.97}_{\pm 0.15}}}$                      & ${{{7.54}_{\pm 0.50}}}$                     & ${{7.47}_{\pm 0.39}}$                    & $\mathbf{{{57.83}_{\pm 0.19}}}$         & ${{31.65}_{\pm 4.15}}$           & ${{23.47}_{\pm 0.72}}$             & ${{23.47}_{\pm 2.39}}$            & ${{{24.04}_{\pm 0.61}}}$        \\
      LoRA (ours)                   & $\mathbf{{{31.51}_{\pm 0.19}}}$            & $\mathbf{{{16.76}_{\pm 1.62}}}$          & $\mathbf{{{23.47}_{\pm 0.74}}}$           & $\mathbf{{{20.40}_{\pm 0.80}}}$         & $\mathbf{{{22.81}_{\pm 0.27}}}$       & ${{{16.59}_{\pm 0.40}}}$                        & ${{{42.03}_{\pm 0.23}}}$                       & ${{{42.38}_{\pm 0.26}}}$              & $\mathbf{{{7.56}_{\pm 3.50}}}$               & $\mathbf{{{10.60}_{\pm 0.41}}}$             & $\mathbf{{{11.36}_{\pm 0.86}}}$          & ${{{57.75}_{\pm 0.11}}}$                & $\mathbf{{{39.16}_{\pm 0.34}}}$  & $\mathbf{{{31.65}_{\pm 0.48}}}$    & $\mathbf{{{32.40}_{\pm 0.13}}}$   & $\mathbf{{{27.10}_{\pm 0.12}}}$ \\ \bottomrule
    \end{tabular}
  }
\end{table}

\end{document}